\documentclass{article} % For LaTeX2e
\usepackage{iclr2025_conference,times}

\usepackage{amsmath,amsfonts,bm}

\def\eqref#1{equation~\ref{#1}}
\def\1{\bm{1}}

\DeclareMathAlphabet{\mathsfit}{\encodingdefault}{\sfdefault}{m}{sl}
\SetMathAlphabet{\mathsfit}{bold}{\encodingdefault}{\sfdefault}{bx}{n}

\usepackage{hyperref}
\usepackage{url}

\usepackage{xurl}
\usepackage{graphicx}
\usepackage{amsmath}
\usepackage{amsthm}
\usepackage{booktabs}
\usepackage{algorithm}
\usepackage{algpseudocode}
\algrenewcommand\alglinenumber[1]{\tiny #1:}
\usepackage[switch]{lineno}

\usepackage{subfigure}

\usepackage{amsfonts,bm}
\usepackage{mathtools}
\usepackage{wrapfig}
\usepackage{soul}
\usepackage{multirow}
\usepackage[most]{tcolorbox}
\usepackage{xcolor,colortbl}
\definecolor{DPLightGreen}{rgb}{0.307, 0.610, 0.446}
\definecolor{DPLightBlue}{rgb}{0.541, 0.585, 0.951}
\definecolor{DPLightRed}{rgb}{0.941, 0.585, 0.551}
\definecolor{DPLightOrange}{RGB}{147, 140, 122}

\definecolor{RBRed}{rgb}{0.98,0.88,0.85}
\definecolor{RBRedLight}{rgb}{1,0.93,0.90}

\definecolor{RBBlue}{rgb}{0.81,0.80,0.94}
\definecolor{RBBlueLight}{rgb}{0.91,0.90,0.98}

\definecolor{RBBlueL}{rgb}{0.84,0.83,0.97}
\definecolor{RBBlueLL}{rgb}{0.86,0.85,0.99}

\definecolor{RBGreen}{rgb}{0.85,0.91,0.84}
\definecolor{RBGreenLight}{rgb}{0.93,0.98,0.92}

\definecolor{RBOrange}{RGB}{250,240,210}

\definecolor{DPBlueD}{rgb}{0.24,0.43,0.77}
\definecolor{DPPurple}{rgb}{0.46,0.12,0.45}

\definecolor{DPGreen}{rgb}{0,0.45,0.24}
\definecolor{DPLightGreen}{rgb}{0,0.65,0.44}

\newtcbox{\dpboxred}{on line,
  colframe=DPLightRed,colback=DPLightRed!10!white,
  boxrule=0.5pt,arc=2pt,boxsep=0pt,left=2pt,right=2pt,top=2pt,bottom=2pt}
  
\newtcbox{\dpboxorange}{on line,
  colframe=DPLightOrange,colback=RBOrange!10!white,
  boxrule=0.5pt,arc=2pt,boxsep=0pt,left=2pt,right=2pt,top=2pt,bottom=2pt}

\newtcbox{\dpboxgreen}{on line,
  colframe=DPLightGreen,colback=DPLightGreen!10!white,
  boxrule=0.5pt,arc=2pt,boxsep=0pt,left=2pt,right=2pt,top=2pt,bottom=2pt}

\newtcbox{\dpboxblue}{on line,
  colframe=DPLightBlue,colback=DPLightBlue!10!white,
  boxrule=0.5pt,arc=2pt,boxsep=0pt,left=2pt,right=2pt,top=2pt,bottom=2pt}

\definecolor{forestgreen}{rgb}{0.60, 0.27, 0.06}

\usepackage{tikz}

\definecolor{primarybox}{RGB}{240,248,255}    % Light Azure
\definecolor{primaryline}{RGB}{70,130,180}    % Steel Blue
\definecolor{secondarybox}{RGB}{255,250,240}  % Cornsilk
\definecolor{secondaryline}{RGB}{210,105,30}  % Chocolate

\newenvironment{takeaways}[1][]{%
    \begin{tcolorbox}[
        colback=secondarybox,
        colframe=secondaryline,
        boxrule=0.5mm,
        sharp corners=all,
        before skip=10pt,
        after skip=10pt,
        left=4mm,
        right=4mm,
        top=2mm,
        bottom=2mm,
        parbox=false,
        breakable,
        title={TL;DR (#1)}
    ]%
}{%
    \end{tcolorbox}%
}

\title{Human–AI Ensembles Improve Deepfake Detection in Low-to-Medium Quality Videos}

\author{
\\
\\
\texttt{\{hippo,brain,jen\}@cs.cranberry-lemon.edu} \\
}

\author{Marco Postiglione, 
Isabel Gortner,
V. S. Subrahmanian \\
Northwestern University\\
\texttt{vss@northwestern.edu}}

\iclrfinalcopy % Uncomment for camera-ready version, but NOT for submission.
\begin{document}

\maketitle

\begin{abstract}
Deepfake detection is widely framed as a machine learning problem, yet how humans and AI detectors compare under realistic conditions remains poorly understood. We evaluate 200 human participants and 95 state-of-the-art AI detectors across two datasets: DF40, a standard benchmark, and CharadesDF, a novel dataset of videos of everyday activities. CharadesDF was recorded using mobile phones leading to low/moderate quality videos compared to the more professionally captured DF40. Humans outperform AI detectors on both datasets, with the gap widening in the case of CharadesDF where AI accuracy collapses to near chance (0.537) while humans maintain robust performance (0.784). Human and AI errors are complementary: humans miss high-quality deepfakes while AI detectors flag authentic videos as fake, and hybrid human-AI ensembles reduce high-confidence errors. These findings suggest that effective real-world deepfake detection, especially in non-professionally produced videos, requires human–AI collaboration rather than AI algorithms alone.
\end{abstract}

\tableofcontents
\section{Introduction}

The emergence of deepfake technology (i.e., artificial intelligence systems capable of generating synthetic media depicting events that may have never occurred) has fundamentally altered the landscape of digital trust~\citep{DBLP:conf/iccv/RosslerCVRTN19,li2020celeb,DBLP:conf/nips/YanYCZFZLWDWY24,kobis2021fooled,appel2022political,sanchezacedo2024mil}. From non-consensual intimate imagery to political disinformation campaigns, deepfakes have been linked to various negative effects~\citep{cuevas2026deepfake,kobis2021fooled,preu2022perception,appel2022political,sanchezacedo2024mil}. As generative AI capabilities continue to advance at a remarkable pace, the question of how society can reliably distinguish authentic from synthetic media has become increasingly urgent~\citep{cooke2024coin,frank2024representative,kamali2025diffusion}.

In early 2025, a US judge asked us to produce a deepfake video of him chopping wood. He sent us a real video, and we produced deepfake versions where the judge's face was swapped with the face of two consenting individuals. AI algorithms were unable to correctly classify the fake videos. We hypothesized that this may have been due to several factors: (i) the source footage was recorded in an open space with substantial camera movement and frequent zooming in and out. (ii) There were also frames in which the subject's face was only partially visible (e.g., when bringing axe down from above the head to the chopping block). (iii) The face itself was not always very large in size. These conditions are typical of authentic user-generated content but largely absent from existing detection benchmarks. The resulting deepfakes proved difficult for both human observers and state-of-the-art AI detectors to identify as fake, underscoring that detection performance measured on high-quality, well-framed benchmark videos may not transfer to the kinds of footage that a normal citizen might record with their mobile phone or similar device.

Despite the extensive focus on AI-based detection, relatively little systematic research has examined how \textit{humans} perform at detecting deepfakes, or whether human and AI detection capabilities might be combined to achieve more robust detection. Existing work shows that humans often struggle to detect deepfakes or other AI-generated media, with performance frequently hovering near chance~\citep{kobis2021fooled,bray2022deepfakefaces,cooke2024coin,frank2024representative,hashmi2024audiovisual,mueller2021audiodeepfake,somoray2025review}. Prior studies have documented that humans struggle to detect deepfakes under some conditions, but these findings are typically based on small stimulus sets, controlled laboratory conditions, or specific forgery techniques that may not generalize~\citep{bray2022deepfakefaces,preu2022perception,pehlivanoglu2026susceptibility,josephs2023artifactmag,somoray2025review}. At the same time, comparisons of human and AI-based detection have yielded mixed results: some works report automated detectors clearly outperforming humans on standard benchmarks~\citep{DBLP:conf/iccv/RosslerCVRTN19,korshunov2021subjective,hashmi2024audiovisual}, whereas others find human observers matching or even exceeding model performance in certain regimes or modalities~\citep{groh2021deepfake,goh2024humansvsmachines,pehlivanoglu2026susceptibility,mueller2021audiodeepfake}. \textit{The field lacks comprehensive, head-to-head comparisons of human and AI detection across datasets that vary systematically in their realism and difficulty}.

This paper addresses these gaps through a comprehensive empirical investigation comparing human and AI deepfake detection. We evaluate 200 human participants and 95 state-of-the-art AI deepfake detectors spanning several architectural paradigms including frequency-based methods~\citep{qian2020f3net,liu2021spsl,luo2021srm}, attention mechanisms~\citep{zhao2021multiattention,dang2020ffd}, reconstruction-based approaches~\citep{cao2022recce}, contrastive learning~\citep{yan2023ucf,zhao2021pcl}, and transformer architectures~\citep{zhuang2022uiavit,bertasius2021timesformer}, on two complementary datasets.

The first dataset is DF40~\citep{DBLP:conf/nips/YanYCZFZLWDWY24}, a recently published benchmark featuring state-of-the-art deepfakes of YouTube-sourced videos. DF40 and similar benchmarks typically feature well-lit subjects in frontal poses with consistently visible faces~\citep{DBLP:conf/iccv/RosslerCVRTN19,li2020celeb}. These conditions diverge substantially from authentic user-generated content, where variability in camera movement, lighting, occlusion, and camera angle is the norm~\citep{josephs2023artifactmag,cooke2024coin,frank2024representative}. This gap matters because real-world detection (e.g., in legal proceedings, at protest marches and/or civic events) frequently involves videos depicting ordinary activities such as nodding in agreement, entering a building, or signing a document. Such recordings might be fabricated to incriminate, or genuine footage might be dismissed as synthetic by those seeking to evade accountability.

\begin{figure}[t]
\centering
\includegraphics[width=\textwidth]{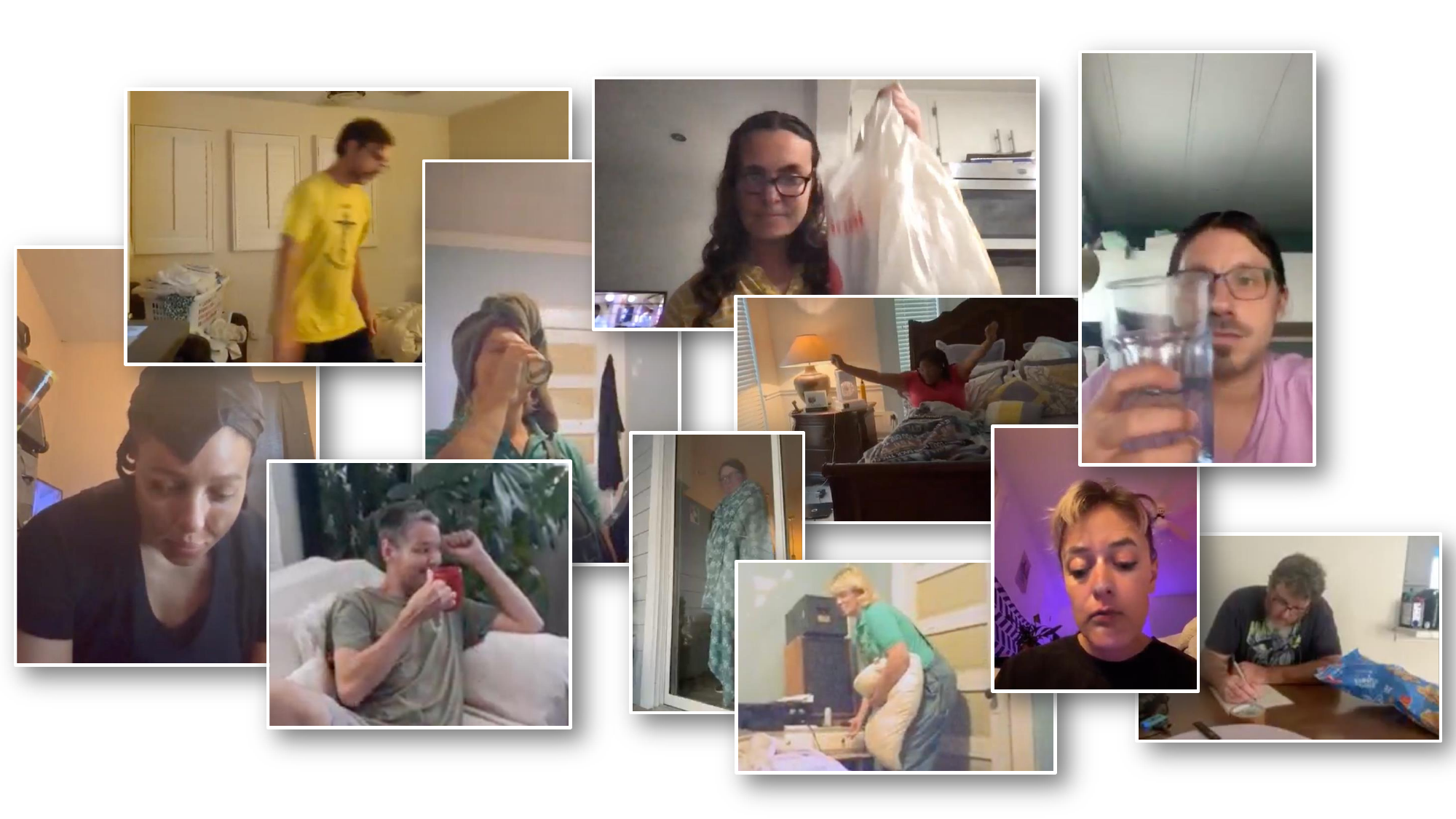}
\caption{\textbf{Examples of video recordings from the CharadesDF dataset.} Participants recorded videos in their home environments while performing everyday activities such as drinking from cups, holding objects, stretching, and working at desks. These conditions produce substantial variability in lighting, camera angles, face visibility, and image quality—challenges common in authentic user-generated content but underrepresented in existing deepfake detection benchmarks.}
\label{fig:charadesdf_examples}
\end{figure}

To address this gap, we developed CharadesDF, a novel dataset of authentic home recordings and corresponding deepfakes (Figure~\ref{fig:charadesdf_examples}). The name derives from the Charades dataset~\citep{DBLP:conf/eccv/SigurdssonVWFLG16}, a video corpus of everyday indoor activities whose name alludes to the parlor game of acting out scenes. We sourced our activity instructions from Charades, asking participants to perform scripted actions such as drinking from a coffee cup, opening a closet, and moving around rooms in their homes. We recruited participants via an IRB-approved study and generated deepfakes using publicly available face-swapping tools, simulating the realistic scenario of non-experts creating synthetic media with accessible software~\citep{tahir2021seeing,preu2022perception}.

We address five research questions:

\begin{enumerate}
    \item[\textbf{RQ1}] \textbf{How does human detection accuracy compare to state-of-the-art AI detectors? }Prior studies have typically compared humans to a small number of AI detectors on a single dataset or modality~\citep{groh2021deepfake,korshunov2021subjective,hashmi2024audiovisual,mueller2021audiodeepfake,goh2024humansvsmachines}, leaving open how these comparisons generalize across different architectures and recording conditions.
    
    \item[\textbf{RQ2}] \textbf{Does combining multiple assessments improve detection, and do humans and AI detectors contribute complementary strength?} Earlier work suggests that ensembles of human crowds and models can outperform either alone, but are vulnerable to automation bias and misleading AI advice~\citep{groh2021deepfake,boyd2022aiguidance,josephs2023artifactmag,ibsen2024conditional,casu2025automationbias}. We extend this line of inquiry by systematically aggregating predictions made by varied model families with human assessments across both controlled and realistic datasets.
    
    \item[\textbf{RQ3}] \textbf{How do video quality factors (e.g., face size, lighting, image noise, and pose variation) influence deepfake detection performance?} Existing results indicate that compression, channel noise, and viewing conditions strongly modulate both human and AI accuracy~\citep{DBLP:conf/iccv/RosslerCVRTN19,prasad2022noisychannels,josephs2023artifactmag,woehler2021perceptual}, but their relative impact across AI detector classes and human observers under realistic, heterogeneous content remains poorly understood.

    \item[\textbf{RQ4}] \textbf{What is the relationship between confidence and accuracy in deepfake detection?} Prior work has documented substantial miscalibration, including truth bias, Dunning--Kruger effects, and conflict detection signals when people misclassify convincing fakes~\citep{kobis2021fooled,miller2023aihyperreal,janssen2024conflict,chen2025selfefficacy,casu2025automationbias}. We examine whether similar patterns arise in our setting and how confidence (in either the human's prediction or an AI detector's prediction) is related to prediction accuracy in the two datasets (DF40 which has relatively quality video, and CharadesDF which has relatively low/moderate quality video).

    \item[\textbf{RQ5}] \textbf{Do demographic characteristics predict individual differences in deepfake detection performance?} Recent studies have linked detection ability to factors such as age, analytic thinking, media literacy, and face-recognition aptitude, with generally small-to-moderate effect sizes and mixed findings~\citep{frank2024representative,pehlivanoglu2026susceptibility,jin2025visual,alsobeh2024unmasking,davis2025superrecogniser,mueller2021audiodeepfake,xie2025seeing}. We test whether demographic variables explain variance in performance in our more ecologically realistic setting.
\end{enumerate}

Our findings demonstrate that humans substantially outperform AI detectors, particularly under realistic conditions where AI detectors approach chance-level performance, extending and qualifying prior mixed evidence on human--AI comparisons~\citep{DBLP:conf/iccv/RosslerCVRTN19,groh2021deepfake,korshunov2021subjective,hashmi2024audiovisual,goh2024humansvsmachines,pehlivanoglu2026susceptibility}. We found that human and AI errors are complementary, enabling hybrid ensembles involving both humans and AI detectors to eliminate highly-confident errors, consistent with prior work~\citep{groh2021deepfake,boyd2022aiguidance,ibsen2024conditional}. Finally, we find that most human demographic factors do not reliably predict human detection ability, suggesting that human performance depends on factors beyond general demographics and technological proficiency~\citep{somoray2025review,frank2024representative,davis2025superrecogniser}. Together, these results indicate that the optimal approach to real-world deepfake detection may not be more sophisticated machine learning, but rather an ensemble of human and AI detectors that leverages complementary strengths~\citep{groh2021deepfake,josephs2023artifactmag,ibsen2024conditional}, especially when the videos are of low/medium quality.

 \section{Results}

We conducted two complementary experiments to compare human and AI deepfake detection capabilities. For AI detection evaluation, we trained 32 state-of-the-art deepfake detection architectures on three training datasets (i.e., FaceForensics++~\citep{DBLP:conf/iccv/RosslerCVRTN19}, CelebDF-v2~\citep{li2020celeb}, and the DF40 training split~\citep{DBLP:conf/nips/YanYCZFZLWDWY24}) yielding 96 detector variants (with one excluded due to architectural incompatibility). Each variant was evaluated on two test sets: a sample of 1{,}000 videos from the held-out DF40 test set (500 real, 500 fake across 10 forgery techniques) and CharadesDF, a novel dataset we developed comprising 1{,}000 videos (500 authentic home recordings and 500 corresponding deepfakes generated using five face-swapping models). The DF40 sample features YouTube-sourced videos with subjects in predominantly frontal poses and consistently visible faces, while CharadesDF was designed to simulate challenging real-world conditions. Participants recorded videos in their home environments using personal devices while following scripted activity instructions, resulting in greater variability in image quality, camera angles, and face visibility.

For human evaluation, we conducted two IRB-approved studies via the Prolific platform, recruiting 100 participants for each dataset. Due to practical constraints on participant time, each participant viewed a randomly sampled subset of 60 videos from the corresponding 1{,}000-video pool, ensuring each video received multiple independent ratings. Participants rated their confidence that each video was fabricated on a 5-point Likert scale, which we converted to continuous probability scores ($p \in [0, 1]$, where $p = 1$ indicates maximum confidence in fabrication). We computed accuracy as the proportion of correct binary classifications at the $p = 0.5$ threshold, along with AUC and F1 scores on the ``deepfake'' class. No overlap exists between training and evaluation data for any detector, and Phase~1 participants (who recorded CharadesDF videos) were excluded from Phase~2 evaluation to prevent familiarity effects. Further details on dataset construction, model training, quality feature extraction, and participant demographics are provided in the Methods section.

\subsection{Research Question 1: Human vs. AI Detection Performance}
\textit{How does human detection accuracy compare to state-of-the-art AI detectors?}  

\begin{takeaways}[RQ1]
    \begin{enumerate}
        \item \textbf{Humans outperform AI at detecting deepfakes.} Across both datasets and analysis levels, human participants achieved significantly higher accuracy than AI detectors.
                
        \item \textbf{AI detectors performance degrades more on challenging datasets.} On CharadesDF, AI deepfake detectors performed near chance level ($\mu =$ 0.537), while humans maintained stronger accuracy ($\mu =$ 0.784), suggesting AI detectors are more sensitive to dataset characteristics.
        
    \end{enumerate}
\end{takeaways}

\begin{figure}[t]
    \centering
    \includegraphics[width=0.48\textwidth]{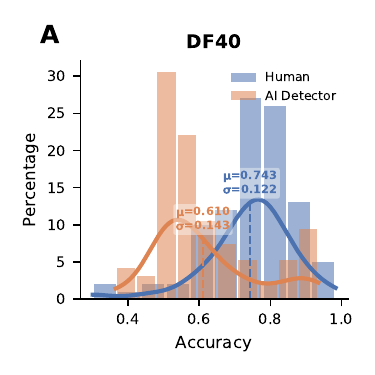}
    \hfill
    \includegraphics[width=0.48\textwidth]{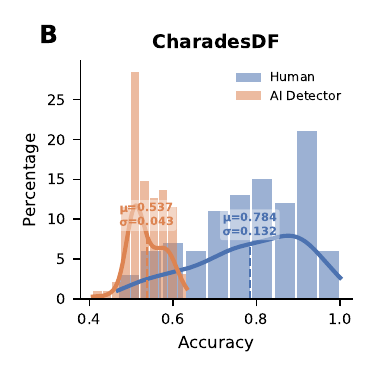}
    \caption{\textbf{Distribution of participant-level accuracy for human participants and AI detectors.} (\textbf{A}) DF40 dataset. (\textbf{B}) CharadesDF dataset. Each data point represents one participant's overall accuracy across all videos they evaluated.}
    \label{fig:RQ1_participant_accuracy}
\end{figure}

We compared performance at the participant level, treating each human participant and each AI classifier as an individual detector. Figure~\ref{fig:RQ1_participant_accuracy} displays the distribution of accuracy scores across all participants and classifiers for both datasets.  Supplementary Figure~\ref{fig:RQ1_video_accuracy} complements this analysis with aggregated predictions at the video level.

On DF40 (Figure~\ref{fig:RQ1_participant_accuracy}A), human participants achieved a mean accuracy of 0.743 (SD = 0.122), significantly outperforming AI detectors ($\mu =$ 0.610, SD = 0.143), $t(193) = 6.970$, $p < .001$, Cohen's $d = 0.999$. This performance gap widened substantially on CharadesDF (Figure~\ref{fig:RQ1_participant_accuracy}B), where humans maintained robust accuracy ($\mu =$ 0.784, SD = 0.132) while AI detectors collapsed to near-chance performance ($\mu =$ 0.537, SD = 0.043), $t(193) = 17.389$, $p < .001$, Cohen's $d = 2.491$. AI detectors in Figure~\ref{fig:RQ1_participant_accuracy}B are narrowly concentrated around 50\%, indicating that the challenging, naturalistic conditions of CharadesDF rendered AI detection essentially ineffective. The reduced variance among AI detectors on CharadesDF (SD = 0.043 vs. 0.143 on DF40) suggests that varied architectural approaches converged to similar failure modes when confronted with realistic recording conditions.

\subsection{Research Question 2: Ensemble Aggregation and Human–AI Complementarity}
\textit{Does combining multiple assessments improve detection, and do humans and AI detectors contribute complementary strength?} 

\begin{takeaways}[RQ2]
    \begin{enumerate}
        \item \textbf{Aggregating assessments substantially improves detection accuracy.} Human ensembles outperformed individual humans by 14–15 percentage points, while AI ensembles improved by 7–26 percentage points depending on dataset.
        
        \item \textbf{Ensemble methods dramatically reduce high-confidence mistakes.} Individual judges were confidently wrong in 17–32\% of cases; ensembles reduced this to under 2\%, with hybrid human-AI ensembles eliminating such errors entirely.
        
        \item \textbf{Human and AI errors do not overlap.} When humans made high-confidence errors, AI detectors tended to be correct, and vice versa.
        
        \item \textbf{Humans and AI detectors fail in qualitatively different ways.} Humans predominantly misclassify deepfakes as real, while AI detectors predominantly misclassify real videos as deepfakes.

        \item \textbf{AI ensemble gains are sensitive to domain shift.} Ensembles of AI detectors improved markedly on DF40 but showed modest gains on CharadesDF, reflecting vulnerability to distribution shift from training data.
        
        \item \textbf{Hybrid ensembles leverage complementary failure patterns.} By combining human and AI predictions, hybrid ensembles exploit the fact that when one group fails, the other typically succeeds.

    \end{enumerate}
\end{takeaways}

To assess whether aggregating assessments could improve detection accuracy, we compared individual performance against ensemble methods that combined predictions within and across human and AI judges (Table~\ref{tab:ensemble_results}). We report accuracy, F1 score, area under the ROC curve (AUC), and catastrophic failure rate (CFR). CFR is defined as the proportion of predictions with absolute error exceeding 0.7, capturing cases where confident predictions were fundamentally wrong. This metric is conceptually related to the complement of threshold accuracy measures \citep{DBLP:conf/nips/EigenPF14}, where the proportion of predictions falling within an acceptable error bound is reported. Here, we instead measure the proportion exceeding it, emphasizing tail failures. We experimented with CFR at multiple thresholds, spanning from 0.5 to 1. 

For ensemble aggregation, we employed quality-weighted voting, which assigns greater influence to more accurate judges. To avoid circularity, we computed weights using a leave-one-out procedure. Specifically, for judge $j$ predicting video $v$, the weight $w_j^{(-v)}$ reflects that judge's accuracy on all videos \textit{except} $v$:
\begin{equation}
w_j^{(-v)} = \frac{1}{|V_j| - 1} \sum_{v' \in V_j \setminus \{v\}} \mathbf{1}\left[\hat{y}_{j,v'} = y_{v'}\right]
\label{eq:loo_weight}
\end{equation}
where $V_j$ is the set of videos evaluated by judge $j$, $\hat{y}_{j,v'}$ is the binarized prediction (threshold = 0.5), and $y_{v'}$ is the ground truth label. The ensemble prediction for video $v$ is then the weighted average of probability scores:
\begin{equation}
p_v^{\text{ensemble}} = \frac{\sum_{j} w_j^{(-v)} \cdot p_{j,v}}{\sum_{j} w_j^{(-v)}}
\label{eq:ensemble_pred}
\end{equation}
where $p_{j,v} \in [0,1]$ is judge $j$'s raw probability score for video $v$.
For hybrid human--AI ensembles, we used a two-stage hierarchical aggregation to prevent the larger group from dominating the final prediction. First, predictions were aggregated separately within human and AI groups using quality-weighted voting (Equations~\ref{eq:loo_weight}--\ref{eq:ensemble_pred}), yielding group-level predictions $p_v^{\text{human}}$ and $p_v^{\text{AI}}$. These group-level predictions were then combined with equal weighting:
\begin{equation}
p_v^{\text{hybrid}} = \frac{1}{2}\left(p_v^{\text{human}} + p_v^{\text{AI}}\right)
\label{eq:hybrid_pred}
\end{equation}
This balanced approach ensures that the complementary strengths of human intuition and algorithmic pattern detection contribute equally to the final prediction, regardless of the number of judges in each group. {Results using alternative aggregation strategies (simple averaging, median voting, and majority voting) are provided in Supplementary Tables~\ref{tab:ensemble_results_soft_mean}--\ref{tab:ensemble_results_hard}}.

\begin{table}[t]
\centering
\small
\caption{\textbf{Comparison of individual versus ensemble performance on deepfake detection. }Values shown as point estimate $\pm$ error. Error represents half-width of 95\% bootstrap CI. CFR = Catastrophic Failure Rate. Bold indicates best ensemble performance.}
\label{tab:ensemble_results}
\scalebox{0.66}{
\begin{tabular}{l cccccccc}
\toprule
 & \multicolumn{4}{c}{\textbf{DF40}} & \multicolumn{4}{c}{\textbf{CharadesDF}} \\
\cmidrule(lr){2-5} \cmidrule(lr){6-9} 
 & Accuracy & F1 & AUC & CFR & Accuracy & F1 & AUC & CFR \\
\midrule
Human (indiv.) & 0.743 {\small $\pm$ 0.023} & 0.759 {\small $\pm$ 0.020} & 0.823 {\small $\pm$ 0.020} & 0.190 {\small $\pm$ 0.017} & 0.784 {\small $\pm$ 0.027} & 0.783 {\small $\pm$ 0.027} & 0.846 {\small $\pm$ 0.024} & 0.171 {\small $\pm$ 0.025} \\
AI detector (indiv.) & 0.610 {\small $\pm$ 0.030} & 0.612 {\small $\pm$ 0.042} & 0.686 {\small $\pm$ 0.037} & 0.279 {\small $\pm$ 0.036} & 0.537 {\small $\pm$ 0.009} & 0.533 {\small $\pm$ 0.041} & 0.583 {\small $\pm$ 0.014} & 0.317 {\small $\pm$ 0.029} \\
\midrule
Human ensemble & 0.890 {\small $\pm$ 0.020} & 0.888 {\small $\pm$ 0.021} & 0.952 {\small $\pm$ 0.013} & 0.019 {\small $\pm$ 0.009} & \textbf{0.928} {\small $\pm$ 0.017} & \textbf{0.926} {\small $\pm$ 0.019} & 0.979 {\small $\pm$ 0.009} & 0.010 {\small $\pm$ 0.007} \\
AI ensemble & 0.869 {\small $\pm$ 0.020} & 0.876 {\small $\pm$ 0.021} & 0.949 {\small $\pm$ 0.012} & 0.001 {\small $\pm$ 0.002} & 0.607 {\small $\pm$ 0.033} & 0.683 {\small $\pm$ 0.034} & 0.697 {\small $\pm$ 0.035} & 0.009 {\small $\pm$ 0.006} \\
Hybrid ensemble & \textbf{0.941} {\small $\pm$ 0.014} & \textbf{0.941} {\small $\pm$ 0.014} & \textbf{0.988} {\small $\pm$ 0.005} & \textbf{0.000} {\small $\pm$ 0.000} & 0.924 {\small $\pm$ 0.018} & 0.924 {\small $\pm$ 0.019} & \textbf{0.979} {\small $\pm$ 0.008} & \textbf{0.000} {\small $\pm$ 0.000} \\
\bottomrule
\end{tabular}}
\end{table}

\paragraph{Quantifying ensemble gains across aggregation methods}
Individual human judges achieved moderate accuracy (DF40: 0.743, 95\% CI [0.718, 0.766]; CharadesDF: 0.784, 95\% CI [0.758, 0.810]), while individual AI detectors performed at or near chance levels (DF40: 0.610, 95\% CI [0.583, 0.639]; CharadesDF: 0.537, 95\% CI [0.529, 0.546]; Table~\ref{tab:ensemble_results}). Aggregating assessments within each group yielded substantial improvements. Human ensembles reached 0.890 (95\% CI [0.870, 0.909]) on DF40 and 0.928 (95\% CI [0.909, 0.945]) on CharadesDF---absolute improvements of 14.7 and 14.4 percentage points over individual humans, respectively (DF40: $t(263) = 9.35$, $p < 0.001$, Cohen's $d = 0.62$; CharadesDF: $t(214) = 8.94$, $p < 0.001$, Cohen's $d = 0.70$). AI ensembles improved markedly on DF40 (0.869, 95\% CI [0.848, 0.889]; $+$25.8 percentage points; $t(214) = 14.19$, $p < 0.001$, Cohen's $d = 0.99$), though gains on CharadesDF were more modest (0.607, 95\% CI [0.573, 0.641]; $+$7.0 percentage points; $t(851) = 3.89$, $p < 0.001$, Cohen's $d = 0.20$), likely reflecting the domain shift from training data that disproportionately affected algorithmic detectors.

\paragraph{High-confidence error analysis}
Beyond improving average accuracy, ensemble methods dramatically reduced catastrophic failures. Individual humans exhibited catastrophic failure rates of 19.0\% (DF40) and 17.0\% (CharadesDF), while individual detectors failed catastrophically in 27.9\% and 31.7\% of cases, respectively. Human ensembles reduced these rates to 1.9\% and 1.0\% (Fisher's exact test: DF40, $p < 0.001$; CharadesDF, $p < 0.001$), while AI ensembles achieved 0.1\% and 0.9\% (DF40, $p < 0.001$; CharadesDF, $p < 0.001$). Hybrid ensembles did not show catastrophic failures.

\begin{figure}[t]
    \centering
    \includegraphics[width=\textwidth]{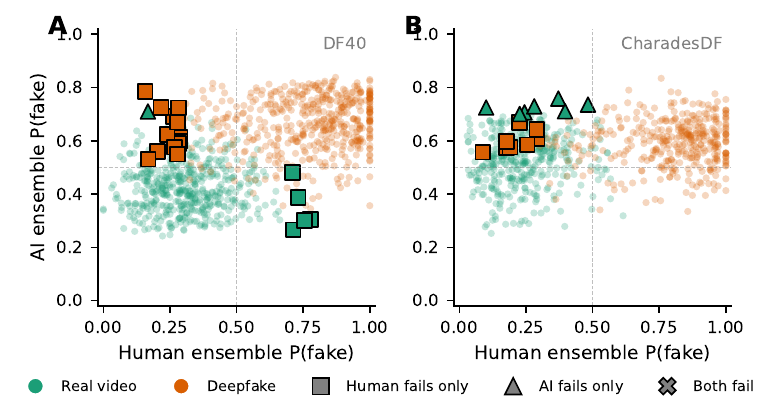}
    \caption{\textbf{Complementary failure patterns between human and AI ensembles.} 
    Each point represents a video, plotted by human ensemble probability (x-axis) versus AI ensemble probability (y-axis) of being a deepfake. 
    Dashed lines indicate the decision threshold (0.5). 
    Small circles show videos without catastrophic failures (teal: real videos; orange: deepfakes). 
    Squares indicate videos where only the human ensemble failed catastrophically; triangles indicate videos where only the AI ensemble failed. 
    No videos caused catastrophic failures in both groups simultaneously (no X markers).
    \textbf{(A)}~DF40 dataset. \textbf{(B)}~CharadesDF dataset.}
    \label{fig:complementarity}
\end{figure}

\paragraph{Characterizing human-AI error independence}
The effectiveness of hybrid ensembles stems from complementary failure patterns between human and AI judges (Figure~\ref{fig:complementarity}). When we classified each video by whether human ensembles, AI ensembles, or both failed catastrophically, we found no overlap: no videos caused catastrophic failures in both groups simultaneously on either dataset. The phi coefficient measuring association between human and AI failures was near zero (DF40: $\phi = -0.004$; CharadesDF: $\phi = -0.010$), confirming that errors were statistically independent (Fisher's exact test: DF40, $p = 1.000$; CharadesDF, $p = 1.000$). 

Moreover, failure patterns differed qualitatively by ground truth. When only humans failed catastrophically, they predominantly classified deepfakes as real (DF40: 14 of 19 failures, 74\%; CharadesDF: 8 of 8, 100\%), suggesting vulnerability to high-quality face manipulations. In contrast, AI-only failures predominantly misclassified real videos as deepfakes (DF40: 1 of 1 failures, 100\%; CharadesDF: 7 of 7, 100\%), potentially due to over-reliance on compression artifacts present in both authentic and manipulated content (Fisher's exact test comparing error types: DF40 humans $p = 0.039$, AI $p = 1.000$; CharadesDF humans $p = 0.004$, AI $p = 0.015$).
This complementarity explains the hybrid ensemble's effectiveness: when humans failed, AI detectors provided correct predictions, and vice versa. By averaging predictions, hybrid ensembles leveraged this complementarity to eliminate catastrophic failures.

\begin{figure}[t]
    \centering
    \includegraphics[width=\linewidth]{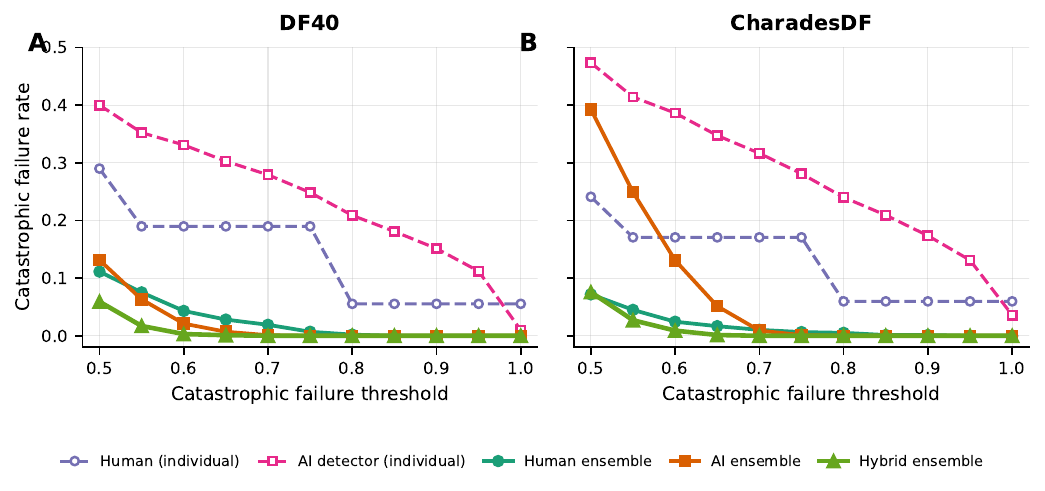}
    \caption{\textbf{Catastrophic failure rates across threshold definitions.} 
    CFR was computed for threshold values ranging from 0.5 (any probability error $\ge 0.5$ is catastrophic) to 1.0 (only complete misclassification with probability error = 1.0). 
    Dashed lines show mean individual performance; solid lines show ensemble performance. 
    Quality-weighted voting was used for all ensembles.
    \textbf{(A)}~DF40 dataset. \textbf{(B)}~CharadesDF dataset.}
\label{fig:cfr_threshold}
\end{figure}

\paragraph{Robustness of ensemble benefits across threshold definitions}
The preceding analyses defined catastrophic failures using a threshold of 0.7 (probability errors $\ge 0.7$). To assess robustness, we examined how catastrophic failure rates varied across threshold definitions from 0.5 to 1.0 (Figure~\ref{fig:cfr_threshold}). Across both datasets, ensemble methods consistently outperformed individual methods at every threshold examined. AI detectors exhibited the highest CFR across all thresholds, particularly on CharadesDF where CFR exceeded 47\% at the most lenient threshold (0.5). Human ensembles reduced CFR substantially compared to individual humans, reaching zero failures by threshold 0.85 on DF40 and 0.95 on CharadesDF. AI ensembles reached zero failures by threshold 0.75 on DF40, though CharadesDF required threshold 0.8, reflecting the domain shift challenges that disproportionately affected algorithmic detectors. Hybrid ensembles achieved zero catastrophic failures by threshold 0.7 on both datasets, demonstrating that complementarity benefits extend beyond any particular operational definition of catastrophic failure. The AI ensemble on CharadesDF showed notably elevated CFR at lenient thresholds (39.3\% at 0.5), but hybrid ensembles mitigated this vulnerability by incorporating human assessments that were less susceptible to distribution shift.

\subsection{Research Question 3: Visual Quality Factors and Detection Performance}
\textit{How do video quality factors influence deepfake detection performance?
}
\begin{takeaways}[RQ3]
    \begin{enumerate}
        \item \textbf{Face size is the strongest and most consistent predictor of detection accuracy.} Larger, more visible faces substantially improve performance for both humans and AI detectors.
        
        \item \textbf{AI detectors are more sensitive to low-level visual quality than humans.} Factors such as signal-to-noise ratio, color balance, contrast, and overall quality primarily affect AI performance, with limited impact on humans.
        
        \item \textbf{Face scaling affects AI detectors and humans differently.} Inter-ocular distance negatively impacts AI accuracy but does not significantly affect humans, indicating a specific vulnerability of AI detectors to scaling-related artifacts.
        
        \item \textbf{Humans benefit from interpretable cues, such as face recognition confidence.} Higher face recognizability improves human accuracy, while AI detectors rely less on this signal.
        
        \item \textbf{Some visual conditions impair humans but can help AI detectors.} For example, exaggerated facial expressions reduce human accuracy but improve AI ensemble performance, suggesting different information-processing strategies.
        
        \item \textbf{Lighting and brightness effects are dataset-dependent.} For instance, increased brightness reduces AI accuracy in CharadesDF but shows no effect in DF40.
        
        \item \textbf{Most other visual features have limited impact.} Variables such as blur, occlusion, facial aspect ratios, and most pose variations do not significantly influence detection after correction for multiple comparisons.
        
        \item \textbf{Visual quality explains only a modest portion of performance.} Detection accuracy is influenced by multiple factors beyond simple visual quality, as reflected by relatively low $R^2$ values.
    \end{enumerate}
\end{takeaways}

\paragraph{Regression analysis of visual quality predictors.}
To identify which visual quality characteristics influence detection performance, we conducted ordinary least squares regressions with standardized coefficients and Benjamini--Hochberg false discovery rate (FDR) correction applied per regression to account for multiple comparisons across the 20 quality features (Table~\ref{tab:quality_regression}). Standard errors were determined by Breusch--Pagan heteroscedasticity tests, with HC3 robust standard errors used where heteroscedasticity was detected ($p < 0.05$) and standard errors otherwise.

\begin{table}[htbp]
\centering
\small
\caption{Quality factors predicting detection accuracy.}
\label{tab:quality_regression}
\scalebox{0.66}{
\begin{tabular}{lrrrrrrrrrr}
\toprule
 & \multicolumn{5}{c}{DF40} & \multicolumn{5}{c}{CharadesDF} \\
\cmidrule(lr){2-6} \cmidrule(lr){7-11} 
Quality Factor & Human & AI & Human Ens. & AI Ens. & Hybrid Ens. & Human & AI & Human Ens. & AI Ens. & Hybrid Ens. \\
\midrule
FR Confidence & 0.008* & 0.001 & 0.001 & 0.011 & 0.002 & 0.024* & -0.003 & 0.014 & -0.012 & 0.035* \\
Bbox Area (ratio) & 0.046*** & 0.034*** & 0.028 & 0.039*** & 0.017 & 0.016 & 0.026** & 0.000 & 0.038 & 0.003 \\
Face Presence (ratio) & -0.008 & 0.002 & -0.006 & -0.011* & -0.002 & 0.008 & 0.002 & 0.010 & -0.001 & 0.008 \\
IOD & -0.029 & -0.057*** & 0.019 & -0.029 & 0.003 & -0.022 & -0.062*** & 0.026 & -0.106** & -0.004 \\
Left EAR & -0.001 & -0.002 & -0.007 & 0.013 & 0.002 & 0.013 & 0.004 & 0.013 & 0.026 & 0.019 \\
Right EAR & -0.004 & -0.002 & 0.012 & -0.015 & -0.003 & 0.003 & -0.005 & 0.016 & -0.012 & 0.011 \\
MAR & -0.012 & 0.000 & -0.013 & 0.016 & 0.009 & 0.002 & -0.010 & -0.001 & -0.017 & -0.004 \\
SNR & 0.033* & 0.040*** & -0.006 & 0.004 & 0.011 & 0.004 & 0.040*** & -0.013 & 0.084** & 0.002 \\
Brightness & 0.000 & -0.012 & 0.003 & -0.004 & 0.000 & -0.002 & -0.026** & 0.000 & -0.043 & 0.000 \\
Sharpness & -0.021 & -0.001 & -0.009 & 0.016 & 0.012 & -0.021 & -0.003 & -0.008 & -0.001 & -0.001 \\
Contrast & 0.015 & 0.018** & -0.004 & 0.035 & 0.008 & 0.006 & 0.015 & -0.019 & 0.029 & -0.028 \\
Color Balance & 0.012 & 0.020*** & -0.010 & 0.014 & -0.002 & 0.007 & 0.018* & 0.003 & 0.058* & -0.007 \\
Blur (blurry) & -0.002 & -0.010 & -0.017 & -0.015 & -0.007 & -0.010 & -0.007 & -0.013 & -0.033 & -0.020 \\
Blur (hazy) & -0.006 & -0.001 & 0.008 & -0.017 & 0.006 & 0.006 & -0.005 & -0.011 & -0.013 & -0.021 \\
Pose (profile) & 0.011 & -0.005 & 0.002 & -0.010 & 0.001 & -0.006 & -0.003 & -0.015 & 0.002 & -0.012 \\
Pose (slight angle) & 0.002 & 0.002 & 0.012 & -0.020 & 0.012 & -0.006 & -0.001 & -0.017 & 0.003 & -0.028* \\
Occlusion & -0.004 & 0.005 & -0.022 & 0.001 & 0.005 & 0.002 & 0.001 & -0.002 & 0.007 & 0.008 \\
Exaggerated expression & 0.000 & 0.000 & 0.000 & 0.000 & 0.000 & -0.020* & 0.005 & -0.035** & 0.027*** & -0.032* \\
Extreme lightning & -0.018 & -0.001 & -0.015 & -0.003 & 0.005 & 0.000 & 0.002 & -0.010 & 0.013 & -0.012 \\
Quality Score & 0.015 & 0.019** & 0.015 & -0.001 & 0.009 & -0.011 & 0.016 & -0.014 & 0.050 & -0.001 \\
\midrule
$R^2$ & 0.099 & 0.137 & 0.037 & 0.048 & 0.019 & 0.059 & 0.093 & 0.041 & 0.051 & 0.061 \\
$N$ & 997 & 997 & 997 & 997 & 997 & 779 & 779 & 779 & 779 & 779 \\
\bottomrule
\end{tabular}}
\vspace{2mm}
\begin{minipage}{\textwidth}
\footnotesize
\textit{Notes:} Standardized OLS coefficients. 
Standard errors determined by Breusch--Pagan heteroscedasticity test: 
HC3 robust SEs used where heteroscedasticity detected ($p<0.05$), standard SEs otherwise. 
$p$-values corrected for multiple comparisons using Benjamini--Hochberg FDR (per regression). 
$^{*}p_{\mathrm{FDR}}<0.05$, $^{**}p_{\mathrm{FDR}}<0.01$, $^{***}p_{\mathrm{FDR}}<0.001$.
\end{minipage}
\end{table}

Face size emerged as the most robust predictor: bounding box area ratio was significantly associated with higher accuracy for humans ($\beta = 0.046$, $P_{\mathrm{FDR}} < 0.001$), AI detectors ($\beta = 0.034$, $P_{\mathrm{FDR}} < 0.001$), and the AI ensemble ($\beta = 0.039$, $P_{\mathrm{FDR}} < 0.001$) in DF40, with a consistent positive effect for AI detectors in CharadesDF ($\beta = 0.026$, $P_{\mathrm{FDR}} = 0.009$). The human ensemble effect in DF40 did not survive FDR correction ($\beta = 0.028$, $P_{\mathrm{FDR}} = 0.245$). Inter-ocular distance (IOD) showed a strong negative association with AI accuracy that survived FDR correction in both datasets (DF40: $\beta = -0.057$, $P_{\mathrm{FDR}} < 0.001$; CharadesDF: $\beta = -0.062$, $P_{\mathrm{FDR}} < 0.001$), with an especially large effect in the CharadesDF AI ensemble ($\beta = -0.106$, $P_{\mathrm{FDR}} = 0.002$), while showing no significant effects for humans after correction---suggesting that AI detectors are specifically susceptible to artifacts associated with face scaling. Signal-to-noise ratio positively predicted AI detection performance across both datasets after FDR correction (DF40: $\beta = 0.040$, $P_{\mathrm{FDR}} < 0.001$; CharadesDF: $\beta = 0.040$, $P_{\mathrm{FDR}} < 0.001$), with a particularly strong effect in the CharadesDF AI ensemble ($\beta = 0.084$, $P_{\mathrm{FDR}} = 0.008$); notably, an effect for humans in DF40 also reached significance ($\beta = 0.033$, $P_{\mathrm{FDR}} = 0.044$). Color balance showed significant positive associations with AI detection accuracy in both datasets after correction (DF40: $\beta = 0.020$, $P_{\mathrm{FDR}} < 0.001$; CharadesDF: $\beta = 0.018$, $P_{\mathrm{FDR}} = 0.020$) and in the CharadesDF AI ensemble ($\beta = 0.058$, $P_{\mathrm{FDR}} = 0.017$) but not human accuracy, indicating that AI detectors leverage low-level color statistics that humans do not consciously process. Contrast significantly predicted AI detection accuracy in DF40 ($\beta = 0.018$, $P_{\mathrm{FDR}} = 0.002$) but not in CharadesDF. The overall Quality Score---an aggregate measure of visual quality---significantly predicted AI detection accuracy in DF40 ($\beta = 0.019$, $P_{\mathrm{FDR}} = 0.001$), but showed no significant effect in CharadesDF ($P_{\mathrm{FDR}} = 0.145$), consistent with the pattern that AI detectors are more sensitive to overall visual quality than humans. Face recognition confidence positively predicted human accuracy in both datasets (DF40: $\beta = 0.008$, $P_{\mathrm{FDR}} = 0.030$; CharadesDF: $\beta = 0.024$, $P_{\mathrm{FDR}} = 0.039$) and hybrid ensemble accuracy in CharadesDF ($\beta = 0.035$, $P_{\mathrm{FDR}} = 0.013$). In the DF40 AI ensemble, face presence ratio was negatively associated with accuracy ($\beta = -0.011$, $P_{\mathrm{FDR}} = 0.042$). Notably, brightness negatively predicted AI detection accuracy in CharadesDF ($\beta = -0.026$, $P_{\mathrm{FDR}} = 0.009$), an effect not observed in DF40.

Exaggerated expression---defined by CLIB-FIQA~\citep{ou2024clib} as the probability that a face displays an atypical or non-neutral expression---emerged as a notable predictor specifically in the CharadesDF dataset, with divergent effects across detector types: it negatively predicted human accuracy ($\beta = -0.020$, $P_{\mathrm{FDR}} = 0.046$), human ensemble accuracy ($\beta = -0.035$, $P_{\mathrm{FDR}} = 0.004$), and hybrid ensemble accuracy ($\beta = -0.032$, $P_{\mathrm{FDR}} = 0.013$), while positively predicting AI ensemble accuracy ($\beta = 0.027$, $P_{\mathrm{FDR}} < 0.001$). This pattern suggests that exaggerated expressions may introduce visual complexity that impairs human judgment while providing exploitable patterns for AI detectors. Additionally, pose (slight angle) negatively predicted hybrid ensemble accuracy in CharadesDF ($\beta = -0.028$, $P_{\mathrm{FDR}} = 0.029$).

Several other factors showed raw significance that did not survive FDR correction, including sharpness for humans (DF40: $P_{\mathrm{FDR}} = 0.077$; CharadesDF: $P_{\mathrm{FDR}} = 0.133$), extreme lighting for humans in DF40 ($P_{\mathrm{FDR}} = 0.098$), and contrast for AI detectors in CharadesDF ($P_{\mathrm{FDR}} = 0.221$). The remaining quality features---including left and right eye aspect ratios, mouth aspect ratio, blur variables, most pose variables, and occlusion---did not reach significance after FDR correction in any model. The overall models explained modest variance ($R^2 = 0.02$--$0.14$), consistent with quality factors representing one of multiple determinants of detection success.

 \paragraph{Performance curves by quality feature.}
To complement the regression analysis, we examined how detection accuracy varies across the range of each quality factor using smoothed performance curves (Figure~\ref{fig:quality_curves}). Curves for all the remaining quality features are reported in Supplementary Material.

\begin{figure}[t]
    \centering
    \includegraphics[width=\textwidth]{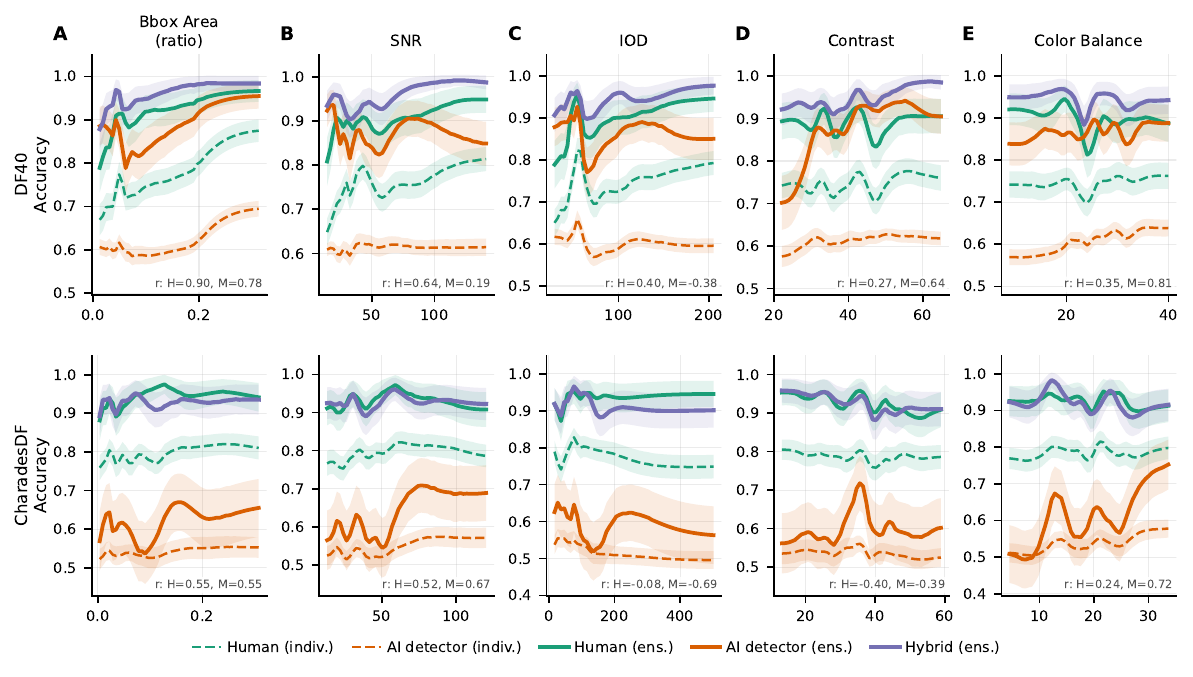}
    \caption{\textbf{Detection accuracy as a function of image quality factors.}
    Smoothed performance curves showing the relationship between quality features and detection accuracy for human individuals (teal, dashed), AI detectors (orange, dashed), human ensembles (teal, solid), AI ensembles (orange, solid), and hybrid ensembles (purple, solid). Shaded regions indicate 95\% confidence intervals. We indicate correlations ($\rho$) in each subplot for both humans ($H$) and AI detectors ($M$). Rows correspond to DF40 (top) and CharadesDF (bottom) datasets. 
    (\textbf{A})~Bounding box area ratio. 
    (\textbf{B})~Inter-ocular distance (IOD).
    (\textbf{C})~Signal-to-noise ratio (SNR).
    (\textbf{D})~Contrast.
    (\textbf{E})~Color balance.}
    \label{fig:quality_curves}
\end{figure}

Face size (bounding box area ratio) showed the strongest positive relationship with accuracy for both humans ($\rho = 0.90$) and AI detectors ($\rho = 0.78$) in DF40, with human accuracy increasing from approximately 67\% at small face sizes to 87\% at large face sizes, compared to a similar but weaker trend for AI detectors (61\% to 69\%). This pattern was attenuated but still present in CharadesDF ($\rho_{\text{human}} = 0.55$, $\rho_{\text{AI}} = 0.55$), with both showing parallel improvements across the feature range. Inter-ocular distance (IOD) exhibited a notably divergent pattern in DF40: human accuracy increased with IOD ($\rho = +0.40$) while AI detection accuracy decreased ($\rho = -0.38$), consistent with the regression finding that AI detectors are specifically susceptible to face-scaling artifacts. This divergence was even more pronounced in CharadesDF, where AI detectors showed a strong negative correlation with IOD ($\rho_{\text{AI}} = -0.69$) while humans showed near-zero correlation ($\rho_{\text{human}} = -0.08$), with AI detection accuracy declining from approximately 54\% at low IOD to 50\% at high IOD. Signal-to-noise ratio showed a strong positive correlation with human accuracy in DF40 ($\rho = 0.64$) but a much weaker relationship for AI detectors ($\rho = 0.19$); in CharadesDF, both showed positive relationships, though AI detectors exhibited a stronger correlation ($\rho = 0.67$) than humans ($\rho = 0.52$). Color balance was substantially more strongly associated with AI detection accuracy ($\rho = 0.81$ in DF40, $\rho = 0.72$ in CharadesDF) than human accuracy ($\rho = 0.35$ and $\rho = 0.24$, respectively), reinforcing the regression finding that AI detectors leverage low-level color statistics that humans do not consciously process. Contrast showed mixed patterns: in DF40, AI detectors exhibited a stronger positive relationship ($\rho = 0.64$) compared to humans ($\rho = 0.27$), while in CharadesDF both showed moderate negative correlations ($\rho_{\text{human}} = -0.40$, $\rho_{\text{AI}} = -0.39$). Across all features and datasets, the hybrid ensemble consistently achieved the highest accuracy, demonstrating that complementary insights from humans and AI detectors can be leveraged regardless of quality conditions.

 \subsection{Research Question 4: Confidence, Calibration, and Metacognition}
\textit{What is the relationship between confidence and accuracy in deepfake detection?}
\begin{takeaways}[RQ4]
    \begin{enumerate}
        \item \textbf{Humans have better insight into when they are wrong.} Human confidence is substantially higher on correct predictions than incorrect ones, indicating genuine metacognitive awareness. AI detectors show much weaker confidence discrimination, meaning their confidence scores are less informative about prediction-level accuracy.
        
        \item \textbf{Both humans and AI detectors show the same asymmetric calibration pattern.} Both groups are overconfident when predicting a video is real but underconfident when predicting a deepfake, despite comparable overall calibration error.
        
        \item \textbf{The Dunning-Kruger effect appears in both humans and AI detectors.} Poor performers overestimate their ability while top performers tend toward underconfidence. This pattern is consistent across datasets and is more pronounced in AI detectors than humans.
        
        \item \textbf{Self-reported confidence before the task does not predict actual performance.} Participants' pre-task beliefs about their deepfake detection ability bear no relationship to their actual accuracy.
    \end{enumerate}
\end{takeaways}
\paragraph{Confidence discrimination and calibration analysis.}
Participants and models rated each video on a 5-point scale from ``Definitely Authentic'' to ``Definitely a Deepfake,'' which we converted to a probability score $p \in [0, 1]$ where 1 indicates maximum confidence that the video is a deepfake. We operationalized \textit{confidence} as certainty in the binary prediction: $\text{confidence} = \max(p, 1-p)$, yielding values in $[0.5, 1.0]$. For example, ratings of $p = 0.9$ (likely deepfake) or $p = 0.1$ (likely real) both yield confidence $= 0.9$, reflecting high certainty regardless of direction; $p = 0.5$ corresponds to minimal confidence.

Human participants demonstrated strong \textit{confidence discrimination}: mean confidence was significantly higher on correct predictions than incorrect predictions (DF40: $\mu = 0.874$ vs.\ $\mu = 0.738$, $t = 29.89$, $p < 0.001$, Cohen's $d = 0.88$; CharadesDF: $\mu = 0.901$ vs.\ $\mu = 0.767$, $t = 26.64$, $p < 0.001$, $d = 0.92$), indicating genuine insight into performance on individual predictions. By contrast, AI detectors showed markedly weaker confidence discrimination (DF40: $\mu = 0.845$ vs.\ $\mu = 0.806$, $t = 36.15$, $p < 0.001$, $d = 0.24$; CharadesDF: $\mu = 0.815$ vs.\ $\mu = 0.798$, $t = 13.69$, $p < 0.001$, $d = 0.10$), suggesting that AI detectors' confidence scores carry less diagnostic information about prediction-level correctness.

\begin{figure}[t]
    \centering
    \includegraphics[width=0.9\textwidth]{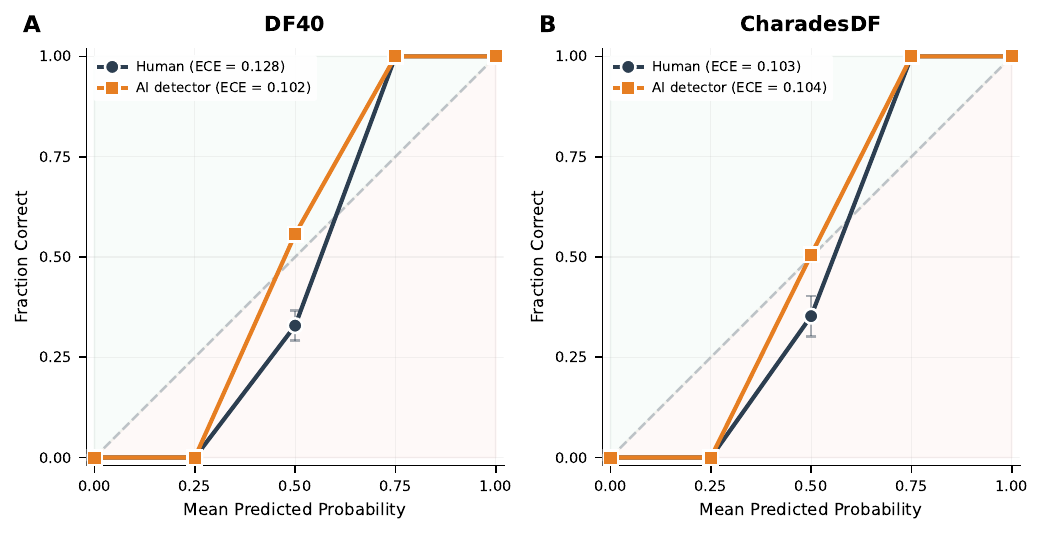}
    \caption{\textbf{Reliability diagrams for deepfake detection confidence: humans vs.\ AI detectors.}
    Calibration curves showing the relationship between mean predicted probability and fraction correct for human participants (dark curves) and AI detectors (orange curves). The dashed line indicates perfect calibration; points above the line indicate underconfidence, points below indicate overconfidence. Error bars show 95\% confidence intervals.
    (\textbf{A})~DF40,
    (\textbf{B})~CharadesDF.}
    \label{fig:calibration_curves}
\end{figure}

We next examined \textit{calibration}, i.e.\ whether confidence magnitude corresponds to actual accuracy. Calibration determines whether confidence can serve as a reliable signal for downstream decisions, such as a content moderator prioritizing low-confidence flags or a hybrid system weighting assessments by stated certainty. A well-calibrated observer reporting 80\% confidence should be correct 80\% of the time. We constructed reliability diagrams by binning trials by predicted probability and computing accuracy within each bin (Figure~\ref{fig:calibration_curves}), quantifying miscalibration using Expected Calibration Error (ECE $= \sum_b \frac{n_b}{N} |\text{accuracy}_b - \text{confidence}_b|$, where $n_b$ is trials in bin $b$, \citep{DBLP:conf/aaai/NaeiniCH15}) and Brier score (mean squared error between predictions and outcomes, \citep{glenn1950verification}). 

Humans and AI detectors showed comparable overall miscalibration (DF40: human ECE $= 0.128$, Brier $= 0.053$; AI detectors ECE $= 0.102$, Brier $= 0.056$; CharadesDF: human ECE $= 0.103$, Brier $= 0.041$; AI detectors ECE $= 0.104$, Brier $= 0.065$), with a shared asymmetric pattern: both groups were overconfident when predicting a video was real (accuracy below the diagonal) but underconfident when predicting a deepfake (accuracy above the diagonal). 

\begin{figure}[t]
    \centering
    \includegraphics[width=0.9\textwidth]{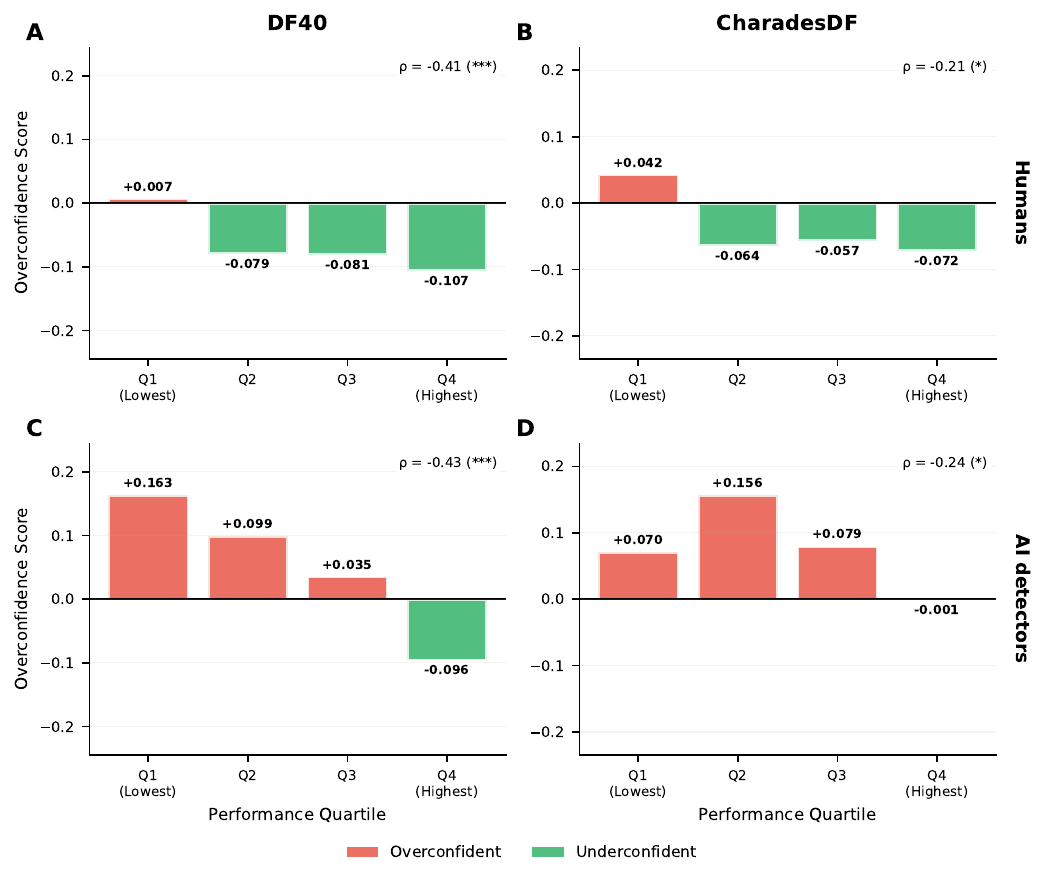}
    \caption{\textbf{Overconfidence score by performance quartile: humans vs.\ AI detectors.}
    Overconfidence computed as normalized confidence minus accuracy; positive values (red) indicate overconfidence, negative values (green) indicate underconfidence. Top row: human participants; bottom row: AI detectors. Poor performers (Q1) are overconfident while top performers (Q4) trend toward underconfidence in both groups, consistent with the Dunning-Kruger effect. The effect is more pronounced for AI detectors.
    (\textbf{A})~DF40 humans: $\rho = -0.405$, $p < 0.001$.
    (\textbf{B})~CharadesDF humans: $\rho = -0.210$, $p = 0.036$.
    (\textbf{C})~DF40 AI detectors: $\rho = -0.431$, $p < 0.001$.
    (\textbf{D})~CharadesDF AI detectors: $\rho = -0.238$, $p = 0.020$.}
    \label{fig:overconfidence}
\end{figure}

\paragraph{Overconfidence by performance quartile.}
We tested whether poor performers overestimate their ability relative to top performers. We computed an \textit{overconfidence score} for each participant or model: $\text{overconfidence} = \frac{\text{confidence} - 0.5}{0.5} - \text{accuracy}$, where the normalization maps confidence from $[0.5, 1.0]$ to $[0, 1]$ for direct comparison with accuracy. Positive values indicate overconfidence; negative values indicate underconfidence. Participants and models were grouped into quartiles by accuracy (Figure~\ref{fig:overconfidence}).

Among humans in DF40, the bottom quartile (accuracy: 0.30--0.68, $n = 25$) showed slight overconfidence ($+0.007$), while the top quartile (accuracy: 0.81--0.98, $n = 25$) was underconfident ($-0.107$). The negative correlation between accuracy and overconfidence was highly significant (Spearman $\rho = -0.405$, $p < 0.001$), with the bottom quartile significantly more overconfident than the top (Mann-Whitney $U = 498.5$, $p < 0.001$). This pattern replicated in CharadesDF: bottom quartile overconfidence was $+0.042$ versus $-0.072$ for the top quartile ($\rho = -0.210$, $p = 0.036$; $U = 399$, $p = 0.023$).
The same Dunning-Kruger pattern emerged among AI detectors. In DF40, the worst-performing models showed pronounced overconfidence ($+0.163$) while top models were underconfident ($-0.096$; $\rho = -0.431$, $p < 0.001$; $U = 501$, $p < 0.001$). In CharadesDF, the bottom-quartile AI detectors exhibited overconfidence ($+0.070$) compared to near-zero overconfidence in the top quartile ($-0.001$; $\rho = -0.238$, $p = 0.020$; $U = 317$, $p = 0.124$). Notably, the magnitude of the Dunning-Kruger effect was larger for AI detectors than humans in DF40: the Q1--Q4 overconfidence gap was $+0.259$ for AI detectors versus $+0.113$ for humans. In CharadesDF, this pattern reversed, with a gap of $+0.071$ for AI detectors versus $+0.114$ for humans.
Self-reported confidence measured before the task (Q9: ``How confident are you in your ability to detect deepfakes?'') did not predict actual accuracy for human participants (DF40: $\rho = -0.015$, $p = 0.880$; CharadesDF: $\rho = 0.062$, $p = 0.538$).

\subsection{Research Question 5: Demographic Predictors of Detection Accuracy}
\textit{Do demographic characteristics predict individual differences in deepfake detection performance?}

\begin{takeaways}[RQ5]
    \begin{enumerate}
        \item \textbf{Demographic factors explain little variance in detection accuracy.} Most demographic variables—including age, gender, tech savviness, social media usage, and deepfake familiarity—do not significantly predict performance.
        
        \item \textbf{Education shows a counterintuitive negative effect in one dataset.} Higher education was associated with lower accuracy in CharadesDF, though this effect did not replicate in DF40.
        
        \item \textbf{Self-reported deepfake experience does not translate to better detection.} Neither familiarity with deepfakes nor prior exposure to deepfake content predicted improved accuracy.
        
        \item \textbf{Self-assessed confidence does not predict actual performance.} Participants' beliefs about their detection ability bore no relationship to their accuracy, consistent with the Dunning-Kruger findings in RQ4.
        
        \item \textbf{Gender differences are absent.} Male and female participants performed comparably, with no significant differences in either dataset.
    \end{enumerate}
\end{takeaways}

\paragraph{Regression analysis of demographic predictors.}
To examine whether individual differences in demographic characteristics predict deepfake detection performance, we conducted ordinary least squares regressions with participant-level accuracy as the dependent variable. Ordinal demographics (age, education, tech savviness, social media usage, deepfake familiarity, prior deepfake exposure, and self-reported confidence) were treated as continuous predictors; gender was dummy-encoded with male as the reference category. Coefficients were standardized, and Benjamini--Hochberg FDR correction was applied to account for multiple comparisons across the eight demographic predictors (Table~\ref{tab:demographic_regression}). Standard errors were determined by Breusch--Pagan heteroscedasticity tests, with standard errors used in both datasets as no significant heteroscedasticity was detected (DF40: $p = 0.401$; CharadesDF: $p = 0.347$).

\begin{table}[t]
\centering
\small
\caption{Demographic factors predicting human deepfake detection accuracy.}
\label{tab:demographic_regression}
\begin{tabular}{lrr}
\toprule
Demographic Factor & DF40 & CharadesDF \\
\midrule
Age & 0.021 & $-$0.025 \\
Education & 0.032 & $-$0.043** \\
Tech Savviness & $-$0.023 & 0.001 \\
Social Media Usage & 0.012 & 0.002 \\
Deepfake Familiarity & $-$0.004 & 0.013 \\
Prior Deepfake Exposure & 0.009 & 0.020 \\
Self-Reported Confidence & 0.002 & $-$0.016 \\
Female (vs.\ Male) & $-$0.021 & 0.011 \\
\midrule
$R^2$ & 0.112 & 0.216 \\
$N$ & 100 & 100 \\
\bottomrule
\end{tabular}
\vspace{2mm}
\begin{minipage}{\textwidth}
\footnotesize
\textit{Notes:} Standardized OLS coefficients. 
Ordinal demographics encoded as continuous; gender dummy-encoded (reference: Male). 
$p$-values corrected using Benjamini--Hochberg FDR.
$^{*}p_{\mathrm{FDR}}<0.05$, $^{**}p_{\mathrm{FDR}}<0.01$, $^{***}p_{\mathrm{FDR}}<0.001$.
\end{minipage}
\end{table}

The overall models explained modest variance, with low adjusted $R^2$ values indicating that demographic factors account for little of the individual variation in detection performance (DF40: Adj.\ $R^2 = 0.024$; CharadesDF: Adj.\ $R^2 = 0.147$). Only one predictor survived FDR correction: education was negatively associated with accuracy in CharadesDF ($\beta = -0.043$, $P_{\mathrm{FDR}} < 0.01$), a counterintuitive finding suggesting that more educated participants performed slightly worse. This effect did not replicate in DF40 ($\beta = 0.032$, $P_{\mathrm{FDR}} > 0.05$), and the inconsistency across datasets suggests this may be a chance finding or reflect dataset-specific characteristics rather than a robust relationship.

Age showed no significant association with accuracy in either dataset (DF40: $\beta = 0.021$; CharadesDF: $\beta = -0.025$; both $P_{\mathrm{FDR}} > 0.05$), contrary to common assumptions that younger, more digitally native individuals might be better at detecting synthetic media. Similarly, tech savviness (DF40: $\beta = -0.023$; CharadesDF: $\beta = 0.001$) and social media usage frequency (DF40: $\beta = 0.012$; CharadesDF: $\beta = 0.002$) were not significant predictors, indicating that general technological proficiency and exposure to social media platforms do not confer detection advantages.

Notably, neither deepfake familiarity ($\beta = -0.004$ and $0.013$) nor prior deepfake exposure ($\beta = 0.009$ and $0.020$) predicted accuracy in either dataset. This finding suggests that self-reported knowledge about deepfakes and experience viewing them does not translate into superior detection ability—potentially because awareness of the phenomenon does not provide the specific perceptual skills needed for accurate discrimination. Self-reported confidence in detection ability was also not significant ($\beta = 0.002$ and $-0.016$), consistent with our RQ4 finding that pre-task confidence does not predict actual performance.

Gender differences were absent: male and female participants performed comparably in both datasets (DF40: $\beta = -0.021$; CharadesDF: $\beta = 0.011$; both $P_{\mathrm{FDR}} > 0.05$). The overall pattern of null findings across demographic variables suggests that deepfake detection ability may depend more on factors not captured by standard demographics—such as individual perceptual sensitivity, attention to detail, or specific training—rather than on age, education, gender, or self-reported digital media experience.

\section{Discussion}

This paper provides comprehensive evidence from two complementary datasets and large-scale experiments with 200 human participants (100 assigned to DF40 and 100 to CharadesDF) and 95 AI detectors that humans currently outperform state-of-the-art deepfake detection algorithms, particularly under challenging real-world conditions. Our findings challenge the prevailing narrative in the deepfake detection literature that positions machine learning as the primary solution to synthetic media threats~\citep{DBLP:conf/iccv/RosslerCVRTN19,korshunov2021subjective,DBLP:journals/corr/abs-2307-01426} while revealing important opportunities for human-AI collaboration~\citep{groh2021deepfake,ibsen2024conditional,boyd2022aiguidance,josephs2023artifactmag}.

\paragraph{Humans as capable deepfake detectors.}

Across both controlled (DF40) and naturalistic (CharadesDF) datasets, human participants achieved significantly higher detection accuracy than AI detectors. This advantage was especially pronounced on CharadesDF, where AI detectors performed near chance level ($\mu =$ 0.537) while humans maintained robust accuracy ($\mu =$ 0.784). The divergence between datasets reflects the vulnerability to distribution shift of current machine learning approaches, consistent with prior work showing that detectors trained on curated benchmarks often fail under new generators, compression pipelines, or recording conditions~\citep{DBLP:conf/iccv/RosslerCVRTN19,korshunov2021subjective,prasad2022noisychannels,xu2025chasingshadows,hasan2026uneven,kamali2025diffusion}. AI detectors, trained on carefully curated datasets with frontal poses and consistent lighting, struggled when confronted with the variability characteristic of authentic user-generated content (e.g., the variable camera angles, partial occlusions, varied lighting conditions in CharadesDF)~\citep{josephs2023artifactmag,cooke2024coin,frank2024representative}. Humans, by contrast, appeared to leverage more generalizable perceptual and cognitive strategies that transferred across contexts, in line with studies showing human advantages on some dynamic or ecologically rich deepfake tasks~\citep{groh2021deepfake,goh2024humansvsmachines,pehlivanoglu2026susceptibility,mueller2021audiodeepfake}, even though other work has emphasized that humans detection ability can be near chance in more constrained or highly realistic regimes~\citep{kobis2021fooled,bray2022deepfakefaces,cooke2024coin,frank2024representative,somoray2025review}.

\paragraph{The power of ensemble methods.}

Our results demonstrate that aggregating multiple assessments substantially improves detection performance, with human ensembles achieving accuracy gains of 14--15 percentage points over individual humans. More striking was the dramatic reduction in high confidence errors (i.e., cases where the absolute difference between ground truth and prediction exceeds 0.7): while individual judges were confidently wrong in 17--32\% of cases, ensembles reduced this to under 2\%, with hybrid ensembles eliminating such errors entirely. This finding has immediate practical implications for content moderation systems, where confident misclassifications carry significant costs. It also echoes prior work showing that aggregating multiple human assessments can rival or exceed individual performance in deepfake detection~\citep{groh2021deepfake,somoray2025review} and that ensembles of human and algorithmic decisions can outperform either alone in related manipulation-detection tasks~\citep{ibsen2024conditional,boyd2022aiguidance,josephs2023artifactmag}.

When we examined high-confidence errors, humans and AI detectors almost never failed on the same videos. Humans predominantly misclassified high-quality deepfakes as real, suggesting vulnerability to sophisticated face manipulations that successfully mimicked natural appearance~\citep{kobis2021fooled,pehlivanoglu2026susceptibility}. AI detectors exhibited the opposite pattern, predominantly misclassifying real videos as deepfakes, potentially due to over-reliance on compression artifacts or low-level regularities present in both authentic and manipulated content~\citep{DBLP:conf/iccv/RosslerCVRTN19,prasad2022noisychannels,xu2025chasingshadows}. By combining predictions from both groups, hybrid ensembles leveraged this complementarity. This finding aligns with broader research on human--AI complementarity and collective intelligence in synthetic media detection~\citep{groh2021deepfake,mueller2021audiodeepfake,boyd2022aiguidance,ibsen2024conditional,casu2025automationbias}, while extending these principles to the specific domain of synthetic media detection at scale.

\paragraph{Quality factors and detection strategies.}

Our analysis of quality factors revealed that humans and AI detectors employ qualitatively different detection strategies. Face size emerged as the most robust predictor for both groups. Larger, more visible faces substantially improved accuracy, consistent with prior findings that visibility and pose strongly modulate human and AI performance on facial manipulations~\citep{DBLP:conf/iccv/RosslerCVRTN19,woehler2021perceptual,bray2022deepfakefaces,47}. But beyond this shared reliance on basic visibility, the groups diverged. AI detectors were significantly more sensitive to low-level visual properties: signal-to-noise ratio, color balance, and contrast all predicted AI but not human accuracy. This pattern suggests AI detectors have learned to exploit statistical regularities in image quality and compression that may not generalize across contexts~\citep{prasad2022noisychannels,xu2025chasingshadows,kamali2025diffusion}, while humans rely on higher-level perceptual and semantic cues such as motion coherence, expression dynamics, and content plausibility~\citep{woehler2021perceptual,41,49}.

These findings have implications for both attack and defense strategies. Adversaries seeking to evade AI detection might focus on matching the low-level statistical properties of authentic content, as suggested by work on benchmark shortcut learning and mirage models~\citep{xu2025chasingshadows,hasan2026uneven,kamali2025diffusion}, while those targeting human judges might invest in high-quality face manipulations that preserve natural appearance and plausible motion~\citep{kobis2021fooled,pehlivanoglu2026susceptibility}. Conversely, detection systems might be improved by explicitly accounting for these differential vulnerabilities, perhaps through ensemble approaches that weight human and AI detectors contributions based on content characteristics such as pose, noise, and scene complexity~\citep{groh2021deepfake,ibsen2024conditional,josephs2023artifactmag}.

\paragraph{Confidence calibration and metacognition.}

Our examination of confidence and accuracy revealed that humans possess genuine metacognitive insight into their detection performance. They were substantially more confident on correct predictions than incorrect ones, aligning with prior work showing that confidence carries useful information about correctness even when absolute calibration is imperfect~\citep{kobis2021fooled,janssen2024conflict,chen2025selfefficacy,somoray2025review}. AI detectors showed much weaker confidence discrimination, meaning their confidence scores carried less diagnostic information about prediction-level correctness, echoing broader concerns that detector scores can be poorly calibrated under distribution shift~\citep{hasan2026uneven,xu2025chasingshadows}.

Both groups exhibited a shared asymmetric calibration pattern: overconfidence when predicting a video was real, underconfidence when predicting a deepfake. This asymmetry may reflect prior expectations about base rates of authentic versus fabricated content, or differential difficulty in the two classification directions. Regardless of mechanism, this pattern suggests that both human and AI detectors' decision thresholds might benefit from recalibration, potentially shifting toward greater skepticism about authenticity in high-stakes settings.

The emergence of the Dunning--Kruger effect in both humans and AI detectors was an unexpected finding. Poor performers systematically overestimated their ability relative to top performers, a pattern that has been documented for humans in related AI-media tasks~\citep{miller2023aihyperreal,cooke2024coin,somoray2025review} and here appears more pronounced in AI detectors than humans. For AI detectors, this phenomenon likely reflects training dynamics that produce overconfident predictions on out-of-distribution samples~\citep{hasan2026uneven,xu2025chasingshadows}. For humans, it suggests that deepfake detection skill may be difficult to introspect—those who lack it may not recognize their deficiency, consistent with prior findings of overconfidence and miscalibration in deepfake detection~\citep{kobis2021fooled,chen2025selfefficacy,casu2025automationbias}.

\paragraph{The limited role of demographics. }

Contrary to common assumptions, demographic factors explained little variance in human detection accuracy. Age, gender, self-reported tech savviness, social media usage frequency, and even self-reported familiarity with deepfakes did not significantly predict performance. This is broadly consistent with prior work finding only small-to-moderate effects of age, experience, and literacy on detection ability~\citep{frank2024representative,mueller2021audiodeepfake,alsobeh2024unmasking,xie2025seeing}, and with evidence that more specialized traits (e.g., analytic thinking, visual literacy, or face-recognition aptitude) are more predictive than coarse demographics~\citep{pehlivanoglu2026susceptibility,jin2025visual,davis2025superrecogniser,somoray2025review}. 

These findings have implications for training and intervention design. If standard demographic characteristics and self-reported experience do not predict detection ability, interventions may need to focus on developing specific perceptual and cognitive skills rather than simply raising awareness~\citep{tahir2021seeing}. The disconnect between self-reported confidence and actual performance further underscores that people lack insight into their own detection capabilities, suggesting that feedback-based, performance-contingent training approaches might be particularly valuable~\citep{kobis2021fooled,chen2025selfefficacy,somoray2025review}.

\subsection{Limitations and future directions}

Several limitations warrant consideration. First, while we evaluated 95 AI detectors variants trained on three datasets, the machine learning landscape changes rapidly, and newer architectures or training strategies might narrow the human-AI performance gap. However, the fundamental vulnerability to distribution shift that we document is unlikely to be resolved by architectural improvements alone, echoing concerns raised by recent analyses of benchmark shortcut learning and cross-generator failures~\citep{xu2025chasingshadows,hasan2026uneven,kamali2025diffusion}.

Second, our human participants evaluated videos in an experimental context that explicitly primed attention to authenticity. In naturalistic social media environments, where attention is not focused on accuracy, sharing behavior may diverge from discernment ability~\citep{cooke2024coin,frank2024representative,sanchezacedo2024mil}. Future work should examine how these findings translate to ecological settings where people are browsing, not explicitly evaluating, and where deepfakes are rare, viewing time is limited, and attention is divided~\citep{josephs2024browsing,cooke2024coin}.

Third, our deepfakes were generated using specific techniques represented in the DF40 benchmark and our novel CharadesDF dataset. Hyper-realistic deepfakes produced by professional visual effects artists, or future generations of generative models (e.g., diffusion and NeRF-based systems), may present different challenges for both human and AI detectors~\citep{hasan2026uneven,kamali2025diffusion}. Similarly, we focused on video deepfakes; audio-only or text-based synthetic content may exhibit different patterns of human-AI complementarity, as suggested by work on audio deepfakes, audiovisual face-voice manipulations, and AI-generated text and images~\citep{mueller2021audiodeepfake,hashmi2024audiovisual,frank2024representative,cooke2024coin}.

Future research might productively explore several directions suggested by our findings. First, understanding what specific perceptual or cognitive strategies successful human detectors employ could inform both training interventions and machine learning architectures~\citep{tahir2021seeing,woehler2021perceptual,49,neo2025strategies}. Second, developing content-aware ensemble methods that dynamically weight human and AI detectors' contributions based on video characteristics might maximize the complementarity we documented~\citep{groh2021deepfake,ibsen2024conditional,boyd2022aiguidance,josephs2023artifactmag,casu2025automationbias}. Third, examining how detection performance changes with video duration, context, base rates, and prior expectations could inform practical deployment strategies~\citep{josephs2024browsing,janssen2024conflict,cooke2024coin,xie2025seeing}. Finally, as generative technologies continue to advance, longitudinal studies tracking the evolution of human and AI-based detection capabilities will be essential for maintaining effective defenses against synthetic media threats~\citep{hasan2026uneven,somoray2025review}.

\subsection{Ethical and societal impact}

This research was conducted to advance scientific understanding of human and AI capabilities in detecting synthetic media, with the goal of informing more effective defenses against potential misuse of deepfake technology~\citep{kobis2021fooled,appel2022political,sanchezacedo2024mil}. Our findings suggest that dismissing human judgment in favor of purely automated detection systems may be premature, and that human-in-the-loop approaches may offer the most robust path forward~\citep{groh2021deepfake,ibsen2024conditional,somoray2025review}.

The demonstration that humans outperform AI detectors on challenging, naturalistic content has direct implications for content moderation practices. Platforms might consider integrating human judgment into detection pipelines rather than relying solely on automated systems, particularly for high-stakes decisions where confident misclassifications carry significant costs~\citep{groh2021deepfake,ibsen2024conditional,sanchezacedo2024mil}. Our finding that hybrid ensembles eliminate catastrophic failures suggests a specific architectural approach: use AI detectors for initial screening, but route uncertain or high-stakes cases to human review, consistent with proposals in security and identity verification domains~\citep{ibsen2024conditional,davis2025superrecogniser,55}.

At the same time, several aspects of our findings warrant caution. The documentation of systematic calibration errors in both humans and AI detectors suggests that neither should be trusted unconditionally~\citep{kobis2021fooled,miller2023aihyperreal,chen2025selfefficacy,casu2025automationbias}. The Dunning--Kruger effect among human detectors implies that confidence should not be taken as a reliable indicator of accuracy~\citep{miller2023aihyperreal,cooke2024coin,somoray2025review}. And the limited role of demographics in predicting performance suggests that simply selecting for particular background characteristics will not guarantee detection competence~\citep{frank2024representative,alsobeh2024unmasking,xie2025seeing}.

We note potential dual-use concerns with detailed findings about what makes deepfakes detectable. Understanding that AI detectors rely on low-level statistical properties could inform adversarial strategies to evade detection~\citep{xu2025chasingshadows,hasan2026uneven,kamali2025diffusion}. We have attempted to balance scientific transparency with responsibility by reporting findings at a level that informs defensive applications without providing a detailed roadmap for evasion. We encourage future work in this area to similarly consider the implications of detailed vulnerability disclosure~\citep{sanchezacedo2024mil,50}.

Finally, our research underscores that deepfake detection is not purely a technical problem. Human perception, cognition, and social context all play important roles in determining how synthetic media is perceived and acted upon~\citep{kobis2021fooled,appel2022political,preu2022perception,cooke2024coin,frank2024representative}. Effective responses will likely require not only improved detection technologies but also media literacy education, platform design interventions, and policy frameworks that account for the fundamentally sociotechnical nature of the challenge~\citep{sanchezacedo2024mil,21,49,50}.

\section{Methods}

\subsection{Consent and ethics}

This research complied with all relevant ethical regulations and was approved by Northwestern University's Institutional Review Board (IRB) as a minimal-risk study under the title ``Understanding the Impact of Face Clarity and Scene Dynamism on Deepfake Generation Quality and Human Detection Performance'' (IRB ID: STU00224530).  The study comprised two phases, each governed by separate consent procedures.

In Phase~1 (video generation for CharadesDF), participants were presented with an informed consent statement describing the study's purpose, the requirement to record videos of themselves, and the potential public release of the resulting dataset for non-commercial academic use. Participants acknowledged the inherent privacy risk that their facial images could be identifying despite the removal of direct identifiers (names, email addresses). In Phase~2 (detection evaluation), participants in both the DF40 and CharadesDF studies were presented with a separate informed consent statement explaining that they would view and evaluate videos containing both authentic recordings and synthetic deepfake versions. Both consent forms described the principal investigator, potential benefits and risks, data handling procedures, and participants' right to withdraw at any time without penalty. Consent was obtained via checkbox agreement on Qualtrics; a waiver of physical signature was granted by the IRB given the online nature of the study. Participants were provided a link to download a PDF copy of their consent form for their records.

All participants were compensated at a rate of \$10 for approximately one hour of participation. Compensation was contingent on completing the study. Participants found to be ineligible after starting were not compensated, and repeat participation was not permitted.

\subsection{Digital experiment interface}

Both detection experiments (Phase~2) were hosted on Qualtrics. Participants first completed a demographic and background questionnaire, then viewed and evaluated 60 videos presented in randomized order. Each video lasted around 30 seconds. For each video, participants were asked to rate their confidence that the video was authentic using a 5-point Likert scale (1~=~``Definitely a Deepfake,'' 2~=~``Likely a Deepfake,'' 3~=~``Unsure,'' 4~=~``Likely Authentic,'' 5~=~``Definitely Authentic''). We converted these ordinal responses to continuous probability scores $p \in [0, 1]$, where $p = 1$ indicates maximum confidence that the video is a deepfake. Specifically, Likert responses of 1 through 5 were mapped to probability scores of 1.0, 0.75, 0.5, 0.25, and 0.0, respectively. 

\subsection{Experiment stimuli}

\paragraph{DF40 dataset.}
The DF40 dataset~\cite{DBLP:conf/nips/YanYCZFZLWDWY24} is a large-scale, comprehensive benchmark for deepfake detection that substantially extends prior datasets in both diversity and scale. From its held-out test set, we sampled 500 fake videos uniformly at random across 10 forgery techniques spanning two categories of face manipulation: face-swapping methods (FSGAN~\cite{DBLP:conf/iccv/NirkinKH19}, FaceSwap~\cite{kowalski_faceswap_github}, SimSwap~\cite{DBLP:journals/corr/abs-2106-06340}, InSwapper~\cite{wang_inswapper_github}, BlendFace~\cite{DBLP:conf/iccv/ShioharaYT23}, UniFace~\cite{DBLP:conf/eccv/XuZHTZTWW022}, MobileFaceSwap~\cite{DBLP:conf/aaai/XuHDZHLD22}) and face-reenactment methods (Wav2Lip~\cite{DBLP:conf/mm/PrajwalMNJ20}, HyperReenact~\cite{DBLP:conf/iccv/BounareliTAPT23}, SadTalker~\cite{DBLP:conf/cvpr/ZhangCWZSGSW23}). These 500 fake videos were balanced with 500 real videos sourced from the original YouTube content in FaceForensics++~\cite{DBLP:conf/iccv/RosslerCVRTN19} (c23 compression) and Celeb-DF~\cite{li2020celeb}, yielding 1{,}000 evaluation stimuli. The DF40 benchmark was designed to address limitations of prior datasets that relied on outdated forgery techniques or limited manipulation diversity. The dataset features subjects in predominantly frontal poses with consistently visible, clearly resolved faces, representing controlled conditions suitable for systematic evaluation of deepfake detection methods.

\paragraph{CharadesDF dataset.}
The CharadesDF dataset was developed through Phase 1 of our IRB-approved study to simulate more challenging, real-world detection conditions. We recruited 100 participants via the Prolific platform, each of whom recorded five videos lasting around 15–30 seconds following specific scenario instructions designed to systematically vary face clarity and scene dynamism. All the instructions were randomly sourced from the Charades dataset \cite{DBLP:conf/eccv/SigurdssonVWFLG16}, a collection of 9,848 crowdsourced videos depicting casual everyday activities performed by 267 individuals in their own homes, annotated with action labels, temporal intervals, and object classes. Scenarios included activities such as speaking directly to the camera, drinking from a mug, moving around a room while speaking, and presenting in front of a microphone. In our study, each participant also provided one high-quality reference photograph of themselves with clear frontal lighting, which served as the source face for controlled face-swapping experiments.

% \footnote{We replicated the Charades protocol in a separate, independent study rather than reusing their original videos because participants in Charades did not consent to AI-based manipulation of their likenesses or distribution of synthetic media derived from their recordings—requirements mandated by our IRB approval.} 

Using this authentic video dataset of 500 recordings and reference faces, we generated corresponding deepfakes using FaceFusion (v3.2.0)\footnote{\url{https://facefusion.io}}, a publicly available face-swapping tool that visually imposes the face of one subject onto the video of another. We selected FaceFusion to emulate the realistic scenario in which individuals without technical backgrounds use publicly available graphical interfaces to create deepfakes without coding skills. To capture a diverse range of synthetic artifacts, we employed five face-swapping models from FaceFusion's available options: \texttt{face\_swapper\_hyperswap\_1a\_256} (the default), \texttt{inswapper\_128\_fp16} (the fastest), \texttt{hififace\_unofficial\_256} (the highest quality), \texttt{inswapper\_128} (a popular user choice), and \texttt{blend\_swap} (an older, lower-quality model). These models were chosen to represent variation in popularity, speed, output quality, and software evolution. Each model was applied to 100 videos, yielding 500 deepfakes. Combined with the 500 authentic recordings, this produced 1,000 evaluation stimuli. The CharadesDF dataset exhibits substantially greater variability than DF40 in image quality, camera angles, and face visibility, including partial occlusions and profile views, reflecting the diversity of conditions encountered in real-world video content.

\subsection{Participants}

\paragraph{Phase~1: Video generation (CharadesDF).}
In Phase~1, we recruited 100 participants from the United States via the Prolific platform. Eligibility criteria required participants to be at least 18 years of age, reside in the United States, have access to a device capable of recording video, and consent to the potential public release of their recordings for non-commercial academic use. Each participant recorded five videos and provided one reference photograph, yielding 500 authentic videos. Participants were compensated \$10 for approximately one hour of effort.

\paragraph{Phase~2: Detection evaluation.}
We conducted two independent detection studies---one using stimuli from DF40 and one using stimuli from CharadesDF---each with 100 participants recruited via Prolific. Eligibility criteria required participants to be at least 18 years of age, reside in the United States, and have access to a device capable of displaying video with adequate screen resolution and audio. Participants who took part in Phase~1 were excluded from Phase~2 to prevent familiarity bias. Due to practical constraints on participant time, each participant viewed a randomly sampled subset of 60 videos from the corresponding 1{,}000-video dataset such that each video was evaluated by multiple independent raters. Participants were compensated \$10 for approximately one hour of effort.

Prior to viewing the videos, participants completed a background questionnaire that collected demographic information including age, gender, education level, self-reported technology savviness, social media usage frequency, familiarity with deepfake technology, prior exposure to deepfake content, and self-reported confidence in their ability to detect deepfakes.

Table~\ref{tab:demographics_descriptives} presents the demographic composition of human participants across both datasets. Participants were predominantly aged 25--54 years (DF40: 80\%; CharadesDF: 80\%), with relatively few participants under 25 or over 65. Gender distribution was approximately balanced (DF40: 47\% male, 52\% female; CharadesDF: 54\% male, 46\% female), with no non-binary participants in either sample. Educational attainment was high: the majority held at least a bachelor's degree (DF40: 59\%; CharadesDF: 56\%). Most participants rated themselves as average or above average in tech savviness (DF40: 98\%; CharadesDF: 98\%) and reported daily or more frequent social media usage (DF40: 88\%; CharadesDF: 80\%). Participants generally reported high familiarity with deepfakes (very or extremely familiar---DF40: 76\%; CharadesDF: 61\%) and prior exposure to deepfake content (probably or definitely yes---DF40: 86\%; CharadesDF: 76\%). Self-reported confidence in detection ability was moderate, with most participants indicating they were moderately or very confident (DF40: 70\%; CharadesDF: 60\%). The demographic similarity across datasets supports the comparability of findings, while the overall sample characteristics suggest a relatively tech-savvy, educated population with substantial self-reported awareness of deepfakes.

\begin{table}[htbp]
\centering
\small
\caption{Participant demographics summary.}
\label{tab:demographics_descriptives}
\begin{tabular}{llrr}
\toprule
Variable & Category & DF40 & CharadesDF \\
\midrule
Age & Under 18 & 0 (0.0\%) & 0 (0.0\%) \\
 & 18--24 years old & 5 (5.0\%) & 2 (2.0\%) \\
 & 25--34 years old & 38 (38.0\%) & 33 (33.0\%) \\
 & 35--44 years old & 28 (28.0\%) & 23 (23.0\%) \\
 & 45--54 years old & 14 (14.0\%) & 24 (24.0\%) \\
 & 55--64 years old & 10 (10.0\%) & 12 (12.0\%) \\
 & 65+ years old & 5 (5.0\%) & 6 (6.0\%) \\
\addlinespace
Gender & Male & 47 (47.0\%) & 54 (54.0\%) \\
 & Female & 52 (52.0\%) & 46 (46.0\%) \\
 & Non-binary / third gender & 0 (0.0\%) & 0 (0.0\%) \\
\addlinespace
Education & Some high school or less & 1 (1.0\%) & 0 (0.0\%) \\
 & High school diploma or GED & 16 (16.0\%) & 15 (15.0\%) \\
 & Some college, but no degree & 14 (14.0\%) & 21 (21.0\%) \\
 & Associates or technical degree & 9 (9.0\%) & 8 (8.0\%) \\
 & Bachelor's degree & 38 (38.0\%) & 32 (32.0\%) \\
 & Graduate or professional degree & 21 (21.0\%) & 24 (24.0\%) \\
\addlinespace
Tech Savviness & Beginner & 0 (0.0\%) & 0 (0.0\%) \\
 & Below average & 2 (2.0\%) & 2 (2.0\%) \\
 & Average & 32 (32.0\%) & 40 (40.0\%) \\
 & Above average & 57 (57.0\%) & 42 (42.0\%) \\
 & Expert & 9 (9.0\%) & 16 (16.0\%) \\
\addlinespace
Social Media Usage & Never & 0 (0.0\%) & 2 (2.0\%) \\
 & Rarely (less than once a week) & 3 (3.0\%) & 1 (1.0\%) \\
 & Sometimes (1--3 times per week) & 6 (6.0\%) & 3 (3.0\%) \\
 & Often (4--6 times per week) & 3 (3.0\%) & 14 (14.0\%) \\
 & Daily & 37 (37.0\%) & 31 (31.0\%) \\
 & Multiple times per day & 51 (51.0\%) & 49 (49.0\%) \\
\addlinespace
Deepfake Familiarity & Not familiar at all & 3 (3.0\%) & 4 (4.0\%) \\
 & Slightly familiar & 5 (5.0\%) & 9 (9.0\%) \\
 & Moderately familiar & 16 (16.0\%) & 26 (26.0\%) \\
 & Very familiar & 50 (50.0\%) & 33 (33.0\%) \\
 & Extremely familiar & 26 (26.0\%) & 28 (28.0\%) \\
\addlinespace
Prior Deepfake Exposure & Definitely not & 4 (4.0\%) & 10 (10.0\%) \\
 & Probably not & 2 (2.0\%) & 7 (7.0\%) \\
 & Might or might not & 8 (8.0\%) & 7 (7.0\%) \\
 & Probably yes & 30 (30.0\%) & 26 (26.0\%) \\
 & Definitely yes & 56 (56.0\%) & 50 (50.0\%) \\
\addlinespace
Self-Reported Confidence & Not confident at all & 4 (4.0\%) & 1 (1.0\%) \\
 & Slightly confident & 17 (17.0\%) & 30 (30.0\%) \\
 & Moderately confident & 46 (46.0\%) & 37 (37.0\%) \\
 & Very confident & 24 (24.0\%) & 23 (23.0\%) \\
 & Extremely confident & 9 (9.0\%) & 9 (9.0\%) \\
\midrule
Total & $N$ & 100 & 100 \\
\bottomrule
\end{tabular}
\end{table}

% The demographic distributions of participants are reported in Table~\ref{tab:demographics}.
% Uncomment and add table reference once demographics table is finalized.

\subsection{Randomization}

In both detection experiments, the order of video presentation was fully randomized for each participant. Each participant encountered each video at most once. In the DF40 study, each participant viewed 60 videos randomly sampled from the pool of 1{,}000 (500 real, 500 fake). In the CharadesDF study, each participant likewise viewed 60 videos randomly sampled from the corresponding pool of 1{,}000 videos. This randomization ensured that each video received multiple independent ratings while controlling for order effects.

\subsection{Quality Features Extraction}
To examine how visual quality characteristics influence detection performance, we extracted 20 quality features from each video using a two-stage pipeline. We first applied face alignment and cropping using dlib's frontal face detector with 81-point landmark prediction, following the preprocessing protocol from DeepfakeBench~\cite{DBLP:journals/corr/abs-2307-01426}\footnote{\url{https://github.com/SCLBD/DeepfakeBench/blob/main/preprocessing/preprocess.py}}. Faces were aligned using a similarity transformation based on five key facial landmarks (left eye, right eye, nose tip, left mouth corner, right mouth corner) and resized to $256 \times 256$ pixels. Videos without detectable faces in all sampled frames were excluded, yielding 997 videos in DF40 and 779 in CharadesDF.

Quality features were extracted at the frame level and aggregated to video-level statistics using the mean across all frames containing detected faces. We used two complementary extraction approaches: (1) geometric and photometric features computed directly from facial landmarks and image statistics on the original, unprocessed frames, and (2) semantic quality factors derived from CLIB-FIQA~\cite{ou2024clib}, a vision-language model trained to assess face image quality through joint learning with multiple quality factors, applied to the preprocessed and aligned face crops.

Table~\ref{tab:quality_features} provides detailed descriptions of all extracted features. Features were standardized (zero mean, unit variance) within each dataset prior to regression analysis.

\begin{table}[htbp]
\centering
\small
\caption{Quality features extracted for each video. Features are computed per-frame and aggregated to video-level statistics using the mean across frames with detected faces.}
\label{tab:quality_features}
\scalebox{0.85}{
\begin{tabular}{p{3.2cm}p{12cm}}
\toprule
\textbf{Feature} & \textbf{Description} \\
\midrule
\multicolumn{2}{l}{\textit{Face Detection and Recognition}} \\
\addlinespace[2pt]
FR Confidence & Face detection confidence score from DeepFace~\cite{serengil2024lightface}, reflecting how reliably the face was detected and localized. Higher values indicate clearer, more frontal faces. \\
\addlinespace[4pt]
Bbox Area (ratio) & Ratio of the face bounding box area to the total frame area, computed as $(w_{\text{bbox}} \times h_{\text{bbox}}) / (w_{\text{frame}} \times h_{\text{frame}})$. Larger values indicate more prominent faces in the frame. \\
\addlinespace[4pt]
Face Presence (ratio) & Proportion of sampled frames in which at least one face was successfully detected. \\
\midrule
\multicolumn{2}{l}{\textit{Facial Geometry}} \\
\addlinespace[2pt]
IOD & Inter-Ocular Distance: Euclidean distance in pixels between the centers of the left and right eyes, computed from 68-point facial landmarks. Serves as a proxy for face scale and distance from camera. \\
\addlinespace[4pt]
Left EAR & Left Eye Aspect Ratio: ratio of vertical to horizontal eye extent, computed as $(||p_1 - p_5|| + ||p_2 - p_4||) / (2 \cdot ||p_0 - p_3||)$ where $p_0$--$p_5$ are the six landmarks of the left eye (indices 36--41 in the 68-point model)~\cite{soukupova2016eye}. Lower values indicate more closed eyes. \\
\addlinespace[4pt]
Right EAR & Right Eye Aspect Ratio, computed analogously for the right eye (indices 42--47). \\
\addlinespace[4pt]
MAR & Mouth Aspect Ratio: computed as $(A + B + C) / (2D)$ where $A$, $B$, $C$ are vertical distances between inner lip landmark pairs (51--59, 53--57, 49--55) and $D$ is the horizontal mouth width (48--54). Higher values indicate more open mouths. \\
\midrule
\multicolumn{2}{l}{\textit{Photometric Quality}} \\
\addlinespace[2pt]
SNR & Signal-to-Noise Ratio: estimated as the ratio of mean grayscale intensity to noise standard deviation, where noise is computed as the residual after Gaussian blur with a $5 \times 5$ kernel. Higher values indicate cleaner images. \\
\addlinespace[4pt]
Brightness & Mean pixel intensity of the grayscale-converted face region, ranging from 0 (black) to 255 (white). \\
\addlinespace[4pt]
Sharpness & Variance of the Laplacian operator applied to the grayscale face region. Higher values indicate sharper images with more high-frequency detail. \\
\addlinespace[4pt]
Contrast & Standard deviation of pixel intensities in the grayscale face region. Higher values indicate greater tonal range. \\
\addlinespace[4pt]
Color Balance & Standard deviation across mean R, G, B channel intensities, computed as $\text{std}([\bar{R}, \bar{G}, \bar{B}])$. Lower values indicate more neutral color balance; higher values indicate color casts. \\
\midrule
\multicolumn{2}{l}{\textit{CLIB-FIQA Semantic Factors}~\cite{ou2024clib}} \\
\addlinespace[2pt]
Blur (blurry) & Probability that the face image is blurry, predicted by CLIB-FIQA. Based on CPBD metric thresholds~\cite{narvekar2011no}: scores $< 0.35$ indicate hazy, $0.35$--$0.7$ indicate blurry, $> 0.7$ indicate clear. \\
\addlinespace[4pt]
Blur (hazy) & Probability that the face image exhibits haze or fog artifacts (CPBD score $< 0.35$). \\
\addlinespace[4pt]
Pose (profile) & Probability that the face is in profile view (yaw angle $> 25°$), classified using Euler angle estimation from facial landmarks. \\
\addlinespace[4pt]
Pose (slight angle) & Probability that the face is at a slight angle ($10°$--$25°$ yaw), intermediate between frontal and profile. \\
\addlinespace[4pt]
Occlusion & Probability that the face is partially occluded (e.g., by hands, objects, hair, or accessories), predicted by a CNN classifier trained on WiderFace~\cite{yang2016wider}. \\
\addlinespace[4pt]
Exaggerated Expression & Probability that the face displays an exaggerated or atypical expression (as opposed to neutral or typical expressions), predicted by a CNN classifier. \\
\addlinespace[4pt]
Extreme Lighting & Probability that the face is under extreme lighting conditions (e.g., harsh shadows, overexposure, backlighting), predicted by a CNN classifier. \\
\addlinespace[4pt]
Quality Score & Overall face image quality score from CLIB-FIQA, computed as a weighted combination of quality factor probabilities mapped to a five-level scale (bad, poor, fair, good, perfect). \\
\bottomrule
\end{tabular}}
\end{table}

\paragraph{Distributional differences between DF40 and CharadesDF.}
We conducted Mann-Whitney U tests with the Benjamini-Hochberg false discovery rate correction to compare quality feature distributions between DF40 and CharadesDF (Table~\ref{tab:dataset_quality_comparison}). Consistent with our design goals, CharadesDF exhibited significantly lower visual quality across multiple dimensions. Overall quality scores were substantially lower in CharadesDF ($\mu = 0.596$, $\text{Mdn} = 0.609$) compared to DF40 ($\mu = 0.676$, $\text{Mdn} = 0.679$), $p < 0.001$. Face presence ratio showed the largest absolute difference: DF40 videos contained detectable faces in nearly all frames ($\mu = 1.000$, $\text{Mdn} = 1.000$), whereas CharadesDF videos exhibited substantial frame-to-frame variability in face visibility ($\mu = 0.811$, $\text{Mdn} = 0.919$), reflecting the naturalistic recording conditions where participants moved freely within their environments.

CharadesDF exhibited greater degradation across photometric quality measures: higher blur prevalence (blurry: $\text{Mdn} = 3.0 \times 10^{-5}$ vs.\ $0.0$; hazy: $\text{Mdn} = 1.1 \times 10^{-6}$ vs.\ $0.0$), lower brightness ($\mu = 89.3$ vs.\ $112.6$), reduced sharpness ($\text{Mdn} = 40.2$ vs.\ $73.7$), and lower contrast ($\mu = 35.7$ vs.\ $40.4$). Pose variability was also greater in CharadesDF, with higher rates of profile views ($\text{Mdn} = 0.001$ vs.\ $4.2 \times 10^{-6}$) and increased occlusion ($\text{Mdn} = 8.8 \times 10^{-6}$ vs.\ $1.8 \times 10^{-7}$). These distributional differences confirm that CharadesDF presents substantially more challenging detection conditions than the controlled DF40 benchmark.

Interestingly, DF40 exhibited higher color balance values ($\mu = 25.3$ vs.\ $18.5$), indicating greater color cast variability, and higher mouth aspect ratios ($\mu = 0.891$ vs.\ $0.827$), potentially reflecting the prevalence of lip-sync deepfake methods in that benchmark. Signal-to-noise ratio was marginally higher in CharadesDF ($\text{Mdn} = 38.6$ vs.\ $34.3$), though both datasets showed high variability in this measure. All comparisons reached statistical significance after FDR correction ($p_{\mathrm{FDR}} < 0.05$), underscoring the systematic differences in recording conditions between the controlled YouTube-sourced videos in DF40 and the naturalistic home recordings in CharadesDF.

\begin{table}[htbp]
\centering
\small
\caption{Distributional comparison of quality features between DF40 and CharadesDF datasets. Mann-Whitney U tests with Benjamini-Hochberg FDR correction.}
\label{tab:dataset_quality_comparison}
\scalebox{0.85}{
\begin{tabular}{lrrrrrrl}
\toprule
 & \multicolumn{3}{c}{\textbf{DF40} ($N = 997$)} & \multicolumn{3}{c}{\textbf{CharadesDF} ($N = 779$)} & \\
\cmidrule(lr){2-4} \cmidrule(lr){5-7}
Quality Feature & Mean & Median & SD & Mean & Median & SD & Direction \\
\midrule
\multicolumn{8}{l}{\textit{Face Detection and Recognition}} \\
FR Confidence & 1.000 & 1.000 & 0.0006 & 0.9995 & 1.000 & 0.0024 & DF40 $>$$^{***\dagger}$ \\
Face Presence (ratio) & 1.000 & 1.000 & 0.0005 & 0.811 & 0.919 & 0.235 & DF40 $>$$^{***}$ \\
Bbox Area (ratio) & 0.076 & 0.054 & 0.069 & 0.078 & 0.052 & 0.081 & DF40 $>$$^{***}$ \\
\addlinespace
\multicolumn{8}{l}{\textit{Facial Geometry}} \\
IOD & 77.08 & 56.37 & 47.25 & 81.40 & 66.72 & 61.38 & DF40 $<$$^{*}$ \\
Left EAR & 0.291 & 0.291 & 0.033 & 0.306 & 0.280 & 0.106 & DF40 $>$$^{***}$ \\
Right EAR & 0.278 & 0.282 & 0.031 & 0.303 & 0.271 & 0.126 & DF40 $>$$^{***}$ \\
MAR & 0.891 & 0.891 & 0.083 & 0.827 & 0.793 & 0.172 & DF40 $>$$^{***}$ \\
\addlinespace
\multicolumn{8}{l}{\textit{Photometric Quality}} \\
SNR & 44.55 & 34.30 & 32.41 & 45.24 & 38.57 & 27.07 & DF40 $<$$^{***}$ \\
Brightness & 112.64 & 110.72 & 22.31 & 89.29 & 91.17 & 26.73 & DF40 $>$$^{***}$ \\
Sharpness & 110.02 & 73.75 & 109.78 & 73.82 & 40.16 & 99.14 & DF40 $>$$^{***}$ \\
Contrast & 40.40 & 39.38 & 10.51 & 35.71 & 36.18 & 10.90 & DF40 $>$$^{***}$ \\
Color Balance & 25.31 & 24.93 & 7.34 & 18.52 & 18.83 & 7.56 & DF40 $>$$^{***}$ \\
\addlinespace
\multicolumn{8}{l}{\textit{CLIB-FIQA Semantic Factors}~\cite{ou2024clib}} \\
Blur (blurry) & 0.126 & 0.000 & 0.323 & 0.294 & $3.0 \times 10^{-5}$ & 0.428 & DF40 $<$$^{***}$ \\
Blur (hazy) & 0.001 & 0.000 & 0.012 & 0.280 & $1.1 \times 10^{-6}$ & 0.436 & DF40 $<$$^{***}$ \\
Pose (profile) & 0.014 & $4.2 \times 10^{-6}$ & 0.097 & 0.138 & 0.001 & 0.285 & DF40 $<$$^{***}$ \\
Pose (slight angle) & 0.251 & 0.003 & 0.393 & 0.137 & 0.003 & 0.270 & DF40 $>$$^{***}$ \\
Occlusion & 0.001 & $1.8 \times 10^{-7}$ & 0.032 & 0.058 & $8.8 \times 10^{-6}$ & 0.203 & DF40 $<$$^{***}$ \\
Exaggerated expression & 0.000 & 0.000 & 0.000 & $3.8 \times 10^{-10}$ & 0.000 & $1.1 \times 10^{-8}$ & DF40 $<$$^{***\dagger}$ \\
Extreme lighting & 0.008 & 0.000 & 0.081 & 0.035 & $1.8 \times 10^{-7}$ & 0.164 & DF40 $<$$^{***}$ \\
Quality Score & 0.676 & 0.679 & 0.038 & 0.596 & 0.609 & 0.091 & DF40 $>$$^{***}$ \\
\bottomrule
\end{tabular}}
\vspace{2mm}
\begin{minipage}{\textwidth}
\footnotesize
\textit{Notes:} All comparisons significant after Benjamini-Hochberg FDR correction. 
$^{*}p_{\mathrm{FDR}}<0.05$, $^{**}p_{\mathrm{FDR}}<0.01$, $^{***}p_{\mathrm{FDR}}<0.001$.
Direction indicates which dataset has higher median values.
$^{\dagger}$Medians are equal; direction based on mean comparison. The Mann-Whitney U test detects significant distributional differences (e.g., in variance or tail behavior) even when medians coincide.
\end{minipage}
\end{table}

Table~\ref{tab:quality_examples} illustrates these quality features with representative frames from both datasets. 

\begin{table*}[htbp]
\centering
\small
\caption{Sampled frames from example video stimuli and their associated quality factor values.}
\label{tab:quality_examples}
\setlength{\tabcolsep}{4pt}
\scalebox{0.85}{
\begin{tabular}{l *{3}{c} c *{3}{c}}
\toprule
& \multicolumn{3}{c}{\textbf{DF40}} & & \multicolumn{3}{c}{\textbf{CharadesDF}} \\
\cmidrule(lr){2-4} \cmidrule(lr){6-8}
& \includegraphics[width=0.13\textwidth]{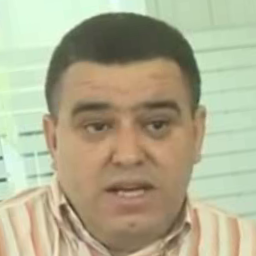}
& \includegraphics[width=0.13\textwidth]{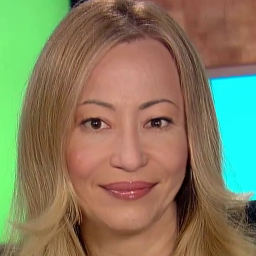}
& \includegraphics[width=0.13\textwidth]{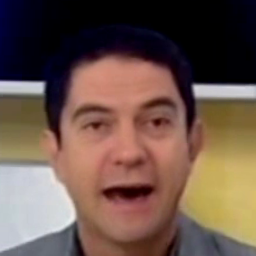}
& \phantom{ab}
& \includegraphics[width=0.13\textwidth]{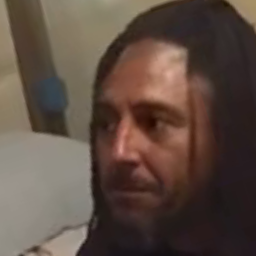}
& \includegraphics[width=0.13\textwidth]{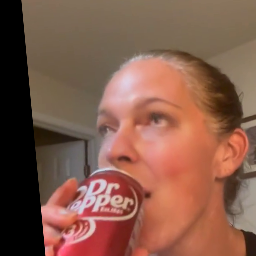}
& \includegraphics[width=0.13\textwidth]{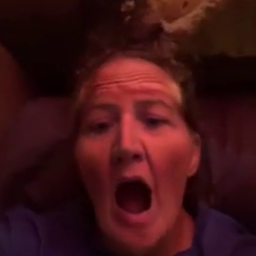} \\
& \small (a) & \small (b) & \small (c) & & \small (d) & \small (e) & \small (f) \\
\midrule
FR Confidence        & 1.000 & 1.000 & 1.000 & & 0.996 & 1.000 & 0.995 \\
Bbox Area (ratio)     & 0.062 & 0.054 & 0.052 & & 0.010 & 0.146 & 0.241 \\
Face Presence (ratio) & 1.00 & 1.00 & 1.00 & & 0.769 & 1.00 & 0.541 \\
IOD                   & 53.931 & 131.245 & 50.110 & & 28.668 & 135.385 & 60.197 \\
Left EAR                     & 0.255 & 0.261 & 0.289 & & 0.317 & 0.294 & 0.396 \\
Right EAR                    & 0.280 & 0.263 & 0.266 & & 0.347 & 0.266 & 0.292 \\
MAR                          & 0.807 & 0.804 & 0.768 & & 0.909 & 0.865 & 1.093 \\
SNR                   & 39.091 & 34.501 & 40.326 & & 25.063 & 141.181 & 33.403 \\
Brightness                   & 129.579 & 108.013 & 112.577 & & 53.813 & 119.192 & 42.627 \\
Sharpness             & 79.671 & 89.321 & 53.663 & & 167.868 & 9.040 & 32.407 \\
Contrast                     & 25.070 & 28.262 & 39.507 & & 27.422 & 32.592 & 26.575 \\
Color Balance                & 23.422 & 30.292 & 25.352 & & 12.479 & 29.534 & 21.420 \\
Blur (blurry)       & 0.000 & 0.000 & 0.000 & & 0.239 & 0.541 & 0.133 \\
Blur (hazy)         & 0.000 & 0.000 & 0.000 & & 0.761 & 0.045 & 0.866 \\
Pose (profile)      & 0.000 & 0.000 & 0.000 & & 0.242 & 0.195 & 0.000 \\
Pose (slight angle) & 0.000 & 0.000 & 0.000 & & 0.172 & 0.440 & 0.506 \\
Occlusion           & 0.000 & 0.000 & 0.000 & & 0.615 & 0.022 & 0.001 \\
Exag.\ Expression   & 0.000 & 0.000 & 0.000 & & 0.000 & 0.000 & 0.000 \\
Extreme Lighting    & 0.000 & 0.000 & 0.000 & & 0.000 & 0.000 & 0.012 \\
Quality Score         & 0.664 & 0.728 & 0.704 & & 0.526 & 0.628 & 0.639 \\
\bottomrule
\end{tabular}}
\vspace{1mm}
\begin{minipage}{\textwidth}
\footnotesize
\textit{Notes:} Each column shows video-level quality factor values for one stimulus. The images above are representative frames sampled from the corresponding video. Values are computed as per-video aggregates. $\uparrow$ indicates higher is better; $\downarrow$ indicates lower is better; --- indicates no direct quality relationship.
\end{minipage}
\end{table*}

\newpage\subsection{AI detectors}

We trained 32 state-of-the-art deepfake detection architectures on three established training datasets: FaceForensics++~\cite{DBLP:conf/iccv/RosslerCVRTN19}, CelebDF-v2~\cite{li2020celeb}, and the DF40 training split~\cite{DBLP:conf/nips/YanYCZFZLWDWY24}. This procedure yielded 96 detector variants, with one entry excluded due to architectural incompatibility: the LSDA detector~\cite{yan2024lsda} could not be trained on CelebDF-v2 because its architecture requires multiple forgery types during training (specifically, four distinct manipulation methods plus real videos) to learn domain-specific encoders and perform latent space augmentation, whereas CelebDF-v2 contains only a single forgery type. Each remaining detector variant was evaluated on all 1{,}000 videos in both the held-out DF40 test sample and the CharadesDF dataset. No overlap exists between training and evaluation data for any detector variant.

The evaluated architectures span diverse detection paradigms, including frequency-based methods (F3Net~\cite{qian2020f3net}, SPSL~\cite{liu2021spsl}, SRM~\cite{luo2021srm}), attention mechanisms (Multi-Attention~\cite{zhao2021multiattention}, FFD~\cite{dang2020ffd}), reconstruction-based approaches (RECCE~\cite{cao2022recce}), contrastive learning (UCF~\cite{yan2023ucf}, PCL-I2G~\cite{zhao2021pcl}), transformer architectures (UIA-ViT~\cite{zhuang2022uiavit}, TimeSformer~\cite{bertasius2021timesformer}), video-level temporal modeling (I3D~\cite{carreira2017i3d}, STIL~\cite{gu2021stil}, VideoMAE~\cite{tong2022videomae}), and foundation model adaptations (CLIP~\cite{radford2021clip}, X-CLIP~\cite{ni2022xclip}). Backbone architectures included EfficientNet-B4~\cite{tan2019efficientnet}, Xception~\cite{rossler2019faceforensics++}, and various ResNet variants. All models were trained following the protocols specified in DeepfakeBench~\cite{DBLP:journals/corr/abs-2307-01426} with regards to preprocessing, data augmentation, and optimization hyperparameters within each training dataset condition.

Each detector produced a continuous probability score $p \in [0,1]$ for each frame, where higher values indicate greater confidence that the frame depicts a deepfake. Frame-level scores were aggregated to video-level predictions by computing the median score across all frames. This aggregation strategy is appropriate for our evaluation setting, where manipulations span the entire video duration; in real-world scenarios where only a subset of frames may be manipulated, alternative aggregation methods---such as maximum pooling or top-$k$ averaging---would be more suitable to avoid diluting localized forgery signals. Binary predictions were obtained by thresholding the aggregated video-level score at $p = 0.5$, consistent with the human evaluation protocol. Tables~\ref{tab:detector_performance_acc}--\ref{tab:detector_performance_f1} report performance metrics (Accuracy, AUC, F1) for all detector--dataset combinations, with missing entries for LSDA on CelebDF-v2 reflecting the architectural incompatibility described above.

\begin{table*}[!htbp]
\centering
\caption{Performance (Accuracy) of deepfake detectors trained on FaceForensics++ (FF++), CelebDF-v2 (CDF), or DF40, and evaluated on our DF40 test sample and CharadesDF. Row and column averages are italicized.}
\label{tab:detector_performance_acc}
\resizebox{\textwidth}{!}{%
\begin{tabular}{lccccccc}
\toprule
\multirow{2}{*}{\textbf{Model}} & \multicolumn{2}{c}{\textbf{Trained on FF++}} & \multicolumn{2}{c}{\textbf{Trained on CDF}} & \multicolumn{2}{c}{\textbf{Trained on DF40}} & \multirow{2}{*}{\textbf{Avg}} \\
\cmidrule(lr){2-3} \cmidrule(lr){4-5} \cmidrule(lr){6-7}
 & \textbf{DF40} & \textbf{CharadesDF} & \textbf{DF40} & \textbf{CharadesDF} & \textbf{DF40} & \textbf{CharadesDF} &  \\
\midrule
Capsule~\cite{nguyen2019capsule} & 0.562 & 0.519 & 0.558 & 0.492 & 0.884 & 0.553 & \textit{0.595} \\
CLIP~\cite{radford2021clip} & 0.500 & 0.493 & 0.501 & 0.493 & 0.541 & 0.515 & \textit{0.507} \\
CORE~\cite{ni2022core} & 0.669 & 0.548 & 0.596 & 0.614 & 0.889 & 0.555 & \textit{0.645} \\
EfficientNet-B4~\cite{tan2019efficientnet} & 0.503 & 0.508 & 0.556 & 0.567 & 0.837 & 0.501 & \textit{0.579} \\
Effort~\cite{yan2025effort} & 0.648 & 0.577 & 0.658 & 0.601 & 0.780 & 0.628 & \textit{0.649} \\
F3Net~\cite{qian2020f3net} & 0.687 & 0.547 & 0.586 & 0.576 & 0.916 & 0.542 & \textit{0.642} \\
FFD~\cite{dang2020ffd} & 0.679 & 0.597 & 0.591 & 0.554 & 0.934 & 0.520 & \textit{0.646} \\
DSP-FWA~\cite{li2019fwa} & 0.511 & 0.501 & 0.378 & 0.417 & 0.500 & 0.510 & \textit{0.470} \\
I3D~\cite{carreira2017i3d} & 0.457 & 0.518 & 0.536 & 0.516 & 0.501 & 0.531 & \textit{0.510} \\
IID~\cite{huang2023iid} & 0.607 & 0.585 & 0.533 & 0.520 & 0.589 & 0.519 & \textit{0.559} \\
Local-relation~\cite{chen2021lrl} & 0.500 & 0.493 & 0.500 & 0.493 & 0.500 & 0.507 & \textit{0.499} \\
LSDA~\cite{yan2024lsda} & 0.508 & 0.509 & --- & --- & 0.614 & 0.565 & \textit{0.549} \\
MesoNet~\cite{afchar2018mesonet} & 0.500 & 0.493 & 0.526 & 0.509 & 0.500 & 0.507 & \textit{0.506} \\
MesoInception~\cite{afchar2018mesonet} & 0.504 & 0.516 & 0.586 & 0.601 & 0.645 & 0.510 & \textit{0.560} \\
Multi-Attention~\cite{zhao2021multiattention} & 0.580 & 0.493 & 0.591 & 0.521 & 0.603 & 0.493 & \textit{0.547} \\
PCL-I2G~\cite{zhao2021pcl} & 0.504 & 0.507 & 0.530 & 0.585 & 0.504 & 0.507 & \textit{0.523} \\
RECCE~\cite{cao2022recce} & 0.675 & 0.578 & 0.586 & 0.561 & 0.918 & 0.565 & \textit{0.647} \\
CNN-Aug~\cite{wang2020cnn} & 0.538 & 0.528 & 0.587 & 0.584 & 0.841 & 0.568 & \textit{0.608} \\
RFM~\cite{wang2021rfm} & 0.520 & 0.539 & 0.579 & 0.527 & 0.905 & 0.536 & \textit{0.601} \\
SBI~\cite{shiohara2022sbi} & 0.533 & 0.469 & 0.496 & 0.453 & 0.515 & 0.482 & \textit{0.492} \\
SIA~\cite{sun2022sia} & 0.498 & 0.530 & 0.536 & 0.542 & 0.707 & 0.493 & \textit{0.551} \\
SLADD~\cite{chen2022sladd} & 0.732 & 0.549 & 0.493 & 0.490 & 0.730 & 0.499 & \textit{0.582} \\
SPSL~\cite{liu2021spsl} & 0.793 & 0.601 & 0.601 & 0.602 & 0.931 & 0.525 & \textit{0.675} \\
SRM~\cite{luo2021srm} & 0.712 & 0.584 & 0.588 & 0.589 & 0.870 & 0.554 & \textit{0.649} \\
STIL~\cite{gu2021stil} & 0.550 & 0.522 & 0.607 & 0.544 & 0.898 & 0.606 & \textit{0.621} \\
TALL~\cite{xu2023tall} & 0.366 & 0.509 & 0.500 & 0.505 & 0.499 & 0.499 & \textit{0.480} \\
TimeSformer~\cite{bertasius2021timesformer} & 0.570 & 0.518 & 0.500 & 0.513 & 0.607 & 0.520 & \textit{0.538} \\
UCF~\cite{yan2023ucf} & 0.733 & 0.631 & 0.573 & 0.583 & 0.920 & 0.556 & \textit{0.666} \\
UIA-ViT~\cite{zhuang2022uiavit} & 0.679 & 0.588 & 0.579 & 0.583 & 0.843 & 0.600 & \textit{0.645} \\
VideoMAE~\cite{tong2022videomae} & 0.459 & 0.516 & 0.458 & 0.516 & 0.526 & 0.515 & \textit{0.498} \\
Xception~\cite{rossler2019faceforensics++} & 0.651 & 0.579 & 0.609 & 0.596 & 0.876 & 0.548 & \textit{0.643} \\
X-CLIP~\cite{ni2022xclip} & 0.366 & 0.509 & 0.366 & 0.509 & 0.499 & 0.508 & \textit{0.459} \\
\midrule
\textbf{Avg} & \textit{0.572} & \textit{0.536} & \textit{0.545} & \textit{0.541} & \textit{0.713} & \textit{0.532} & \textbf{0.573} \\
\bottomrule
\end{tabular}
}
\end{table*}

\begin{table*}[!htbp]
\centering
\caption{Performance (AUC) of deepfake detectors trained on FaceForensics++ (FF++), CelebDF-v2 (CDF), or DF40, and evaluated on our DF40 test sample and CharadesDF. Row and column averages are italicized.}
\label{tab:detector_performance_auc}
\resizebox{\textwidth}{!}{%
\begin{tabular}{lccccccc}
\toprule
\multirow{2}{*}{\textbf{Model}} & \multicolumn{2}{c}{\textbf{Trained on FF++}} & \multicolumn{2}{c}{\textbf{Trained on CDF}} & \multicolumn{2}{c}{\textbf{Trained on DF40}} & \multirow{2}{*}{\textbf{Avg}} \\
\cmidrule(lr){2-3} \cmidrule(lr){4-5} \cmidrule(lr){6-7}
 & \textbf{DF40} & \textbf{CharadesDF} & \textbf{DF40} & \textbf{CharadesDF} & \textbf{DF40} & \textbf{CharadesDF} &  \\
\midrule
Capsule~\cite{nguyen2019capsule} & 0.706 & 0.561 & 0.619 & 0.545 & 0.983 & 0.643 & \textit{0.676} \\
CLIP~\cite{radford2021clip} & 0.530 & 0.545 & 0.521 & 0.590 & 0.551 & 0.527 & \textit{0.544} \\
CORE~\cite{ni2022core} & 0.768 & 0.668 & 0.634 & 0.635 & 0.982 & 0.582 & \textit{0.712} \\
EfficientNet-B4~\cite{tan2019efficientnet} & 0.605 & 0.626 & 0.584 & 0.583 & 0.929 & 0.520 & \textit{0.641} \\
Effort~\cite{yan2025effort} & 0.821 & 0.753 & 0.721 & 0.674 & 0.953 & 0.709 & \textit{0.772} \\
F3Net~\cite{qian2020f3net} & 0.768 & 0.676 & 0.616 & 0.635 & 0.969 & 0.577 & \textit{0.707} \\
FFD~\cite{dang2020ffd} & 0.825 & 0.666 & 0.631 & 0.623 & 0.982 & 0.562 & \textit{0.715} \\
DSP-FWA~\cite{li2019fwa} & 0.572 & 0.586 & 0.320 & 0.409 & 0.465 & 0.597 & \textit{0.492} \\
I3D~\cite{carreira2017i3d} & 0.533 & 0.528 & 0.549 & 0.531 & 0.816 & 0.544 & \textit{0.583} \\
IID~\cite{huang2023iid} & 0.825 & 0.634 & 0.583 & 0.584 & 0.834 & 0.570 & \textit{0.672} \\
Local-relation~\cite{chen2021lrl} & 0.591 & 0.608 & 0.600 & 0.579 & 0.495 & 0.486 & \textit{0.560} \\
LSDA~\cite{yan2024lsda} & 0.771 & 0.662 & --- & --- & 0.731 & 0.641 & \textit{0.701} \\
MesoNet~\cite{afchar2018mesonet} & 0.623 & 0.499 & 0.606 & 0.564 & 0.574 & 0.559 & \textit{0.571} \\
MesoInception~\cite{afchar2018mesonet} & 0.607 & 0.644 & 0.627 & 0.623 & 0.908 & 0.543 & \textit{0.659} \\
Multi-Attention~\cite{zhao2021multiattention} & 0.432 & 0.500 & 0.621 & 0.606 & 0.478 & 0.500 & \textit{0.523} \\
PCL-I2G~\cite{zhao2021pcl} & 0.535 & 0.515 & 0.664 & 0.650 & 0.676 & 0.527 & \textit{0.595} \\
RECCE~\cite{cao2022recce} & 0.819 & 0.690 & 0.622 & 0.609 & 0.978 & 0.594 & \textit{0.719} \\
CNN-Aug~\cite{wang2020cnn} & 0.723 & 0.611 & 0.622 & 0.656 & 0.958 & 0.637 & \textit{0.701} \\
RFM~\cite{wang2021rfm} & 0.765 & 0.639 & 0.624 & 0.566 & 0.971 & 0.586 & \textit{0.692} \\
SBI~\cite{shiohara2022sbi} & 0.537 & 0.456 & 0.513 & 0.436 & 0.510 & 0.403 & \textit{0.476} \\
SIA~\cite{sun2022sia} & 0.537 & 0.606 & 0.556 & 0.558 & 0.891 & 0.496 & \textit{0.607} \\
SLADD~\cite{chen2022sladd} & 0.852 & 0.700 & 0.492 & 0.492 & 0.979 & 0.588 & \textit{0.684} \\
SPSL~\cite{liu2021spsl} & 0.867 & 0.634 & 0.647 & 0.631 & 0.983 & 0.546 & \textit{0.718} \\
SRM~\cite{luo2021srm} & 0.804 & 0.662 & 0.633 & 0.616 & 0.982 & 0.605 & \textit{0.717} \\
STIL~\cite{gu2021stil} & 0.816 & 0.594 & 0.671 & 0.582 & 0.971 & 0.627 & \textit{0.710} \\
TALL~\cite{xu2023tall} & 0.385 & 0.504 & 0.527 & 0.484 & 0.630 & 0.562 & \textit{0.515} \\
TimeSformer~\cite{bertasius2021timesformer} & 0.579 & 0.526 & 0.542 & 0.554 & 0.641 & 0.528 & \textit{0.562} \\
UCF~\cite{yan2023ucf} & 0.801 & 0.691 & 0.616 & 0.599 & 0.982 & 0.585 & \textit{0.712} \\
UIA-ViT~\cite{zhuang2022uiavit} & 0.769 & 0.692 & 0.617 & 0.608 & 0.946 & 0.621 & \textit{0.709} \\
VideoMAE~\cite{tong2022videomae} & 0.507 & 0.527 & 0.489 & 0.523 & 0.535 & 0.522 & \textit{0.517} \\
Xception~\cite{rossler2019faceforensics++} & 0.818 & 0.681 & 0.652 & 0.639 & 0.958 & 0.585 & \textit{0.722} \\
X-CLIP~\cite{ni2022xclip} & 0.386 & 0.509 & 0.399 & 0.512 & 0.357 & 0.506 & \textit{0.445} \\
\midrule
\textbf{Avg} & \textit{0.671} & \textit{0.606} & \textit{0.585} & \textit{0.577} & \textit{0.800} & \textit{0.565} & \textbf{0.635} \\
\bottomrule
\end{tabular}
}
\end{table*}

\begin{table*}[!htbp]
\centering
\caption{Performance (F1 score) of deepfake detectors trained on FaceForensics++ (FF++), CelebDF-v2 (CDF), or DF40, and evaluated on our DF40 test sample and CharadesDF. Row and column averages are italicized.}
\label{tab:detector_performance_f1}
\resizebox{\textwidth}{!}{%
\begin{tabular}{lccccccc}
\toprule
\multirow{2}{*}{\textbf{Model}} & \multicolumn{2}{c}{\textbf{Trained on FF++}} & \multicolumn{2}{c}{\textbf{Trained on CDF}} & \multicolumn{2}{c}{\textbf{Trained on DF40}} & \multirow{2}{*}{\textbf{Avg}} \\
\cmidrule(lr){2-3} \cmidrule(lr){4-5} \cmidrule(lr){6-7}
 & \textbf{DF40} & \textbf{CharadesDF} & \textbf{DF40} & \textbf{CharadesDF} & \textbf{DF40} & \textbf{CharadesDF} &  \\
\midrule
Capsule~\cite{nguyen2019capsule} & 0.681 & 0.665 & 0.629 & 0.635 & 0.895 & 0.674 & \textit{0.697} \\
CLIP~\cite{radford2021clip} & 0.667 & 0.661 & 0.667 & 0.661 & 0.526 & 0.607 & \textit{0.631} \\
CORE~\cite{ni2022core} & 0.727 & 0.683 & 0.553 & 0.618 & 0.876 & 0.597 & \textit{0.676} \\
EfficientNet-B4~\cite{tan2019efficientnet} & 0.668 & 0.666 & 0.534 & 0.620 & 0.848 & 0.568 & \textit{0.650} \\
Effort~\cite{yan2025effort} & 0.723 & 0.688 & 0.563 & 0.463 & 0.719 & 0.569 & \textit{0.621} \\
F3Net~\cite{qian2020f3net} & 0.722 & 0.678 & 0.587 & 0.649 & 0.913 & 0.594 & \textit{0.690} \\
FFD~\cite{dang2020ffd} & 0.739 & 0.690 & 0.644 & 0.657 & 0.933 & 0.612 & \textit{0.713} \\
DSP-FWA~\cite{li2019fwa} & 0.659 & 0.620 & 0.421 & 0.425 & 0.667 & 0.663 & \textit{0.576} \\
I3D~\cite{carreira2017i3d} & 0.624 & 0.519 & 0.394 & 0.087 & 0.643 & 0.531 & \textit{0.466} \\
IID~\cite{huang2023iid} & 0.363 & 0.501 & 0.168 & 0.070 & 0.705 & 0.660 & \textit{0.411} \\
Local-relation~\cite{chen2021lrl} & 0.667 & 0.661 & 0.667 & 0.661 & 0.000 & 0.000 & \textit{0.442} \\
LSDA~\cite{yan2024lsda} & 0.668 & 0.667 & --- & --- & 0.695 & 0.676 & \textit{0.676} \\
MesoNet~\cite{afchar2018mesonet} & 0.667 & 0.661 & 0.664 & 0.634 & 0.000 & 0.000 & \textit{0.438} \\
MesoInception~\cite{afchar2018mesonet} & 0.659 & 0.667 & 0.619 & 0.669 & 0.451 & 0.188 & \textit{0.542} \\
Multi-Attention~\cite{zhao2021multiattention} & 0.704 & 0.661 & 0.659 & 0.650 & 0.716 & 0.661 & \textit{0.675} \\
PCL-I2G~\cite{zhao2021pcl} & 0.024 & 0.000 & 0.161 & 0.424 & 0.024 & 0.000 & \textit{0.105} \\
RECCE~\cite{cao2022recce} & 0.732 & 0.695 & 0.578 & 0.637 & 0.914 & 0.566 & \textit{0.687} \\
CNN-Aug~\cite{wang2020cnn} & 0.678 & 0.668 & 0.603 & 0.472 & 0.858 & 0.672 & \textit{0.659} \\
RFM~\cite{wang2021rfm} & 0.674 & 0.676 & 0.571 & 0.626 & 0.907 & 0.624 & \textit{0.680} \\
SBI~\cite{shiohara2022sbi} & 0.499 & 0.148 & 0.434 & 0.321 & 0.490 & 0.186 & \textit{0.346} \\
SIA~\cite{sun2022sia} & 0.663 & 0.673 & 0.455 & 0.490 & 0.609 & 0.075 & \textit{0.494} \\
SLADD~\cite{chen2022sladd} & 0.763 & 0.672 & 0.460 & 0.470 & 0.786 & 0.663 & \textit{0.636} \\
SPSL~\cite{liu2021spsl} & 0.762 & 0.667 & 0.529 & 0.623 & 0.928 & 0.602 & \textit{0.685} \\
SRM~\cite{luo2021srm} & 0.744 & 0.691 & 0.586 & 0.639 & 0.883 & 0.682 & \textit{0.704} \\
STIL~\cite{gu2021stil} & 0.688 & 0.616 & 0.452 & 0.229 & 0.889 & 0.552 & \textit{0.571} \\
TALL~\cite{xu2023tall} & 0.511 & 0.375 & 0.667 & 0.637 & 0.658 & 0.629 & \textit{0.579} \\
TimeSformer~\cite{bertasius2021timesformer} & 0.609 & 0.499 & 0.667 & 0.611 & 0.488 & 0.424 & \textit{0.550} \\
UCF~\cite{yan2023ucf} & 0.708 & 0.687 & 0.446 & 0.575 & 0.915 & 0.624 & \textit{0.659} \\
UIA-ViT~\cite{zhuang2022uiavit} & 0.718 & 0.679 & 0.499 & 0.527 & 0.826 & 0.579 & \textit{0.638} \\
VideoMAE~\cite{tong2022videomae} & 0.629 & 0.538 & 0.628 & 0.538 & 0.342 & 0.134 & \textit{0.468} \\
Xception~\cite{rossler2019faceforensics++} & 0.725 & 0.690 & 0.589 & 0.640 & 0.867 & 0.574 & \textit{0.681} \\
X-CLIP~\cite{ni2022xclip} & 0.511 & 0.375 & 0.511 & 0.375 & 0.000 & 0.005 & \textit{0.296} \\
\midrule
\textbf{Avg} & \textit{0.643} & \textit{0.595} & \textit{0.536} & \textit{0.527} & \textit{0.655} & \textit{0.475} & \textbf{0.572} \\
\bottomrule
\end{tabular}
}
\end{table*}

\section{Data availability}
The DF40 dataset is publicly available from the original authors at \url{https://github.com/YZY-stack/DF40}. The CharadesDF dataset, including all 500 authentic home recordings and 500 corresponding deepfakes, will be made publicly available for non-commercial academic use upon publication. The original Charades dataset from which activity instructions were sourced is available at \url{https://prior.allenai.org/projects/charades}. 

\section{Code availability}
All code for reproducing the analyses reported in this paper, including scripts for quality feature extraction, ensemble aggregation, statistical analyses, and figure generation, will be made publicly available upon publication. AI detectors were trained using the DeepfakeBench framework (\url{https://github.com/SCLBD/DeepfakeBench}). Deepfakes for the CharadesDF dataset were generated using FaceFusion v3.2.0 (\url{https://facefusion.io}). Face quality assessment was performed using CLIB-FIQA (\url{https://github.com/oufuzhao/CLIB-FIQA}). All additional dependencies and environment specifications are documented in the code repository.

\newpage 
\bibliographystyle{plainnat}
\bibliography{sample-base} % for a file named science_template.bib

@inproceedings{DBLP:conf/iccv/RosslerCVRTN19,
  author       = {Andreas R{\"{o}}ssler and
                  Davide Cozzolino and
                  Luisa Verdoliva and
                  Christian Riess and
                  Justus Thies and
                  Matthias Nie{\ss}ner},
  title        = {FaceForensics++: Learning to Detect Manipulated Facial Images},
  booktitle    = {2019 {IEEE/CVF} International Conference on Computer Vision, {ICCV}
                  2019, Seoul, Korea (South), October 27 - November 2, 2019},
  pages        = {1--11},
  publisher    = {{IEEE}},
  year         = {2019},
  url          = {https://doi.org/10.1109/ICCV.2019.00009},
  doi          = {10.1109/ICCV.2019.00009},
  bibsource    = {dblp computer science bibliography, https://dblp.org}
}

@article{serengil2024lightface,
  title={A benchmark of facial recognition pipelines and co-usability performances of modules},
  author={Serengil, Sefik Ilkin and Ozpinar, Alper},
  journal={Journal of Information Technologies},
  year={2024}
}

@inproceedings{soukupova2016eye,
  title={Eye blink detection using facial landmarks},
  author={Soukupova, Tereza and Cech, Jan},
  booktitle={21st Computer Vision Winter Workshop},
  year={2016}
}

@article{narvekar2011no,
  title={A no-reference image blur metric based on the cumulative probability of blur detection (CPBD)},
  author={Narvekar, Niranjan D and Karam, Lina J},
  journal={IEEE Transactions on Image Processing},
  volume={20},
  number={9},
  pages={2678--2683},
  year={2011}
}

@inproceedings{yang2016wider,
  title={Wider face: A face detection benchmark},
  author={Yang, Shuo and Luo, Ping and Loy, Chen-Change and Tang, Xiaoou},
  booktitle={Proceedings of the IEEE Conference on Computer Vision and Pattern Recognition},
  pages={5525--5533},
  year={2016}
}

@article{DBLP:journals/corr/abs-2307-01426,
  title={DeepfakeBench: A comprehensive benchmark of deepfake detection},
  author={Yan, Zhiyuan and Zhang, Yong and Yuan, Xinhang and Lyu, Siwei and Wu, Baoyuan},
  journal={arXiv preprint arXiv:2307.01426},
  year={2023}
}

@inproceedings{ou2024clib,
  title={CLIB-FIQA: Face Image Quality Assessment with Confidence Calibration},
  author={Ou, Fu-Zhao and Li, Chongyi and Wang, Shiqi and Kwong, Sam},
  booktitle={Proceedings of the IEEE/CVF Conference on Computer Vision and Pattern Recognition},
  pages={1694--1704},
  year={2024}
}

@inproceedings{DBLP:conf/eccv/SigurdssonVWFLG16,
  author       = {Gunnar A. Sigurdsson and
                  G{\"{u}}l Varol and
                  Xiaolong Wang and
                  Ali Farhadi and
                  Ivan Laptev and
                  Abhinav Gupta},
  editor       = {Bastian Leibe and
                  Jiri Matas and
                  Nicu Sebe and
                  Max Welling},
  title        = {Hollywood in Homes: Crowdsourcing Data Collection for Activity Understanding},
  booktitle    = {Computer Vision - {ECCV} 2016 - 14th European Conference, Amsterdam,
                  The Netherlands, October 11-14, 2016, Proceedings, Part {I}},
  series       = {Lecture Notes in Computer Science},
  volume       = {9905},
  pages        = {510--526},
  publisher    = {Springer},
  year         = {2016},
  url          = {https://doi.org/10.1007/978-3-319-46448-0\_31},
  doi          = {10.1007/978-3-319-46448-0\_31},
  bibsource    = {dblp computer science bibliography, https://dblp.org}
}

@inproceedings{li2020celeb,
  title={Celeb-df: A large-scale challenging dataset for deepfake forensics},
  author={Li, Yuezun and Yang, Xin and Sun, Pu and Qi, Honggang and Lyu, Siwei},
  booktitle={Proceedings of the IEEE/CVF conference on computer vision and pattern recognition},
  pages={3207--3216},
  year={2020}
}

@article{47,
  author = {Robert Nichols and C. Rathgeb and P. Drozdowski and C. Busch},
  title = {Psychophysical Evaluation of Human Performance in Detecting Digital Face Image Manipulations},
  journal = {IEEE Access},
  year = {2022},
  doi = {10.1109/ACCESS.2022.3160596},
  volume = {10},
  pages = {31359-31376}
}

@inproceedings{DBLP:conf/nips/EigenPF14,
  author       = {David Eigen and
                  Christian Puhrsch and
                  Rob Fergus},
  editor       = {Zoubin Ghahramani and
                  Max Welling and
                  Corinna Cortes and
                  Neil D. Lawrence and
                  Kilian Q. Weinberger},
  title        = {Depth Map Prediction from a Single Image using a Multi-Scale Deep
                  Network},
  booktitle    = {Advances in Neural Information Processing Systems 27: Annual Conference
                  on Neural Information Processing Systems 2014, December 8-13 2014,
                  Montreal, Quebec, Canada},
  pages        = {2366--2374},
  year         = {2014},
  url          = {https://proceedings.neurips.cc/paper/2014/hash/7bccfde7714a1ebadf06c5f4cea752c1-Abstract.html},
  bibsource    = {dblp computer science bibliography, https://dblp.org}
}

@article{kowalski_faceswap_github,
  author       = {Marek Kowalski},
  title        = {FaceSwap: 3D Face Swapping Implemented in Python},
  year         = {n.d.},
  howpublished = {\url{https://github.com/MarekKowalski/FaceSwap}},
  note         = {GitHub repository, accessed February 2026},
}

@inproceedings{DBLP:conf/cvpr/ZhangCWZSGSW23,
  author       = {Wenxuan Zhang and
                  Xiaodong Cun and
                  Xuan Wang and
                  Yong Zhang and
                  Xi Shen and
                  Yu Guo and
                  Ying Shan and
                  Fei Wang},
  title        = {SadTalker: Learning Realistic 3D Motion Coefficients for Stylized
                  Audio-Driven Single Image Talking Face Animation},
  booktitle    = {{IEEE/CVF} Conference on Computer Vision and Pattern Recognition,
                  {CVPR} 2023, Vancouver, BC, Canada, June 17-24, 2023},
  pages        = {8652--8661},
  publisher    = {{IEEE}},
  year         = {2023},
  url          = {https://doi.org/10.1109/CVPR52729.2023.00836},
  doi          = {10.1109/CVPR52729.2023.00836},
  bibsource    = {dblp computer science bibliography, https://dblp.org}
}

@inproceedings{DBLP:conf/iccv/BounareliTAPT23,
  author       = {Stella Bounareli and
                  Christos Tzelepis and
                  Vasileios Argyriou and
                  Ioannis Patras and
                  Georgios Tzimiropoulos},
  title        = {HyperReenact: One-Shot Reenactment via Jointly Learning to Refine
                  and Retarget Faces},
  booktitle    = {{IEEE/CVF} International Conference on Computer Vision, {ICCV} 2023,
                  Paris, France, October 1-6, 2023},
  pages        = {7115--7125},
  publisher    = {{IEEE}},
  year         = {2023},
  url          = {https://doi.org/10.1109/ICCV51070.2023.00657},
  doi          = {10.1109/ICCV51070.2023.00657},
  bibsource    = {dblp computer science bibliography, https://dblp.org}
}

@inproceedings{DBLP:conf/mm/PrajwalMNJ20,
  author       = {K. R. Prajwal and
                  Rudrabha Mukhopadhyay and
                  Vinay P. Namboodiri and
                  C. V. Jawahar},
  editor       = {Chang Wen Chen and
                  Rita Cucchiara and
                  Xian{-}Sheng Hua and
                  Guo{-}Jun Qi and
                  Elisa Ricci and
                  Zhengyou Zhang and
                  Roger Zimmermann},
  title        = {A Lip Sync Expert Is All You Need for Speech to Lip Generation In
                  the Wild},
  booktitle    = {{MM} '20: The 28th {ACM} International Conference on Multimedia, Virtual
                  Event / Seattle, WA, USA, October 12-16, 2020},
  pages        = {484--492},
  publisher    = {{ACM}},
  year         = {2020},
  url          = {https://doi.org/10.1145/3394171.3413532},
  doi          = {10.1145/3394171.3413532},
  bibsource    = {dblp computer science bibliography, https://dblp.org}
}

@inproceedings{DBLP:conf/aaai/XuHDZHLD22,
  author       = {Zhiliang Xu and
                  Zhibin Hong and
                  Changxing Ding and
                  Zhen Zhu and
                  Junyu Han and
                  Jingtuo Liu and
                  Errui Ding},
  title        = {MobileFaceSwap: {A} Lightweight Framework for Video Face Swapping},
  booktitle    = {Thirty-Sixth {AAAI} Conference on Artificial Intelligence, {AAAI}
                  2022, Thirty-Fourth Conference on Innovative Applications of Artificial
                  Intelligence, {IAAI} 2022, The Twelveth Symposium on Educational Advances
                  in Artificial Intelligence, {EAAI} 2022 Virtual Event, February 22
                  - March 1, 2022},
  pages        = {2973--2981},
  publisher    = {{AAAI} Press},
  year         = {2022},
  url          = {https://doi.org/10.1609/aaai.v36i3.20203},
  doi          = {10.1609/AAAI.V36I3.20203},
  bibsource    = {dblp computer science bibliography, https://dblp.org}
}

@inproceedings{DBLP:conf/eccv/XuZHTZTWW022,
  author       = {Chao Xu and
                  Jiangning Zhang and
                  Yue Han and
                  Guanzhong Tian and
                  Xianfang Zeng and
                  Ying Tai and
                  Yabiao Wang and
                  Chengjie Wang and
                  Yong Liu},
  editor       = {Shai Avidan and
                  Gabriel J. Brostow and
                  Moustapha Ciss{\'{e}} and
                  Giovanni Maria Farinella and
                  Tal Hassner},
  title        = {Designing One Unified Framework for High-Fidelity Face Reenactment
                  and Swapping},
  booktitle    = {Computer Vision - {ECCV} 2022 - 17th European Conference, Tel Aviv,
                  Israel, October 23-27, 2022, Proceedings, Part {XV}},
  series       = {Lecture Notes in Computer Science},
  volume       = {13675},
  pages        = {54--71},
  publisher    = {Springer},
  year         = {2022},
  url          = {https://doi.org/10.1007/978-3-031-19784-0\_4},
  doi          = {10.1007/978-3-031-19784-0\_4},
  bibsource    = {dblp computer science bibliography, https://dblp.org}
}

@article{wang_inswapper_github,
  author       = {Haofan Wang},
  title        = {InSwapper: One-click Face Swapper and Restoration powered by InsightFace},
  year         = {2023},
  howpublished = {\url{https://github.com/haofanwang/inswapper}},
  note         = {GitHub repository, accessed February 2026},
}

@article{DBLP:journals/corr/abs-2106-06340,
  author       = {Renwang Chen and
                  Xuanhong Chen and
                  Bingbing Ni and
                  Yanhao Ge},
  title        = {SimSwap: An Efficient Framework For High Fidelity Face Swapping},
  journal      = {CoRR},
  volume       = {abs/2106.06340},
  year         = {2021},
  url          = {https://arxiv.org/abs/2106.06340},
  eprinttype    = {arXiv},
  eprint       = {2106.06340},
  bibsource    = {dblp computer science bibliography, https://dblp.org}
}

@inproceedings{DBLP:conf/iccv/NirkinKH19,
  author       = {Yuval Nirkin and
                  Yosi Keller and
                  Tal Hassner},
  title        = {{FSGAN:} Subject Agnostic Face Swapping and Reenactment},
  booktitle    = {2019 {IEEE/CVF} International Conference on Computer Vision, {ICCV}
                  2019, Seoul, Korea (South), October 27 - November 2, 2019},
  pages        = {7183--7192},
  publisher    = {{IEEE}},
  year         = {2019},
  url          = {https://doi.org/10.1109/ICCV.2019.00728},
  doi          = {10.1109/ICCV.2019.00728},
  bibsource    = {dblp computer science bibliography, https://dblp.org}
}

@inproceedings{DBLP:conf/iccv/ShioharaYT23,
  author       = {Kaede Shiohara and
                  Xingchao Yang and
                  Takafumi Taketomi},
  title        = {BlendFace: Re-designing Identity Encoders for Face-Swapping},
  booktitle    = {{IEEE/CVF} International Conference on Computer Vision, {ICCV} 2023,
                  Paris, France, October 1-6, 2023},
  pages        = {7600--7610},
  publisher    = {{IEEE}},
  year         = {2023},
  url          = {https://doi.org/10.1109/ICCV51070.2023.00702},
  doi          = {10.1109/ICCV51070.2023.00702},
  bibsource    = {dblp computer science bibliography, https://dblp.org}
}

@inproceedings{DBLP:conf/nips/YanYCZFZLWDWY24,
  author       = {Zhiyuan Yan and
                  Taiping Yao and
                  Shen Chen and
                  Yandan Zhao and
                  Xinghe Fu and
                  Junwei Zhu and
                  Donghao Luo and
                  Chengjie Wang and
                  Shouhong Ding and
                  Yunsheng Wu and
                  Li Yuan},
  editor       = {Amir Globersons and
                  Lester Mackey and
                  Danielle Belgrave and
                  Angela Fan and
                  Ulrich Paquet and
                  Jakub M. Tomczak and
                  Cheng Zhang},
  title        = {{DF40:} Toward Next-Generation Deepfake Detection},
  booktitle    = {Advances in Neural Information Processing Systems 38: Annual Conference
                  on Neural Information Processing Systems 2024, NeurIPS 2024, Vancouver,
                  BC, Canada, December 10 - 15, 2024},
  year         = {2024},
  url          = {http://papers.nips.cc/paper\_files/paper/2024/hash/34239f60eca7ce9bee5280aaf81362d8-Abstract-Datasets\_and\_Benchmarks\_Track.html},
  bibsource    = {dblp computer science bibliography, https://dblp.org}
}

@inproceedings{afchar2018mesonet,
  title={MesoNet: a Compact Facial Video Forgery Detection Network},
  author={Afchar, Darius and Nozick, Vincent and Yamagishi, Junichi and Echizen, Isao},
  booktitle={IEEE International Workshop on Information Forensics and Security (WIFS)},
  pages={1--7},
  year={2018},
  organization={IEEE}
}

@inproceedings{rossler2019faceforensics++,
  title={FaceForensics++: Learning to Detect Manipulated Facial Images},
  author={R{\"o}ssler, Andreas and Cozzolino, Davide and Verdoliva, Luisa and Riess, Christian and Thies, Justus and Nie{\ss}ner, Matthias},
  booktitle={Proceedings of the IEEE/CVF International Conference on Computer Vision (ICCV)},
  pages={1--11},
  year={2019}
}

@inproceedings{wang2020cnn,
  title={CNN-generated Images are Surprisingly Easy to Spot... for Now},
  author={Wang, Sheng-Yu and Wang, Oliver and Zhang, Richard and Owens, Andrew and Efros, Alexei A},
  booktitle={Proceedings of the IEEE/CVF Conference on Computer Vision and Pattern Recognition (CVPR)},
  pages={8695--8704},
  year={2020}
}

@inproceedings{tan2019efficientnet,
  title={EfficientNet: Rethinking Model Scaling for Convolutional Neural Networks},
  author={Tan, Mingxing and Le, Quoc},
  booktitle={International Conference on Machine Learning (ICML)},
  pages={6105--6114},
  year={2019},
  organization={PMLR}
}

@inproceedings{nguyen2019capsule,
  title={Capsule-Forensics: Using Capsule Networks to Detect Forged Images and Videos},
  author={Nguyen, Huy H and Yamagishi, Junichi and Echizen, Isao},
  booktitle={IEEE International Conference on Acoustics, Speech and Signal Processing (ICASSP)},
  pages={2307--2311},
  year={2019},
  organization={IEEE}
}

@inproceedings{li2019fwa,
  title={Exposing DeepFake Videos By Detecting Face Warping Artifacts},
  author={Li, Yuezun and Lyu, Siwei},
  booktitle={IEEE Conference on Computer Vision and Pattern Recognition Workshops (CVPRW)},
  pages={46--52},
  year={2019}
}

@inproceedings{dang2020ffd,
  title={On the Detection of Digital Face Manipulation},
  author={Dang, Hao and Liu, Feng and Stehouwer, Joel and Liu, Xiaoming and Jain, Anil K},
  booktitle={Proceedings of the IEEE/CVF Conference on Computer Vision and Pattern Recognition (CVPR)},
  pages={5781--5790},
  year={2020}
}

@inproceedings{ni2022core,
  title={CORE: Consistent Representation Learning for Face Forgery Detection},
  author={Ni, Yunsheng and Meng, Depu and Yu, Changqian and Quan, Chengbin and Ren, Dongchun and Zhao, Youjian},
  booktitle={Proceedings of the IEEE/CVF Conference on Computer Vision and Pattern Recognition Workshops (CVPRW)},
  pages={12--21},
  year={2022}
}

@inproceedings{cao2022recce,
  title={End-to-End Reconstruction-Classification Learning for Face Forgery Detection},
  author={Cao, Junyi and Ma, Chao and Yao, Taiping and Chen, Shen and Ding, Shouhong and Yang, Xiaokang},
  booktitle={Proceedings of the IEEE/CVF Conference on Computer Vision and Pattern Recognition (CVPR)},
  pages={4113--4122},
  year={2022}
}

@inproceedings{yan2023ucf,
  title={UCF: Uncovering Common Features for Generalizable Deepfake Detection},
  author={Yan, Zhiyuan and Zhang, Yong and Fan, Yanbo and Wu, Baoyuan},
  booktitle={Proceedings of the IEEE/CVF International Conference on Computer Vision (ICCV)},
  pages={22412--22423},
  year={2023}
}

@inproceedings{chen2021lrl,
  title={Local Relation Learning for Face Forgery Detection},
  author={Chen, Shen and Yao, Taiping and Chen, Yang and Ding, Shouhong and Li, Jilin and Ji, Rongrong},
  booktitle={Proceedings of the AAAI Conference on Artificial Intelligence},
  volume={35},
  number={2},
  pages={1081--1088},
  year={2021}
}

@inproceedings{huang2023iid,
  title={Implicit Identity Driven Deepfake Face Swapping Detection},
  author={Huang, Baojin and Wang, Zhongyuan and Yang, Jifan and Ai, Jiaxin and Zou, Qin and Wang, Qian and Ye, Dengpan},
  booktitle={Proceedings of the IEEE/CVF Conference on Computer Vision and Pattern Recognition (CVPR)},
  pages={4490--4499},
  year={2023}
}

@inproceedings{wang2021rfm,
  title={Representative Forgery Mining for Fake Face Detection},
  author={Wang, Chengrui and Deng, Weihong},
  booktitle={Proceedings of the IEEE/CVF Conference on Computer Vision and Pattern Recognition (CVPR)},
  pages={14923--14932},
  year={2021}
}

@inproceedings{sun2022sia,
  title={An Information Theoretic Approach for Attention-Driven Face Forgery Detection},
  author={Sun, Ke and Liu, Hong and Ye, Qixiang and Gao, Yue and Liu, Jianzhuang and Shao, Ling and Ji, Rongrong},
  booktitle={European Conference on Computer Vision (ECCV)},
  pages={111--127},
  year={2022},
  organization={Springer}
}

@inproceedings{chen2022sladd,
  title={Self-supervised Learning of Adversarial Example: Towards Good Generalizations for Deepfake Detection},
  author={Chen, Liang and Zhang, Yong and Song, Yibing and Liu, Lingqiao and Wang, Jue},
  booktitle={Proceedings of the IEEE/CVF Conference on Computer Vision and Pattern Recognition (CVPR)},
  pages={18710--18719},
  year={2022}
}

@inproceedings{zhuang2022uiavit,
  title={UIA-ViT: Unsupervised Inconsistency-Aware Method based on Vision Transformer for Face Forgery Detection},
  author={Zhuang, Wanyi and Chu, Qi and Tan, Zhentao and Liu, Qiankun and Yuan, Haojie and Miao, Changtao and Luo, Zixiang and Yu, Nenghai},
  booktitle={European Conference on Computer Vision (ECCV)},
  pages={391--407},
  year={2022},
  organization={Springer}
}

@inproceedings{radford2021clip,
  title={Learning Transferable Visual Models From Natural Language Supervision},
  author={Radford, Alec and Kim, Jong Wook and Hallacy, Chris and Ramesh, Aditya and Goh, Gabriel and Agarwal, Sandhini and Sastry, Girish and Askell, Amanda and Mishkin, Pamela and Clark, Jack and others},
  booktitle={International Conference on Machine Learning (ICML)},
  pages={8748--8763},
  year={2021},
  organization={PMLR}
}

@inproceedings{shiohara2022sbi,
  title={Detecting Deepfakes with Self-Blended Images},
  author={Shiohara, Kaede and Yamasaki, Toshihiko},
  booktitle={Proceedings of the IEEE/CVF Conference on Computer Vision and Pattern Recognition (CVPR)},
  pages={18720--18729},
  year={2022}
}

@inproceedings{zhao2021pcl,
  title={Learning Self-Consistency for Deepfake Detection},
  author={Zhao, Tianchen and Xu, Xiang and Xu, Mingze and Ding, Hui and Xiong, Yuanjun and Xia, Wei},
  booktitle={Proceedings of the IEEE/CVF International Conference on Computer Vision (ICCV)},
  pages={15023--15033},
  year={2021}
}

@inproceedings{zhao2021multiattention,
  title={Multi-Attentional Deepfake Detection},
  author={Zhao, Hanqing and Zhou, Wenbo and Chen, Dongdong and Wei, Tianyi and Zhang, Weiming and Yu, Nenghai},
  booktitle={Proceedings of the IEEE/CVF Conference on Computer Vision and Pattern Recognition (CVPR)},
  pages={2185--2194},
  year={2021}
}

@inproceedings{yan2024lsda,
  title={Transcending Forgery Specificity with Latent Space Augmentation for Generalizable Deepfake Detection},
  author={Yan, Zhiyuan and Luo, Yuhao and Lyu, Siwei and Liu, Qingshan and Wu, Baoyuan},
  booktitle={Proceedings of the IEEE/CVF Conference on Computer Vision and Pattern Recognition (CVPR)},
  year={2024}
}

@article{neo2025strategies,
  author = {Celene Neo and D. Goh and Rachel Wan Ying Chun and C. S. Lee},
  title = {Uncovering strategies for identifying deepfakes},
  journal = {Inf. Res.},
  year = {2025},
  doi = {10.47989/ir30iconf47209},
  volume = {30},
  pages = {752-760}
}

@article{50,
  author = {Alberto Sanchez-Acedo and Alejandro Carbonell-Alcocer and Manuel Gértrudix and J. Rubio-Tamayo},
  title = {The challenges of media and information literacy in the artificial intelligence ecology: deepfakes and misinformation},
  journal = {Communication \&amp; Society},
  year = {2024},
  doi = {10.15581/003.37.4.223-239}
}

@article{glenn1950verification,
  title={Verification of forecasts expressed in terms of probability},
  author={Glenn, W Brier and others},
  journal={Monthly weather review},
  volume={78},
  number={1},
  pages={1--3},
  year={1950},
  publisher={War Department, Office of the Chief Signal Officer}
}

@inproceedings{DBLP:conf/aaai/NaeiniCH15,
  author       = {Mahdi Pakdaman Naeini and
                  Gregory F. Cooper and
                  Milos Hauskrecht},
  editor       = {Blai Bonet and
                  Sven Koenig},
  title        = {Obtaining Well Calibrated Probabilities Using Bayesian Binning},
  booktitle    = {Proceedings of the Twenty-Ninth {AAAI} Conference on Artificial Intelligence,
                  January 25-30, 2015, Austin, Texas, {USA}},
  pages        = {2901--2907},
  publisher    = {{AAAI} Press},
  year         = {2015},
  url          = {https://doi.org/10.1609/aaai.v29i1.9602},
  doi          = {10.1609/AAAI.V29I1.9602},
  bibsource    = {dblp computer science bibliography, https://dblp.org}
}

@article{cuevas2026deepfake,
  title={Deepfake Pornography is Resilient to Regulatory and Platform Shocks},
  author={Cuevas, Alejandro and Ribeiro, Manoel Horta},
  journal={arXiv preprint arXiv:2602.02754},
  year={2026}
}

@article{21,
  author = {D. Goh},
  title = {“He looks very real”: Media, knowledge, and search‐based strategies for deepfake identification},
  journal = {Journal of the Association for Information Science and Technology},
  year = {2024},
  doi = {10.1002/asi.24867},
  volume = {75},
  pages = {643 - 654}
}

@article{49,
  author = {D. Goh},
  title = {“He looks very real”: Media, knowledge, and search‐based strategies for deepfake identification},
  journal = {Journal of the Association for Information Science and Technology},
  year = {2024},
  doi = {10.1002/asi.24867},
  volume = {75},
  pages = {643 - 654}
}

@article{55,
  author = {Sankini Rancha Godage and Frøy Løvåsdal and S. Venkatesh and K. Raja and Raghavendra Ramachandra and C. Busch},
  title = {Analyzing Human Observer Ability in Morphing Attack Detection—Where Do We Stand?},
  journal = {IEEE Transactions on Technology and Society},
  year = {2022},
  doi = {10.1109/TTS.2022.3231450},
  volume = {4},
  pages = {125-145}
}

@inproceedings{yan2025effort,
  title={Orthogonal Subspace Decomposition for Generalizable AI-Generated Image Detection},
  author={Yan, Zhiyuan and Wang, Jiangming and Jin, Peng and Zhang, Ke-Yue and Liu, Chengchun and Chen, Shen and Yao, Taiping and Ding, Shouhong and Wu, Baoyuan and Yuan, Li},
  booktitle={International Conference on Machine Learning (ICML)},
  year={2025},
  note={Spotlight}
}

@inproceedings{qian2020f3net,
  title={Thinking in Frequency: Face Forgery Detection by Mining Frequency-Aware Clues},
  author={Qian, Yuyang and Yin, Guojun and Sheng, Lu and Chen, Zixuan and Shao, Jing},
  booktitle={European Conference on Computer Vision (ECCV)},
  pages={86--103},
  year={2020},
  organization={Springer}
}

@inproceedings{liu2021spsl,
  title={Spatial-Phase Shallow Learning: Rethinking Face Forgery Detection in Frequency Domain},
  author={Liu, Honggu and Li, Xiaodan and Zhou, Wenbo and Chen, Yuefeng and He, Yuan and Xue, Hui and Zhang, Weiming and Yu, Nenghai},
  booktitle={Proceedings of the IEEE/CVF Conference on Computer Vision and Pattern Recognition (CVPR)},
  pages={772--781},
  year={2021}
}

@inproceedings{luo2021srm,
  title={Generalizing Face Forgery Detection with High-frequency Features},
  author={Luo, Yuchen and Zhang, Yong and Yan, Junchi and Liu, Wei},
  booktitle={Proceedings of the IEEE/CVF Conference on Computer Vision and Pattern Recognition (CVPR)},
  pages={16317--16326},
  year={2021}
}

@inproceedings{xu2023tall,
  title={TALL: Thumbnail Layout for Deepfake Video Detection},
  author={Xu, Yuting and Liang, Jian and Jia, Gengyun and Yang, Ziming and Zhang, Yanhao and He, Ran},
  booktitle={Proceedings of the IEEE/CVF International Conference on Computer Vision (ICCV)},
  pages={22658--22668},
  year={2023}
}

@inproceedings{gu2021stil,
  title={Spatiotemporal Inconsistency Learning for DeepFake Video Detection},
  author={Gu, Zhihao and Chen, Yang and Yao, Taiping and Ding, Shouhong and Li, Jilin and Huang, Feiyue and Ma, Lizhuang},
  booktitle={Proceedings of the 29th ACM International Conference on Multimedia},
  pages={3473--3481},
  year={2021}
}

@inproceedings{bertasius2021timesformer,
  title={Is Space-Time Attention All You Need for Video Understanding?},
  author={Bertasius, Gedas and Wang, Heng and Torresani, Lorenzo},
  booktitle={International Conference on Machine Learning (ICML)},
  pages={813--824},
  year={2021},
  organization={PMLR}
}

@inproceedings{tong2022videomae,
  title={VideoMAE: Masked Autoencoders are Data-Efficient Learners for Self-Supervised Video Pre-Training},
  author={Tong, Zhan and Song, Yibing and Wang, Jue and Wang, Limin},
  booktitle={Advances in Neural Information Processing Systems (NeurIPS)},
  volume={35},
  pages={10078--10093},
  year={2022}
}

@inproceedings{ni2022xclip,
  title={Expanding Language-Image Pretrained Models for General Video Recognition},
  author={Ni, Bolin and Peng, Houwen and Chen, Minghao and Zhang, Songyang and Meng, Gaofeng and Fu, Jianlong and Xiang, Shiming and Ling, Haibin},
  booktitle={European Conference on Computer Vision (ECCV)},
  pages={1--18},
  year={2022},
  organization={Springer}
}

@inproceedings{carreira2017i3d,
  title={Quo Vadis, Action Recognition? A New Model and the Kinetics Dataset},
  author={Carreira, Jo{\~a}o and Zisserman, Andrew},
  booktitle={Proceedings of the IEEE/CVF Conference on Computer Vision and Pattern Recognition (CVPR)},
  pages={6299--6308},
  year={2017}
}

@article{kobis2021fooled,
  title={Fooled twice: People cannot detect deepfakes but think they can},
  author={K{\"o}bis, Nils C and Dole{\v{z}}alov{\'a}, Barbora and Soraperra, Ivan},
  journal={Iscience},
  volume={24},
  number={11},
  year={2021},
  publisher={Elsevier}
}

@article{appel2022political,
  title={The detection of political deepfakes},
  author={Appel, Markus and Prietzel, Fabian},
  journal={Journal of Computer-Mediated Communication},
  volume={27},
  number={4},
  pages={zmac008},
  year={2022},
  publisher={Oxford University Press}
}

@article{sanchezacedo2024mil,
  title={The challenges of media and information literacy in the artificial intelligence ecology: deepfakes and misinformation},
  author={Sanchez-Acedo, Alberto and Carbonell-Alcocer, Alejandro and Gertrudix, Manuel and Rubio-Tamayo, Jose-Luis},
  journal={Communication \& Society},
  pages={223--239},
  year={2024}
}

@inproceedings{preu2022perception,
  title={Perception vs. reality: Understanding and evaluating the impact of synthetic image deepfakes over college students},
  author={Preu, Ethan and Jackson, Mark and Choudhury, Nazim},
  booktitle={2022 IEEE 13th Annual Ubiquitous Computing, Electronics \& Mobile Communication Conference (UEMCON)},
  pages={0547--0553},
  year={2022},
  organization={IEEE}
}

@article{kamali2025diffusion,
  author = {Negar Kamali and Karyn Nakamura and Aakriti Kumar and Angelos Chatzimparmpas and Jessica Hullman and Matthew Groh},
  title = {Characterizing Photorealism and Artifacts in Diffusion Model-Generated Images},
  journal = {Proceedings of the 2025 CHI Conference on Human Factors in Computing Systems},
  year = {2025},
  doi = {10.1145/3706598.3713962}
}

@article{frank2024representative,
  author = {Joel Frank and Franziska Herbert and Jonas Ricker and L. Schönherr and Thorsten Eisenhofer and Asja Fischer and Markus Dürmuth and Thorsten Holz},
  title = {A Representative Study on Human Detection of Artificially Generated Media Across Countries},
  journal = {2024 IEEE Symposium on Security and Privacy (SP)},
  year = {2023},
  doi = {10.1109/SP54263.2024.00159},
  pages = {55-73}
}

@article{cooke2024coin,
  author = {Di Cooke and Abigail Edwards and Sophia Barkoff and Kathryn Kelly},
  title = {As Good as a Coin Toss: Human Detection of AI-Generated Content},
  journal = {Communications of the ACM},
  year = {2025},
  doi = {10.1145/3729417},
  volume = {68},
  pages = {100 - 109}
}

@article{goh2024humansvsmachines,
  author = {D. Goh and Jonathan Pan and C. S. Lee},
  title = {Humans Versus Machines: A Deepfake Detection Faceoff},
  journal = {Proceedings of the Association for Information Science and Technology},
  year = {2024},
  doi = {10.1002/pra2.1139},
  volume = {61}
}

@article{groh2021deepfake,
  author = {Matthew Groh and Ziv Epstein and C. Firestone and Rosalind W. Picard},
  title = {Deepfake detection by human crowds, machines, and machine-informed crowds},
  journal = {Proceedings of the National Academy of Sciences of the United States of America},
  year = {2021},
  doi = {10.1073/pnas.2110013119},
  volume = {119}
}

@article{korshunov2021subjective,
  author = {Pavel Korshunov and S. Marcel},
  title = {Subjective and Objective Evaluation of Deepfake Videos},
  journal = {ICASSP 2021 - 2021 IEEE International Conference on Acoustics, Speech and Signal Processing (ICASSP)},
  year = {2021},
  doi = {10.1109/ICASSP39728.2021.9414258},
  pages = {2510-2514}
}

@article{josephs2023artifactmag,
  author = {Emilie Josephs and Camilo Luciano Fosco and A. Oliva},
  title = {Artifact magnification on deepfake videos increases human detection and subjective confidence},
  journal = {ArXiv},
  year = {2023},
  doi = {10.48550/arXiv.2304.04733},
  volume = {abs/2304.04733}
}

@article{pehlivanoglu2026susceptibility,
  author = {Didem Pehlivanoglu and Mengdi Zhu and Jialong Zhen and Aude A Gagnon-Roberge and Rebecca K Kern and Damon Woodard and Brian S Cahill and Natalie C Ebner},
  title = {Is this real? Susceptibility to deepfakes in machines and humans},
  journal = {Cognitive Research: Principles and Implications},
  year = {2026},
  doi = {10.1186/s41235-025-00700-y},
  volume = {11}
}

@article{somoray2025review,
  author = {Xinyi Jin and Guoyan Wang and Zhuoyue Zhang and Wenbo Zhou and Nenghai Yu and Bowen Gao and Shuqing Gao},
  title = {Identifying individual differences in deepfake discernment: the effects of cognitive disposition and visual literacy},
  journal = {Information, Communication \& Society},
  year = {2025},
  doi = {10.1080/1369118X.2025.2496902},
  volume = {28},
  pages = {3066 - 3085}
}

@article{mueller2021audiodeepfake,
  author = {N. Müller and Karla Markert and Konstantin Böttinger},
  title = {Human Perception of Audio Deepfakes},
  journal = {Proceedings of the 1st International Workshop on Deepfake Detection for Audio Multimedia},
  year = {2021},
  doi = {10.1145/3552466.3556531}
}

@article{hashmi2024audiovisual,
  author = {Ammarah Hashmi and Sahibzada Adil Shahzad and Chia-Wen Lin and Yu Tsao and Hsin-Min Wang},
  title = {Unmasking Illusions: Understanding Human Perception of Audiovisual Deepfakes},
  journal = {ArXiv},
  year = {2024},
  doi = {10.48550/arXiv.2405.04097},
  volume = {abs/2405.04097}
}

@article{bray2022deepfakefaces,
  author = {Sergi D. Bray and Shane D. Johnson and Bennett Kleinberg},
  title = {Testing Human Ability To Detect Deepfake Images of Human Faces},
  journal = {J. Cybersecur.},
  year = {2022},
  doi = {10.48550/arXiv.2212.05056},
  volume = {9}
}

@article{ibsen2024conditional,
  author = {M. Ibsen and Robert Nichols and C. Rathgeb and David J. Robertson and Josh P. Davis and Frøy Løvåsdal and Kiran Raja and Ryan E. Jenkins and Christoph Busch},
  title = {Conditional Face Image Manipulation Detection: Combining Algorithm and Human Examiner Decisions},
  journal = {Proceedings of the 2024 ACM Workshop on Information Hiding and Multimedia Security},
  year = {2024},
  doi = {10.1145/3658664.3659649}
}

@article{tahir2021seeing,
  author = {Rashid Tahir and Brishna Batool and Hira Jamshed and Mahnoor Jameel and Mubashir Anwar and Faizan Ahmed and M. Zaffar and Muhammad Fareed Zaffar},
  title = {Seeing is Believing: Exploring Perceptual Differences in DeepFake Videos},
  journal = {Proceedings of the 2021 CHI Conference on Human Factors in Computing Systems},
  year = {2021},
  doi = {10.1145/3411764.3445699}
}

@article{josephs2024browsing,
  author = {Emilie Josephs and Camilo Luciano Fosco and A. Oliva},
  title = {Effects of Browsing Conditions and Visual Alert Design on Human Susceptibility to Deepfakes},
  journal = {Journal of Online Trust and Safety},
  year = {2024},
  doi = {10.54501/jots.v2i2.144}
}

@article{casu2025automationbias,
  author = {Mirko Casu and Luca Guarnera and Ignazio Zangara and P. Caponnetto and S. Battiato},
  title = {A (Mid)journey Through Reality: Assessing Accuracy, Impostor Bias, and Automation Bias in Human Detection of AI‐Generated Images},
  journal = {Human Behavior and Emerging Technologies},
  year = {2025},
  doi = {10.1155/hbe2/9977058}
}

@inproceedings{boyd2022aiguidance,
  title={The value of ai guidance in human examination of synthetically-generated faces},
  author={Boyd, Aidan and Tinsley, Patrick and Bowyer, Kevin and Czajka, Adam},
  booktitle={Proceedings of the AAAI Conference on Artificial Intelligence},
  volume={37},
  number={5},
  pages={5930--5938},
  year={2023}
}

@article{chen2025selfefficacy,
  author = {Chen Chen and D. Goh},
  title = {The Role of Self‐Efficacy, Critical Thinking, and Media Literacy in Human Deepfake Video Detection},
  journal = {Proceedings of the Association for Information Science and Technology},
  year = {2025},
  doi = {10.1002/pra2.1310},
  volume = {62}
}

@article{janssen2024conflict,
  author = {Eva M. Janssen and Yarno F. Mutis and Tamara van Gog},
  title = {Looks real, feels fake: conflict detection in deepfake videos},
  journal = {Thinking \& Reasoning},
  year = {2024},
  doi = {10.1080/13546783.2024.2391794},
  volume = {31},
  pages = {237 - 247}
}

@article{miller2023aihyperreal,
  author = {Elizabeth J. Miller and B. A. Steward and Zak Witkower and C. Sutherland and Eva G. Krumhuber and Amy Dawel},
  title = {AI Hyperrealism: Why AI Faces Are Perceived as More Real Than Human Ones},
  journal = {Psychological Science},
  year = {2023},
  doi = {10.1177/09567976231207095},
  volume = {34},
  pages = {1390 - 1403}
}

@article{jin2025visual,
  author = {Jaron Mink and Miranda Wei and Collins W. Munyendo and Kurt Hugenberg and Tadayoshi Kohno and Elissa M. Redmiles and Gang Wang},
  title = {It's Trying Too Hard To Look Real: Deepfake Moderation Mistakes and Identity-Based Bias},
  journal = {Proceedings of the 2024 CHI Conference on Human Factors in Computing Systems},
  year = {2024},
  doi = {10.1145/3613904.3641999}
}

@article{alsobeh2024unmasking,
  author = {Anas Alsobeh and Alan Franklin and Belle Woodward and M. Porche and Joseph Siegelman},
  title = {UNMASKING MEDIA ILLUSION: ANALYTICAL SURVEY OF DEEPFAKE VIDEO DETECTION AND EMOTIONAL INSIGHTS},
  journal = {Issues In Information Systems},
  year = {2024},
  doi = {10.48009/2_iis_2024_108}
}

@article{davis2025superrecogniser,
  author = {Josh P. Davis and David J. Robertson and Ryan E. Jenkins and M. Ibsen and Robert Nichols and Martha Babbs and C. Rathgeb and Frøy Løvåsdal and Kiran Raja and Christoph Busch},
  title = {The Super‐Recogniser Advantage Extends to the Detection of Digitally Manipulated Faces},
  journal = {Applied Cognitive Psychology},
  year = {2025},
  doi = {10.1002/acp.70053}
}

@article{prasad2022noisychannels,
  author = {S. S. Prasad and O. Hadar and Thang Vu and I. Polian},
  title = {Human vs. Automatic Detection of Deepfake Videos Over Noisy Channels},
  journal = {2022 IEEE International Conference on Multimedia and Expo (ICME)},
  year = {2022},
  doi = {10.1109/ICME52920.2022.9859954},
  pages = {1-6}
}

@article{xu2025chasingshadows,
  author = {Ying Xu and Philipp Terhörst and Marius Pedersen and Kiran B. Raja},
  title = {Chasing Shadows: Solving Deepfake Detection Benchmarks Using Irrelevant Features Only},
  journal = {2025 IEEE 19th International Conference on Automatic Face and Gesture Recognition (FG)},
  year = {2025},
  doi = {10.1109/FG61629.2025.11099163},
  pages = {1-9}
}

@article{woehler2021perceptual,
  author = {L. Wöhler and Susana Castillo and Martin Zembaty and M. Magnor},
  title = {Towards Understanding Perceptual Differences between Genuine and Face-Swapped Videos},
  journal = {Proceedings of the 2021 CHI Conference on Human Factors in Computing Systems},
  year = {2021},
  doi = {10.1145/3411764.3445627}
}

@article{xie2025seeing,
  author = {Yijun Xie},
  title = {Seeing is no longer believing: a study of Deep Fake identification ability and its social impact},
  journal = {Advances in Engineering Innovation},
  year = {2025},
  doi = {10.54254/2977-3903/2025.28976}
}

@article{41,
  author = {Parul Gupta and Komal Chugh and Abhinav Dhall and Ramanathan Subramanian},
  title = {The eyes know it: FakeET- An Eye-tracking Database to Understand Deepfake Perception},
  journal = {Proceedings of the 2020 International Conference on Multimodal Interaction},
  year = {2020},
  doi = {10.1145/3382507.3418857}
}

@article{hasan2026uneven,
  author = {Md. Tarek Hasan and Sanjay Saha and Shaojing Fan and Swakkhar Shatabda and Terence Sim},
  title = {Deepfake Synthesis vs. Detection: An Uneven Contest},
  journal = {Unknown Journal},
  year = {2026}
}

\section{Acknowledgments}

N.A.

\section{Author contributions}

M.P. and V.S.S. conceived and designed the study. M.P. developed the CharadesDF dataset, trained and evaluated all AI detectors, designed and administered the human participant studies, performed all statistical analyses, created all figures and tables, and wrote the manuscript. I.G. generated the deepfake videos for the CharadesDF dataset. V.S.S. supervised the research, secured funding, and provided critical revisions to the manuscript. All authors reviewed and approved the final manuscript.

\section{Competing interests}
The authors declare no competing interests.

\appendix
\newpage\section{Supplementary Information}

\subsection{RQ1: Video-level performance}  
\begin{figure}[htbp]
    \centering
    \includegraphics[width=0.48\textwidth]{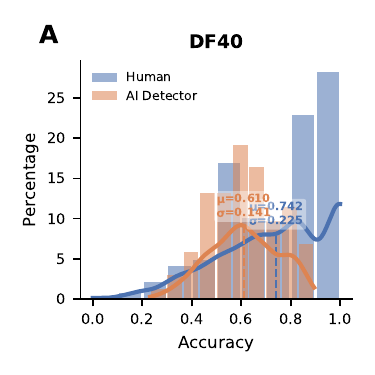}
    \hfill
    \includegraphics[width=0.48\textwidth]{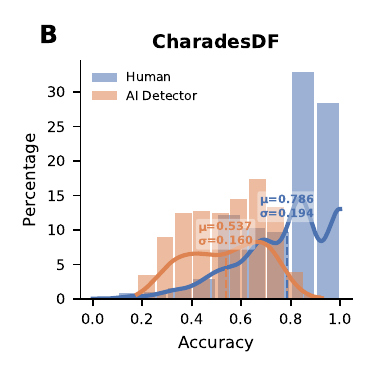}
    \caption{\textbf{Distribution of video-level accuracy for human participants and AI detectors.} (\textbf{A}) DF40 dataset. (\textbf{B}) CharadesDF dataset. Each data point represents the proportion of correct classifications for a single video, aggregated across all human participants or AI detectors.}
    \label{fig:RQ1_video_accuracy}
\end{figure}

To complement the participant-level analysis, we aggregated predictions at the video level, computing the proportion of correct classifications each video received across all human participants and all AI detectors (Figure~\ref{fig:RQ1_video_accuracy}). %This analysis reveals which videos were consistently easy or difficult to classify and whether human and machine performance patterns hold across individual stimuli.

On DF40 (Figure~\ref{fig:RQ1_video_accuracy}A), human accuracy per video ($\mu =$ 0.742, SD = 0.225) exceeded AI detectors accuracy ($\mu =$ 0.610, SD = 0.141), paired $t(996) = 17.564$, $p < .001$, $d = 0.556$. The broader distribution of human accuracy (SD = 0.225 vs. 0.141 for AI detectors) indicates greater variability in how consistently humans classified individual videos, with some videos achieving near-perfect human consensus while others proved challenging. The gap between humans and AI detectors was more pronounced on CharadesDF (Figure~\ref{fig:RQ1_video_accuracy}B): human accuracy ($\mu =$ 0.786, SD = 0.194) substantially exceeded AI detectors accuracy ($\mu =$ 0.537, SD = 0.160), paired $t(778) = 29.432$, $p < .001$, $d = 1.055$. 

\newpage\subsection{RQ2: Additional ensemble techniques}

To assess the robustness of our ensemble findings in RQ2, we evaluated four alternative aggregation strategies beyond the quality-weighted voting presented in Table~\ref{tab:ensemble_results}. {Soft Mean} (Table~\ref{tab:ensemble_results_soft_mean}) computes the simple arithmetic average of probability scores across all judges. {Soft Median} (Table~\ref{tab:ensemble_results_soft_median}) uses the median probability score, which is more robust to outliers. {Soft Max} (Table~\ref{tab:ensemble_results_soft_max}) takes the maximum probability score across judges, effectively flagging a video as a deepfake if any single judge is highly confident. {Hard Ensemble} (Table~\ref{tab:ensemble_results_hard}) applies majority voting on binarized predictions (threshold = 0.5) rather than aggregating continuous probability scores.

\begin{table}[h!]
\centering
\small
\caption{\textbf{Soft Mean Ensemble.} Values shown as point estimate $\pm$ error. Error represents half-width of 95\% bootstrap CI. CFR = Catastrophic Failure Rate. Bold indicates best ensemble performance.}
\label{tab:ensemble_results_soft_mean}
\scalebox{0.66}{
\begin{tabular}{l cccccccc}
\toprule
 & \multicolumn{4}{c}{\textbf{DF40}} & \multicolumn{4}{c}{\textbf{CharadesDF}} \\
\cmidrule(lr){2-5} \cmidrule(lr){6-9} 
 & Accuracy & F1 & AUC & CFR & Accuracy & F1 & AUC & CFR \\
\midrule
Human (indiv.) & 0.743 {\small $\pm$ 0.023} & 0.759 {\small $\pm$ 0.020} & 0.823 {\small $\pm$ 0.020} & 0.190 {\small $\pm$ 0.017} & 0.784 {\small $\pm$ 0.027} & 0.783 {\small $\pm$ 0.027} & 0.846 {\small $\pm$ 0.024} & 0.171 {\small $\pm$ 0.025} \\
AI detector (indiv.) & 0.610 {\small $\pm$ 0.030} & 0.612 {\small $\pm$ 0.042} & 0.686 {\small $\pm$ 0.037} & 0.279 {\small $\pm$ 0.036} & 0.537 {\small $\pm$ 0.009} & 0.533 {\small $\pm$ 0.041} & 0.583 {\small $\pm$ 0.014} & 0.317 {\small $\pm$ 0.029} \\
\midrule
Human ensemble & 0.881 {\small $\pm$ 0.020} & 0.881 {\small $\pm$ 0.022} & 0.948 {\small $\pm$ 0.013} & 0.020 {\small $\pm$ 0.009} & \textbf{0.928} {\small $\pm$ 0.017} & \textbf{0.926} {\small $\pm$ 0.019} & 0.975 {\small $\pm$ 0.010} & 0.012 {\small $\pm$ 0.007} \\
AI ensemble & 0.796 {\small $\pm$ 0.025} & 0.818 {\small $\pm$ 0.025} & 0.912 {\small $\pm$ 0.017} & 0.003 {\small $\pm$ 0.004} & 0.603 {\small $\pm$ 0.034} & 0.681 {\small $\pm$ 0.034} & 0.692 {\small $\pm$ 0.034} & 0.008 {\small $\pm$ 0.006} \\
Hybrid ensemble & \textbf{0.925} {\small $\pm$ 0.016} & \textbf{0.926} {\small $\pm$ 0.016} & \textbf{0.980} {\small $\pm$ 0.007} & \textbf{0.000} {\small $\pm$ 0.000} & 0.914 {\small $\pm$ 0.018} & 0.914 {\small $\pm$ 0.020} & \textbf{0.975} {\small $\pm$ 0.009} & \textbf{0.000} {\small $\pm$ 0.000} \\
\bottomrule
\end{tabular}

}
\end{table}

\begin{table}[h!]
\centering
\small
\caption{\textbf{Soft Median Ensemble.} Values shown as point estimate $\pm$ error. Error represents half-width of 95\% bootstrap CI. CFR = Catastrophic Failure Rate. Bold indicates best ensemble performance.}
\label{tab:ensemble_results_soft_median}
\scalebox{0.66}{
\begin{tabular}{l cccccccc}
\toprule
 & \multicolumn{4}{c}{\textbf{DF40}} & \multicolumn{4}{c}{\textbf{CharadesDF}} \\
\cmidrule(lr){2-5} \cmidrule(lr){6-9} 
 & Accuracy & F1 & AUC & CFR & Accuracy & F1 & AUC & CFR \\
\midrule
Human (indiv.) & 0.743 {\small $\pm$ 0.023} & 0.759 {\small $\pm$ 0.020} & 0.823 {\small $\pm$ 0.020} & 0.190 {\small $\pm$ 0.017} & 0.784 {\small $\pm$ 0.027} & 0.783 {\small $\pm$ 0.027} & 0.846 {\small $\pm$ 0.024} & 0.171 {\small $\pm$ 0.025} \\
AI detector (indiv.) & 0.610 {\small $\pm$ 0.030} & 0.612 {\small $\pm$ 0.042} & 0.686 {\small $\pm$ 0.037} & 0.279 {\small $\pm$ 0.036} & 0.537 {\small $\pm$ 0.009} & 0.533 {\small $\pm$ 0.041} & 0.583 {\small $\pm$ 0.014} & 0.317 {\small $\pm$ 0.029} \\
\midrule
Human ensemble & 0.857 {\small $\pm$ 0.022} & 0.860 {\small $\pm$ 0.023} & 0.924 {\small $\pm$ 0.016} & 0.067 {\small $\pm$ 0.015} & \textbf{0.920} {\small $\pm$ 0.018} & \textbf{0.919} {\small $\pm$ 0.020} & 0.961 {\small $\pm$ 0.013} & 0.045 {\small $\pm$ 0.014} \\
AI ensemble & 0.767 {\small $\pm$ 0.025} & 0.803 {\small $\pm$ 0.024} & 0.907 {\small $\pm$ 0.017} & 0.038 {\small $\pm$ 0.012} & 0.579 {\small $\pm$ 0.035} & 0.682 {\small $\pm$ 0.033} & 0.696 {\small $\pm$ 0.034} & 0.083 {\small $\pm$ 0.020} \\
Hybrid ensemble & \textbf{0.903} {\small $\pm$ 0.019} & \textbf{0.906} {\small $\pm$ 0.019} & \textbf{0.974} {\small $\pm$ 0.008} & \textbf{0.006} {\small $\pm$ 0.005} & 0.910 {\small $\pm$ 0.020} & 0.910 {\small $\pm$ 0.021} & \textbf{0.966} {\small $\pm$ 0.012} & \textbf{0.001} {\small $\pm$ 0.003} \\
\bottomrule
\end{tabular}

}
\end{table}

\begin{table}[h!]
\centering
\small
\caption{\textbf{Soft Max Ensemble. }Values shown as point estimate $\pm$ error. Error represents half-width of 95\% bootstrap CI. CFR = Catastrophic Failure Rate. Bold indicates best ensemble performance.}
\label{tab:ensemble_results_soft_max}
\scalebox{0.66}{
\begin{tabular}{l cccccccc}
\toprule
 & \multicolumn{4}{c}{\textbf{DF40}} & \multicolumn{4}{c}{\textbf{CharadesDF}} \\
\cmidrule(lr){2-5} \cmidrule(lr){6-9} 
 & Accuracy & F1 & AUC & CFR & Accuracy & F1 & AUC & CFR \\
\midrule
Human (indiv.) & 0.743 {\small $\pm$ 0.023} & 0.759 {\small $\pm$ 0.020} & 0.823 {\small $\pm$ 0.020} & 0.190 {\small $\pm$ 0.017} & 0.784 {\small $\pm$ 0.027} & 0.783 {\small $\pm$ 0.027} & 0.846 {\small $\pm$ 0.024} & 0.171 {\small $\pm$ 0.025} \\
AI detector (indiv.) & 0.610 {\small $\pm$ 0.030} & 0.612 {\small $\pm$ 0.042} & 0.686 {\small $\pm$ 0.037} & 0.279 {\small $\pm$ 0.036} & 0.537 {\small $\pm$ 0.009} & 0.533 {\small $\pm$ 0.041} & 0.583 {\small $\pm$ 0.014} & 0.317 {\small $\pm$ 0.029} \\
\midrule
Human ensemble & \textbf{0.568} {\small $\pm$ 0.032} & \textbf{0.697} {\small $\pm$ 0.027} & 0.847 {\small $\pm$ 0.022} & \textbf{0.331} {\small $\pm$ 0.028} & \textbf{0.597} {\small $\pm$ 0.033} & \textbf{0.710} {\small $\pm$ 0.031} & 0.848 {\small $\pm$ 0.023} & \textbf{0.317} {\small $\pm$ 0.033} \\
AI ensemble & 0.499 {\small $\pm$ 0.032} & 0.666 {\small $\pm$ 0.028} & 0.812 {\small $\pm$ 0.026} & 0.501 {\small $\pm$ 0.032} & 0.496 {\small $\pm$ 0.035} & 0.663 {\small $\pm$ 0.031} & 0.675 {\small $\pm$ 0.037} & 0.504 {\small $\pm$ 0.035} \\
Hybrid ensemble & 0.502 {\small $\pm$ 0.032} & 0.667 {\small $\pm$ 0.028} & \textbf{0.936} {\small $\pm$ 0.014} & 0.430 {\small $\pm$ 0.032} & 0.496 {\small $\pm$ 0.035} & 0.663 {\small $\pm$ 0.031} & \textbf{0.902} {\small $\pm$ 0.022} & 0.402 {\small $\pm$ 0.033} \\
\bottomrule
\end{tabular}
}
\end{table}

\begin{table}[h!]
\centering
\small
\caption{\textbf{Hard Ensemble. }Values shown as point estimate $\pm$ error. Error represents half-width of 95\% bootstrap CI. CFR = Catastrophic Failure Rate. Bold indicates best ensemble performance.}
\label{tab:ensemble_results_hard}
\scalebox{0.66}{
\begin{tabular}{l cccccccc}
\toprule
 & \multicolumn{4}{c}{\textbf{DF40}} & \multicolumn{4}{c}{\textbf{CharadesDF}} \\
\cmidrule(lr){2-5} \cmidrule(lr){6-9} 
 & Accuracy & F1 & AUC & CFR & Accuracy & F1 & AUC & CFR \\
\midrule
Human (indiv.) & 0.743 {\small $\pm$ 0.023} & 0.759 {\small $\pm$ 0.020} & 0.823 {\small $\pm$ 0.020} & 0.190 {\small $\pm$ 0.017} & 0.784 {\small $\pm$ 0.027} & 0.783 {\small $\pm$ 0.027} & 0.846 {\small $\pm$ 0.024} & 0.171 {\small $\pm$ 0.025} \\
AI detector (indiv.) & 0.610 {\small $\pm$ 0.030} & 0.612 {\small $\pm$ 0.042} & 0.686 {\small $\pm$ 0.037} & 0.279 {\small $\pm$ 0.036} & 0.537 {\small $\pm$ 0.009} & 0.533 {\small $\pm$ 0.041} & 0.583 {\small $\pm$ 0.014} & 0.317 {\small $\pm$ 0.029} \\
\midrule
Human ensemble & \textbf{0.821} {\small $\pm$ 0.026} & \textbf{0.835} {\small $\pm$ 0.025} & 0.822 {\small $\pm$ 0.025} & 0.179 {\small $\pm$ 0.026} & \textbf{0.897} {\small $\pm$ 0.021} & \textbf{0.899} {\small $\pm$ 0.022} & \textbf{0.898} {\small $\pm$ 0.021} & 0.103 {\small $\pm$ 0.021} \\
AI ensemble & 0.767 {\small $\pm$ 0.025} & 0.803 {\small $\pm$ 0.024} & 0.767 {\small $\pm$ 0.023} & 0.233 {\small $\pm$ 0.025} & 0.579 {\small $\pm$ 0.035} & 0.682 {\small $\pm$ 0.033} & 0.582 {\small $\pm$ 0.025} & 0.421 {\small $\pm$ 0.035} \\
Hybrid ensemble & 0.719 {\small $\pm$ 0.028} & 0.780 {\small $\pm$ 0.026} & \textbf{0.900} {\small $\pm$ 0.019} & \textbf{0.059} {\small $\pm$ 0.015} & 0.597 {\small $\pm$ 0.035} & 0.709 {\small $\pm$ 0.032} & 0.890 {\small $\pm$ 0.021} & \textbf{0.050} {\small $\pm$ 0.015} \\
\bottomrule
\end{tabular}
}
\end{table}

\newpage\subsection{RQ3: Performance curves by quality feature}

\begin{figure}[h!]
    \centering
    \includegraphics[width=.99\textwidth]{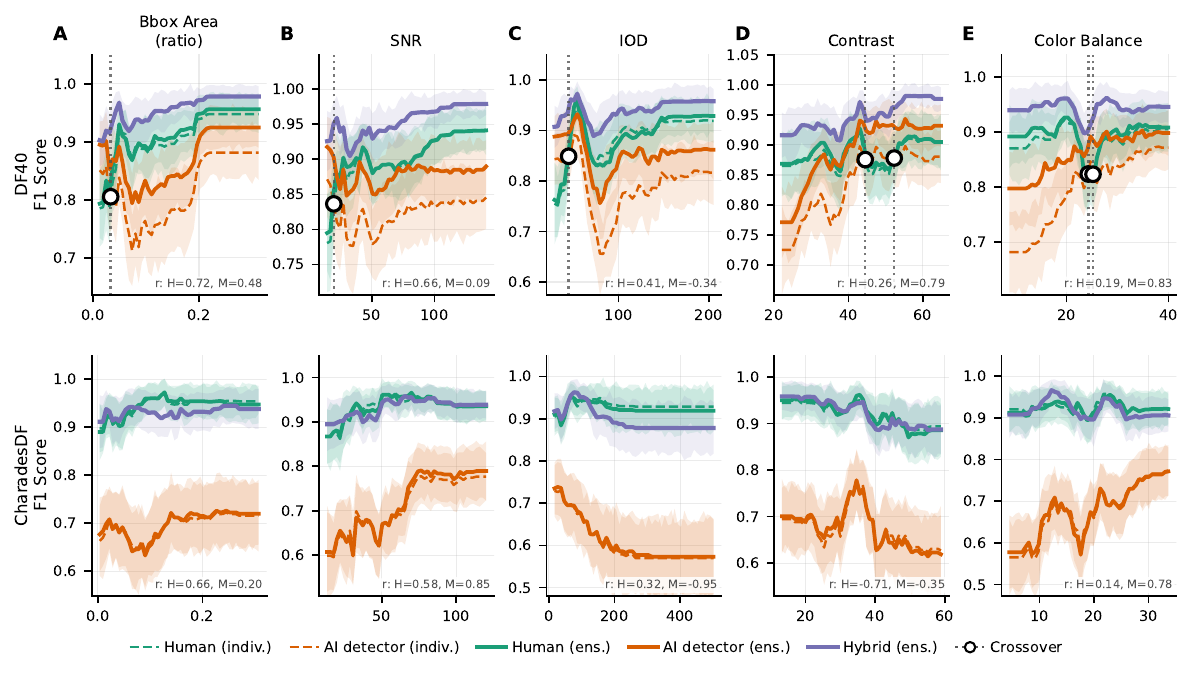}
    \caption{\textbf{Detection F1 as a function of quality features.}
    Smoothed performance curves showing the relationship between quality features and F1 for human individuals (teal, dashed), AI detectors (orange, dashed), human ensembles (teal, solid), AI ensembles (orange, solid), and hybrid ensembles (purple, solid). Shaded regions indicate 95\% confidence intervals. We indicate correlations ($\rho$) in each subplot for both humans ($H$) and AI detectors ($M$). Rows correspond to DF40 (top) and CharadesDF (bottom) datasets. Dashed vertical lines indicate human-AI crossovers.  
    (\textbf{A})~Bounding box area ratio. 
    (\textbf{B})~Signal-to-noise ratio (SNR).
    (\textbf{C})~Inter-ocular distance (IOD).
    (\textbf{D})~Contrast.
    (\textbf{E})~Color balance.}
    \label{fig:quality_curves_f1}
\end{figure}

\begin{figure}[h!]
    \centering
    \includegraphics[width=.99\textwidth]{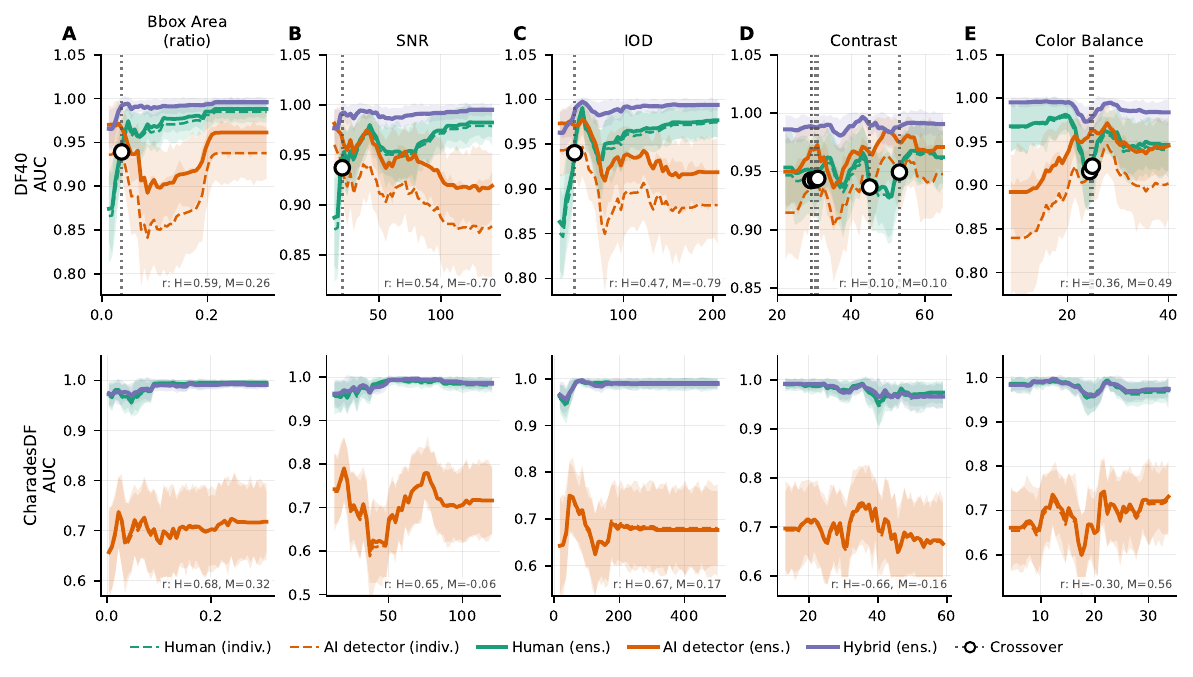}
    \caption{\textbf{Detection AUC as a function of quality features.}
    Smoothed performance curves showing the relationship between quality features and AUC for human individuals (teal, dashed), AI detectors (orange, dashed), human ensembles (teal, solid), AI ensembles (orange, solid), and hybrid ensembles (purple, solid). Shaded regions indicate 95\% confidence intervals. We indicate correlations ($\rho$) in each subplot for both humans ($H$) and AI detectors ($M$). Rows correspond to DF40 (top) and CharadesDF (bottom) datasets. Dashed vertical lines indicate human-AI crossovers.  
    (\textbf{A})~Bounding box area ratio. 
    (\textbf{B})~Signal-to-noise ratio (SNR).
    (\textbf{C})~Inter-ocular distance (IOD).
    (\textbf{D})~Contrast.
    (\textbf{E})~Color balance.}
    \label{fig:quality_curves_auc}
\end{figure}

\end{document}